\documentclass{article}

\usepackage[preprint,nonatbib]{nips_2018} % for uploading to arxiv
\usepackage{amsthm}
\usepackage{amssymb}
\usepackage{amsmath}
\usepackage{amsthm}
\usepackage{graphicx}
\usepackage{algorithm}
\usepackage{algorithmic}
 \usepackage{tcolorbox}
 \usepackage{geometry}
 \usepackage{bibentry}
  \usepackage{environ}
\nobibliography*
\usepackage{subfigure}

\newtheorem{theorem}{Theorem}
\newtheorem{lemma}{Lemma}
\newtheorem{assumption}{Assumption}
\newtheorem{remark}{Remark}
\newtheorem{corollary}{Corollary}
\usepackage[utf8]{inputenc} % allow utf-8 input
\usepackage[T1]{fontenc}    % use 8-bit T1 fonts
\usepackage{hyperref}       % hyperlinks
\usepackage{url}            % simple URL typesetting
\usepackage{booktabs}       % professional-quality tables
\usepackage{amsfonts}       % blackboard math symbols
\usepackage{nicefrac}       % compact symbols for 1/2, etc.
\usepackage{microtype}      % microtypography

\newcommand{\EL}{\mathcal{L}}
\newcommand{\MM}{\min_{x\in X}\max_{y\in Y}}
\newcommand{\MMD}{\min_{x\in \Delta_X}\max_{y\in \Delta_Y}}
\newcommand{\MMDt}{\min_{x\in \Delta_{X,\theta}}\max_{y\in \Delta_{Y,\theta}}}
\newcommand{\MMDd}{\min_{x\in \Delta_{X,\delta}}\max_{y\in \Delta_{Y,\delta}}}
\newcommand{\SPreg}{\mathsf{SP}\text{-}\mathsf{Regret}}
\newcommand{\Indreg}{\mathsf{Ind}\text{-}\mathsf{Regret}}
\newcommand{\COCOreg}{\mathsf{Regret}}

\global\long\def\Ex{\mathrm{E}}

\title{The Online Saddle Point Problem and Online Convex Optimization with Knapsacks}

\author{
  Adrian Rivera Cardoso\\
  School of Industrial and Systems Engineering\\
  Georgia Institute of Technology\\
  Atlanta, GA 30332 \\
  \texttt{adrian.riv@gatech.edu} \\
   \And
   He Wang \\
   School of Industrial and Systems Engineering\\
   Georgia Institute of Technology\\
   Atlanta, GA 30332 \\
   \texttt{he.wang@isye.gatech.edu} \\
   \And
   Huan Xu \\
   School of Industrial and Systems Engineering\\
   Georgia Institute of Technology\\
   Atlanta, GA 30332 \\
   \texttt{huan.xu@isye.gatech.edu} \\
}

\begin{document}

\maketitle

\begin{abstract}
We study the online saddle point problem, an online learning problem where at each iteration a pair of actions need to be chosen without knowledge of the current and future (convex-concave) payoff functions.
The objective is to minimize the gap between the cumulative payoffs and the saddle point value of the aggregate payoff function, which we measure using a metric called ``SP-Regret''. The problem generalizes the online convex optimization framework but here we must ensure both players incur cumulative payoffs close to that of the Nash equilibrium of the sum of the games.
We propose an algorithm that achieves SP-Regret proportional to $\sqrt{\ln(T)T}$ in the general case, and $\log(T)$ SP-Regret for the strongly convex-concave case. We also consider the special case where the payoff functions are bilinear and the decision sets are the probability simplex. In this setting we are able to design algorithms that reduce the bounds on SP-Regret from a linear dependence in the dimension of the problem to a \textit{logarithmic} one. We also study the problem under bandit feedback and provide an algorithm that achieves sublinear SP-Regret. We then consider an online convex optimization with knapsacks problem motivated by a wide variety of applications such as: dynamic pricing, auctions, and crowdsourcing. We relate this problem to the online saddle point problem and establish $O(\sqrt{T})$ regret using a primal-dual algorithm.
\end{abstract}

\section{Introduction}\label{Intro}

In this paper, we study the \emph{online saddle point} (OSP) problem. The OSP problem involves a sequence of two-player zero-sum convex-concave games which are selected arbitrarily by Nature.
In each iteration, player 1 chooses an action to minimize its payoffs, while player 2 chooses an action to maximize its payoffs. Both players choose actions without knowledge of the current and future payoff functions.
Our goal is to \emph{jointly} choose a pair of actions for both players at each iteration, such that each player's cumulative payoff at the end
is as close as possible to that of the Nash equilibrium (i.e.\ saddle point) of the aggregate game.

More formally, we define the OSP problem as follows. There is a sequence of unknown functions $\{\EL_t(x, y)\}_{t=1}^T$ that are convex in $x \in X$ and concave in $y \in Y$. Here, $X$ and $Y$ are compact convex sets in Euclidean space. As a result, there exists a saddle point $(x^*, y^*) \in X \times Y$ such that
\[
\MM \sum_{t=1}^T \EL_t(x,y) = \sum_{t=1}^T \EL_t(x^*,y^*) = \max_{y\in Y} \min_{x\in X} \sum_{t=1}^T \EL_t(x,y) .
\]
At each iteration $t$, the decision makers jointly choose a pair of actions $(x_t , y_t) \in X \times Y$, and then the function $\EL_t$ is revealed. 
The goal is to design an algorithm to minimize the cumulative \emph{saddle-point regret} (SP-Regret), defined as 
\begin{equation}\label{eq:sp-reg}
\SPreg (T) = \left|\sum_{t=1}^T \EL_t(x_t,y_t) -\MM \sum_{t=1}^T \EL_t(x,y)\right|.
\end{equation} 
In other words, we would like to obtain a cumulative payoff that is as close as possible to the saddle-point value if we had known all the functions $\{\EL_t\}_{t=1}^T$ in advance.

We would like to emphasize an important distinction between the OSP problem and the standard Online Convex Optimization (OCO) problem \cite{hazan2016introduction}. In the OCO problem, Nature selects an arbitrary sequence of convex functions $\{f_t(\cdot)\}_{t=1}^{T}$, and the decision maker chooses an action $x_t \in X$ before each function $f_t(\cdot)$ is revealed. The objective is to minimize the regret defined as
\[
\sum_{t=1}^T f_t(x_t) - \min_{x\in X} \sum_{t=1}^T f_t(x).
\]
{ The objective in the OSP problem is to choose the actions of two players \emph{jointly} such that the aggregate payoffs of both players are close to the Nash equilibrium payoff.} In contrast, OCO involves only an individual player against Nature.
The OCO framework can be viewed as a {special case} of the OSP problem where the action set of the second player $Y$ is a singleton.
Moreover, the standard OCO setting is applicable to the OSP problem when only one of the players' payoff is optimized at a time.
To be specific, we define the \emph{individual-regret} of players 1 and 2 as
\begin{subequations}\label{eq:indiv-reg}
\begin{align}
\Indreg_x (T) = \sum_{t=1}^T \EL_t(x_t,y_t) -\min_{x\in X}\sum_{t=1}^T \EL_t(x,y_t), \label{eq:indiv-reg-x} \\
\Indreg_y (T) = \max_{y\in Y}\sum_{t=1}^T \EL_t(x_t,y)- \sum_{t=1}^T \EL_t(x_t,y_t). \label{eq:indiv-reg-y}
\end{align}
\end{subequations}
The individual-regret measures each player's own regret while fixing the other player's actions. 
It is easy to see that minimizing individual-regret \eqref{eq:indiv-reg-x} or \eqref{eq:indiv-reg-y} can be cast as a standard OCO problem.

However, we will show that SP-Regret and individual-regret do not imply one another, so existing OCO algorithms cannot be directly applied to the OSP problem. More surprisingly, we show that any OCO algorithm with a sublinear ($o(T)$) individual-regret will inevitably have a linear ($\Omega(T)$) SP-Regret in the general OSP problem (see details in  \S\ref{choose_one}).

In addition to establishing general results for the OSP problem,
we focus on one of its prominent applications: the \emph{online convex optimization with knapsacks} (OCOwK) problem.  
Several variants of the OCOwK problem have recently received a lot of attention in recent literature, but we found its connection to the OSP problem has not been well exploited.
We show that the OCOwK problem is closely related to the OSP problem through Lagrangian duality; thus, we are able to apply our results for the OSP problem to the OCOwK problem.

In the OCOwK problem, a decision maker is endowed with a fixed budget of resource at the beginning of $T$ periods. 
In each period $t=1,\ldots,T$, 
the decision maker chooses an action $x_t \in X$, and then  Nature reveals a reward function $r_t$ and a budget consumption function $c_t$. 
The objective is to maximize total reward $\sum_{t=1}^T r_t(x_t)$ while keeping the total consumption $\sum_{t=1}^T c_t(x_t)$ within the given budget. 

The OCOwK model also generalizes the standard OCO problem by having an additional budget constraint. 
Additionally, it also has a wide range of practical applications (see more discussion in \cite{badanidiyuru2018bandits}), some notable examples include:
\begin{itemize}
\item Dynamic pricing: a retailer is selling a fixed amount of goods in a finite horizon. The actions correspond to pricing decisions, the reward is the retailer's revenue, and the budget represents finite item inventory. The reward functions are unknown initially due to high uncertainty in customer demand.
\item Online ad auction: a firm is bidding for advertising on a platform (e.g. Google) with limited daily budget. The actions refer to auction bids, and the reward represents impressions received from displayed ads. The reward function is unknown because the firm is unaware of other firms' bidding strategies.
\item Crowdsourcing: suppose an organization is purchasing labor tasks on a crowdsourcing platform (e.g. Amazon Mechanical Turk). The actions correspond to prices offered for each micro-task, and the budget corresponds to the maximum amount of money to be spent on acquiring these tasks. The reward functions are unknown a priori because of uncertainty in the crowd's abilities.
\end{itemize}

\subsection{Main Contributions}

%We first propose an algorithm called \textsf{SP-FTL} (Saddle-Point Follow-the-Leader) 
%for the online saddle point problem, and show that this algorithm has a SP-Regret of $\tilde{O}(\sqrt{T})$, which matches the lower bound of $\Omega(\sqrt{T})$ up to a logarithmic factor.
%In the special case where the payoff function $\EL_t(x, y)$ is strongly-convex in $x$ and strongly-concave in $y$, the algorithm has a SP-regret of $O(\log T)$, which is optimal.

{ We first propose an algorithm called \textsf{SP-FTL} (Saddle-Point Follow-the-Leader) 
for the online saddle point problem when the payoff function $\EL_t(x, y)$ is Lipschitz continuous and strongly-convex in $x$ and strongly-concave in $y$, the algorithm has a SP-regret that scales as $\ln(T)$, which is optimal. When the payoff functions are convex-concave we show that a variant of \textsf{SP-FTL} attains a SP-Regret that scales as $\sqrt{\ln(T) T}$, which matches the lower bound of $\Omega(\sqrt{T})$ up to a logarithmic factor.
In the special case where the payoff functions are bilinear and the decision sets are the probability simplex, a setup which we call Online Matrix Games, we show that a variant of \textsf{SP-FTL} can attain SP-Regret that scales almost optimally with $T$ and  \textit{logarithmically} with the dimension of the problem. This is in contrast to the general convex-concave case where the SP-Regret scales \textit{linearly} with the dimension of the problem. 
We also study Online Matrix Games under bandit feedback. Here the players only observe the loss function evaluated at their decisions instead of observing the whole payoff function, this makes the problem significantly more challenging. For this setting we derive an algorithm that attains sublinear SP-Regret. }

In addition, we show that no algorithm can simultaneously achieve sublinear (i.e.\ $o(T)$) SP-Regret and sublinear individual-regrets (defined in \eqref{eq:sp-reg} and
\eqref{eq:indiv-reg}) in the general OSP problem. This impossibility result further illustrates the contribution of the  \textsf{SP-FTL} algorithm, as existing OCO algorithms designed to achieve sublinear individual-regret are not able to achieve sublinear SP-Regret.

Then, we consider the OCOwK problem.  We show that this problem is related to the OSP problem by Lagrangian duality, and a sufficient condition to achieve a sublinear regret for OCOwK is that an algorithm must have \emph{both} sublinear SP-Regret and sublinear individual-regret for the OSP problem. 
In light of the previous impossibility result, we consider the OCOwK problem in a stochastic setting where the reward and consumption functions are sampled i.i.d.\ from some unknown distribution. 
By applying the \textsf{SP-FTL} algorithm and exploiting the connection between OCOwK and OSP problems, we obtain a $\tilde{O}(T^{5/6})$ regret. We then propose a new algorithm called \textsf{PD-RFTL} (Primal-Dual Regularized-Follow-the-Leader) that achieves an $O(\sqrt{T})$ regret bound. The result matches the lower bound $\Omega(\sqrt{T})$ for the OCOwK problem in the stochastic setting. 
We then provide numerical experiments to compare the empirical performances of \textsf{SP-FTL} and \textsf{PD-RFTL}.

\section{Literature Review}

Saddle point problems emerge from a variety of fields such as machine learning, statistics, computer science, and economics. Some applications of the saddle point problem include: minimizing the maximum of smooth convex functions, minimizing the maximal eigenvalue, $l_1$-minimization (an important tool in sparsity-oriented Signal processing), nuclear norm minimization, robust learning problems, and two-player zero-sum games
\cite{presentation_arkadi,lu2007large,cox2017decomposition,ahmadinejad2016duels,myerson1993incentives,laslier2002distributive,chowdhury2013experimental,kovenock2012coalitional}. 

%A few papers have studied the saddle point problem in online learning settings.
%Motivated by joint optimization and estimation problems, Ho-Nguyen and K{\i}l{\i}n{\c{c}}-Karzan \cite{ho2016role}
%consider an online saddle point problem similar to ours. They show that the so-called ``online saddle point gap,'' or using our terminology, the sum of both players' individual-regret, is sublinear. However, they did not consider SP-regret.
%Cesa-Bianchi and Lugosi \cite{cesa2006prediction} provide a detailed overview of online learning for static two-player zero-sum games,  where the (convex-concave) payoffs are given by $\EL_t(x,y)=\EL(x,y)$ for all $t=1,\ldots,T$.
%They show that if both players minimize their individual-regrets, then the average of actions $(\bar{x},\bar{y})$ satisfy $|\EL(\bar{x},\bar{y})-\EL(x^*,y^*)|\to 0$ as $T \to \infty$, where $(x^*,y^*)$ is a Nash equilibrium.
%Abernethy and Wang  \cite{abernethy2017frank}, using the same scheme, establish a connection between online learning and projection-free optimization by considering a specific static two-player zero-sum game, where both players apply individual regret minimization algorithms. They also mention this idea has had applications in boosting, differential privacy, linear programming and flow optimization. This line of research has been continued in \cite{wang2018acceleration,abernethy2018faster}.
{
We now discuss some related works that focus on learning in games. \cite{singh2000nash} study a two player, two-action general sum static game. They show that if both players use Infinitesimal Gradient Ascent, either the strategy pair will converge to a Nash equilibrium (NE), or even if they do not, then the average payoffs are close to that of the NE. A result of similar flavor was derived in \cite{cesa2007improved} for any zero-sum convex-concave game. Given a payoff function $\EL(x,y)$, they show that if both players minimize their individual-regrets, then the average of actions $(\bar{x},\bar{y})$ will satisfy $|\EL(\bar{x},\bar{y})-\EL(x^*,y^*)|\to 0$ as $T \to \infty$, where $(x^*,y^*)$ is a NE.  \cite{bowling2001convergence} improve upon the result of \cite{singh2000nash} by proposing an algorithm called WoLF (Win or Learn Fast), which is a modification of gradient ascent; they show that the iterates of their algorithm indeed converge to a NE. \cite{conitzer2007awesome} further improve the results in \cite{singh2000nash} and \cite{bowling2005convergence} 
%(which introduced and algorithm called GIGA-WoLF) 
by developing an algorithm called GIGA-WoLF for multi-player nonzero sum static games. Their algorithm learns to play optimally against stationary opponents; when used in self-play, the actions chosen by the algorithm converge to a NE. 
More recently, \cite{balduzzi2018mechanics} studied general multi-player static games and show that by decomposing and classifying the second order dynamics of these games, one can prevent cycling behavior to find NE. 
%The decomposition they use motivates an algorithm called Symplectic Gradient Adjustment, which is successfully used to train a Generative Adversarial Network on a toy problem. 
We note that unlike our paper, all of the papers above consider repeated games with a static payoff matrix, whereas we allow the payoff matrix to change arbitrarily.
An exception is the work by \cite{ho2016role}, who consider the same setting as our OMG problem; however their paper only shows that the sum of the individual regrets of both players is sublinear and does not study SP-Regret.
%, which they use to solve a joint optimization and estimation problem.

Related to the OMG problem with bandit feedback is the seminal work of \cite{flaxman2005online}.
They provide the first sublinear regret bound for Online Convex Optimization with bandit feedback,
using a one-point estimate of the gradient. The one-point gradient estimate used in \cite{flaxman2005online} is similar to those independently proposed 
%in \cite{granichin1989stochastic} and 
in \cite{spall1997one}. The regret bound provided in \cite{flaxman2005online} is $O(T^{3/4})$, which is suboptimal. In \cite{abernethy2009competing}, the authors give the first $O(\sqrt{T})$ bound for the special case when the functions are linear. More recently, \cite{hazan2016optimal} and \cite{bubeck2016kernel} designed the first efficient algorithms with $\tilde{O}(poly(d)\sqrt{T})$ regret  for the general online convex optimization case; unfortunately, the dependence on the dimension $d$ in the regret rate is a very large polynomial. Our one-point matrix estimate is most closely related to the random estimator in \cite{auer1995gambling} for linear functions. It is possible to use the more sophisticated techniques from \cite{abernethy2009competing,hazan2016optimal,bubeck2016kernel} to improve our SP-Regret bound in Section~\ref{section:omg_bandit}; however, the result does not seem to be immediate and we leave this as future work. 
 }
 
The Online Convex Optimization with Knapsacks (OCOwK) problem studied in this paper is related to several previous works on constrained multi-armed bandit problems, online linear programming, and online convex programming. We next give an overview of the work related to OCOwK . Agrawal et al. \cite{agrawal2014dynamic}  and Agrawal and Devanur \cite{agrawal2014fast} consider 
online linear/convex programming problems. A key difference between the online linear/convex programming problems and the OCOwK problem is that we assume the action must be chosen without knowledge of the function associated with the current iteration. In \cite{agrawal2014dynamic,agrawal2014fast}, it is assumed that these functions are revealed \emph{before} the action is chosen. Related work is that of Buchbinder and Naor \cite{buchbinder2009online}, where they study an online fractional covering/packing problem, and that of Gupta and Molinaro \cite{gupta2016experts} where they consider a packing/covering multiple choice LP problem in a random permutation model. Another relevant paper is \cite{lan2016algorithms} where the authors provide an algorithm to solve convex problems with expectation constraints, such as the benchmark in Section \ref{OCOwK_setup}. However it is unclear if their optimization algorithm has any sublinear regret properties.  

Mannor et al.\ \cite{mannor2009online} consider a variant of the online convex optimization (OCO) problem where the adversary may choose extra constraints that must be satisfied. They construct an example such that no algorithm can attain an $\epsilon$-approximation to the offline problem. In view of such result, several papers \cite{mahdavi2012online,neely2017online,yu2017online} study problems similar to \cite{mannor2009online} with further restrictions on how constraints are selected by the adversary.
The objective in this line of work is to choose a sequence of decisions to achieve the offline optimum while making sure the constraints are (almost) satisfied. In this line of research, the most relevant work to ours is that of \cite{neely2017online}. They study OCO with time-varying constraints, the model is similar to that of \cite{mannor2009online}, however in view of the existing 
3
 negative results they consider three different settings. In the first one, both the cost functions and the constraints are arbitrary sequences of convex functions, however in view of the negative result from \cite{mannor2009online}, the constraints must all be non positive over a common subset of $\mathbb{R}^n$. In the second setting the sequences of loss functions remain adversarially chosen however the constraints are sampled i.i.d. from some unknown distribution. Finally, in the third setting both the sequences of loss functions and constraints are sampled i.i.d. from some unknown distribution. They develop algorithms for all the three different settings that ensure the total loss incurred by the algorithm is not too far from the offline optimum and such that the constraints are almost satisfied. The setup and results are different than ours because they only require the cumulative constraint violation to be sublinear whereas in OCOwK, once the player exceeds the budget it can no longer collect rewards. Closely related is the problem of ``Online Convex Optimization with Long Term Constraints''. The setup is similar to that of OCO where the functions are chosen adversarially with the difference that it is not required that the decisions the player makes at each step belong to the set. Instead, it is required that the average decision lies in the set (which is fully known in advance). As the authors explain, this problem is useful to avoid the projection step of online gradient descent (OGD) and it allows to solve problems such as multi-objective online classification \cite{bernstein2010online}, and for using the popular online-to-batch conversion. The algorithms they develop consist of simultaneously running two copies of variants of OGD on convex-concave functions. Better rates and slightly different guarantees were obtained for the same problem in \cite{yuan2018online,jenatton2015adaptive,yu2016low}. In \cite{paternain2015online}, the authors study a continuous time version of a problem similar to that of \cite{mahdavi2012online} and show that a continuous time version of primal-dual online gradient descent in continuous time guarantees small regret.
Motivated by an application in low-latency fog computing, \cite{chen2018harnessing} consider a problem similar to that in \cite{mahdavi2012trading} however there is bandit feedback in the loss function. The algorithm proposed in \cite{chen2018harnessing} is primal-dual online gradient descent that combines ideas from \cite{flaxman2005online} to deal with bandit feedback. 

Most closely related to our model is the Bandits with Knapsacks problem studied by Badanidiyuru et al.\ \cite{badanidiyuru2018bandits}
and Wu et al.\ \cite{wu2015algorithms}. In this problem, there is a finite set of arms, and each arm yields a random reward and consumes resources when it is pulled. The goal is to maximize total reward without exceeding a total budget. The Bandits with Knapsacks problem can be viewed as a special case of the OCOwK problem, where the reward and consumption functions are both linear. 
Agrawal and Devanur \cite{agrawal2014bandits} study a generalization of 
bandits with concave rewards and convex knapsack constraints. Similar problems have also been studied in specific application contexts, such as online ad auction
\cite{balseiro2017learning} and dynamic pricing  \cite{besbes2012blind, ferreira2017online}.
{
Recently, \cite{immorlica2019adversarial} study an adversarial version of the multi-armed bandits with knapsack problem. A key part of their algorithm uses a primal-dual approach similar to the one we propose in Section~\ref{OCOwK_setup}.}

\section{Preliminaries}

We introduce some notation and definitions that will be used in later sections. By default, all vectors are column vectors. A vector with entries $x_1,...,x_n$ is written as $x = [x_1;...;x_n] = [x_1,...,x_n]^\top$, where $\top$ denotes the transpose.
{Let $\Vert\cdot\Vert$ be any norm of a vector; the ones we will frequently use are $\Vert\cdot\Vert_2, \Vert\cdot\Vert_1, \Vert\cdot\Vert_\infty$.}  

{We say a function $\EL(x,y)$ is \emph{convex-concave} if it is convex in $x \in X$, for every fixed $y\in Y$,  and concave in $y \in Y$, for every fixed $x\in X$.}
A pair $(x^*,y^*)$ is called a saddle point for $\EL$ if for any $x\in X$ and any $y \in Y$, we have
\begin{equation} \label{def_sp}
\EL(x^*,y) \leq \EL(x^*,y^*) \leq \EL(x,y^*).
\end{equation}
It is well known that if $\EL$ is convex-concave, and $X$ and $Y$ are convex compact sets, there always exists at least one saddle point (see e.g.~\cite{boyd2004convex}).

We say that a function $f:X\rightarrow \mathbb{R}$ is $H$-strongly convex if for any $x_1, x_2 \in X$,  it holds that
\begin{align*}
f(x_1) \geq f(x_2) + \nabla f(x_2)^\top(x_1-x_2) + \frac{H}{2}\Vert x_1-x_2 \Vert^2.
\end{align*}
Here, $\nabla f(x)$ denotes a subgradient of $f$ at $x$. Strong convexity implies that the problem $\min_{x \in X} f(x)$ has
a unique solution. We say a function $g$ is $H$-strongly concave if $-g$ is $H$-strongly convex. 

Furthermore, we say a function $\EL(x,y)$ is $H$-strongly convex-concave if for any fixed $y_0 \in Y$, the function $\EL(x,y_0)$ is $H$-strongly convex in $x$, and for any fixed $x_0\in X$, the function $\EL(x_0,y)$ is $H$-strongly concave in $y$. If $\EL$ is $H$-strongly convex-concave, then there exists a unique saddle point.

{
We say a function $\EL(x,y)$ is $G$-Lipschitz continuous with respect to norm $\Vert\cdot\Vert$, if 
\begin{align*}
|\EL(x_1, y_1)- \EL(x_2, y_2)| \leq G \Vert [x_1;y_1] - [x_2;y_2] \Vert.
\end{align*}
It is well known that the previous inequality holds if and only if
\begin{align*}
\Vert [\nabla_x \EL(x,y); \nabla_y \EL(x,y)] \Vert_* \leq G 
\end{align*}
for all $x\in X$, $y \in Y$ where $\Vert\cdot\Vert_*$ is the norm dual to $\Vert\cdot\Vert$ (see Lemma 2.6 in \cite{shalev2012online}).
}

Throughout the paper we will use the big $O$ notation to hide constant factors. 
For two functions $f(T)$ and $g(T)>0$, we write $f(T)=O(g(T))$ if there exists a constant $M_1$ and a constant $T_1$ such that $f(T) \leq M_1 g(T)$ for all $T \geq T_1$; we write $f(T)=\Omega(g(T))$ if there exists a constant $M_2$ and a constant $T_2$ such that $f(T) \geq M_2 g(T)$ for all $T \geq T_2$.
We use the $\tilde{O}$ notation to hide constant factors and poly-logarithmic factors. More specifically, for two functions $f(T)$ and $g(T)>0$, we write $f(T)=\tilde{O}(g(T))$ if there exists constants $M_3$, $T_3$ and an integer $k\geq 0$ such that $f(T) \leq M_3 g(T)\log^k (g(T))$ for all $T \geq T_3$. 

\section{The Online Saddle Point Problem}\label{setup_problem}
%As mentioned in Section \ref{Intro} the OCO setup makes explicit consideration of only one player. Since there are many examples of decision making environments where several players (with conflicting goals) interact with each other we generalize the setting of OCO and explicitly incorporate into the game a second player. 
%We now formally describe the Online Saddle Point (OSP) problem.
%
%A game between two players and an adversary will be played for $T$ time steps. In every time step $t=1,...,T$ the adversary will choose function $\EL_t(x,y)$ which is $H$-strongly convex in $x$ and $H$-strongly concave in $y$ (we later relax this to convex-concave games). Before observing $\EL_t$, Player 1 chooses action $x_t\in X$ and Player 2 chooses action $y_t\in Y$ (here $X$ and $Y$ are convex and bounded sets). After the decisions have been made Player 1 incurs a loss of $\EL_t(x_t,y_t)$ and Player 2 gains $\EL_t(x_t,y_t)$ (i.e. $\EL_t$ is a zero-sum game) then both players observe $\EL_t$.
%
%The goal for both players is to obtain sublinear saddle point regret (SP Regret) defined as 
%\begin{equation}
% |\sum_{t=1}^T \EL_t(x_t,y_t) -\MM \sum_{t=1}^T \EL_t(x,y)|. 
%\end{equation} 
%That is, both players want to make sure that their average loss$\slash$gain is not too far from that of the Nash Equilibrium of the average game in hindsight.  

\subsection{The Strongly Convex-Concave Case}
We now present algorithms for the OSP problem with guaranteed sublinear SP-Regret. Recall that the SP-regret defined in \eqref{eq:sp-reg} measures the gap between the cumulative value achieved by an online algorithm and the value of the game under the Nash equilibrium if all functions are known in hindsight.

For simplicity we assume $T$ is known in advance (this assumption can be relaxed using the well known doubling trick from \cite{cesa2006prediction,shalev2012online}). We first consider the case where the functions $\{\EL_t\}_{t=1}^T$ are strongly convex-concave.
We show that the following simple algorithm Saddle-Point Follow-the-Leader (\textsf{SP-FTL}), which is
a variant of the Follow-the-Leader (\textsf{FTL}) algorithm by Kalai and Vempala \cite{kalai2002}, attains sublinear SP-Regret. 

\begin{algorithm}[tbh]
\caption{Saddle-Point Follow-the-Leader (\textsf{SP-FTL})}
\label{alg: SP-FTL}
\begin{algorithmic}
  \STATE {\bfseries input:} $x_1 \in X$, $y_1 \in Y$
  \FOR{$t=1,..., T$}
  \STATE \quad Choose actions $(x_t, y_t)$
  \STATE \quad Observe function $\EL_t$
  \STATE \quad Set $x_{t+1} \leftarrow \arg \min_{x\in X} \max_{y\in Y} \sum_{\tau=1}^t \EL_{\tau}(x,y)$ 
  \STATE \quad Set $y_{t+1} \leftarrow \arg  \max_{y\in Y} \min_{x\in X} \sum_{\tau=1}^t \EL_{\tau}(x,y)$ 
  \ENDFOR
\end{algorithmic}
\end{algorithm}

The main difference between \textsf{SP-FTL} and \textsf{FTL} is that in \textsf{SP-FTL} both players update jointly and play the (unique) saddle point of the sum of the games observed so far. In contrast, the updates for Follow-the-Leader would be $x^{FTL}_{t+1}\leftarrow \arg \min_{x\in X} \sum_{\tau=1}^t \EL_\tau(x, y^{FTL}_\tau)$ and $y^{FTL}_{t+1}\leftarrow \arg \max_{y\in Y}  \sum_{\tau=1}^t \EL_\tau(x^{FTL}_\tau, y)$ for $t=2,...,T$ and $x^{FTL}_1$, $y^{FTL}_1$ are arbitrarily chosen from their respective sets $X$ and $Y$. It is easy to see that the sequence of iterates is in general not the same. In fact, in view of Theorem~\ref{thm:impossible} we will see that \textsf{FTL} can not achieve sublinear $\SPreg$ when the sequence of functions is chosen arbitrarily. 

\begin{theorem} \label{thm:sp_regret_str} 
Let $\{\EL_t(x,y)\}_{t=1}^T$ be any  sequence of $H$-strongly convex-concave,  $G$-Lipschitz functions. Then, the $\mathsf{SP}$-$\mathsf{FTL}$ algorithm guarantees
\begin{align*}
\SPreg (T) =
\Bigl| \sum_{t=1}^T \EL_t(x_t,y_t) -\MM \sum_{t=1}^T \EL_t(x,y) \Bigr| \leq \frac{8G^2}{H} (1+ \log T).
\end{align*}
\end{theorem}

{We remark that since Theorem \ref{thm:sp_regret_str} holds against \textit{all} sequences of functions $\{\EL_t(x,y)\}_{t=1}^T$, this means that the sequence $\{\EL_t(x,y)\}_{t=1}^T$ can even be chosen by an adaptive adversary. In particular, this means that $\EL_t(x,y)$ can be a function of all the previous iterates $\{(x_\tau,y_\tau)\}_{\tau=1}^{t-1}$.
%and even the iterates $(x_t,y_t)$. 
}

The proof of Theorem~\ref{thm:sp_regret_str} is based on the following two lemmas. We first analyze a quantity that is similar to SP-Regret, but with actions $(x_t, y_t)$ replaced by $(x_{t+1},y_{t+1})$ (Lemma~\ref{loss_BTL}).
This analysis framework is known as the Follow-the-Leader vs.\ Be-the-Leader scheme \cite{kalai2002}. 
We then show that consecutive iterates of \textsf{SP-FTL} have distances diminishing proportionally to $1/t$. The proof heavily utilizes the KKT conditions associated with points  $(x_t, y_t)$ and $(x_{t+1},y_{t+1})$ (Lemma~\ref{FTL_close_iterates}).

\begin{lemma}\label{loss_BTL} 
{
Let $\{\EL_t\}_{t=1}^T$ be an arbitrary sequence of convex-concave functions that are $G$-Lipschitz with respect to norm $\Vert \cdot \Vert$. Here, $\EL_t:X \times Y \rightarrow \mathbb{R}$, where $X\subset \mathbb{R}^{d_1}$ and $Y\subset \mathbb{R}^{d_2}$ are convex compact sets. Let $\{(x_t,y_t)\}_{t=1}^T$ be the iterates of $\mathsf{SP}$-$\mathsf{FTL}$ when run on the sequence of functions $\{\EL_t\}_{t=1}^T$. It holds that
\begin{equation}
-G \sum_{t=1}^T \Vert x_t-x_{t+1} \Vert \leq \sum_{t=1}^T \EL_t(x_{t+1},y_{t+1}) - \MM \sum_{t=1}^T \EL_t(x,y) \leq  G \sum_{t=1}^T \Vert y_t-y_{t+1} \Vert.
\end{equation}
}
\end{lemma}

\proof{Proof.}
We first prove the second inequality, namely
\[
\sum_{t=1}^T \EL_t(x_{t+1},y_{t+1}) - \MM \sum_{t=1}^T \EL_t(x,y) \leq  G \sum_{t=1}^T \Vert y_t-y_{t+1} \Vert.
\]
We proceed by induction. The base case $t=1$ holds by definition of $(x_2,y_2)$:
\[
\EL_1(x_2,y_2) - \MM \EL_1(x,y) = \EL_1(x_2,y_2) - \EL_1(x_2,y_2) = 0 \leq G \Vert y_1-y_2\Vert.
\]
We now assume the following claim holds for $T-1$,
\begin{equation}\label{ind_hypo}
\MM \sum_{t=1}^{T-1} \EL_t(x,y)  \geq \sum_{t=1}^{T-1} \EL_t(x_{t+1},y_{t+1}) - G \sum_{t=1}^{T-1} \Vert y_t-y_{t+1}\Vert,
\end{equation}
and show it holds for $T$.  By definition of $(x_{T+1},y_{T+1})$, we have
\begin{align*}
\MM \sum_{t=1}^T \EL_t(x,y) = & \sum_{t=1}^{T-1} \EL_t(x_{T+1},y_{T+1}) + \EL_T(x_{T+1},y_{T+1})\\
\geq & \sum_{t=1}^{T-1} \EL_t(x_{T+1},y_T) + \EL_{T}(x_{T+1},y_T)\\
\geq &\sum_{t=1}^{T-1} \EL_t(x_T,y_T) + \EL_{T}(x_{T+1},y_T).
\end{align*}
The first inequality holds because $(x_{T+1},y_{T+1})$ is the saddle point of $\sum_{t=1}^T\EL_t(x,y)$, so $y_{T+1}$ is a maximizer for function $\sum_{t=1}^T\EL_t(x_{T+1},y)$, see Equation \eqref{def_sp}. 
Similarly, the second inequality follows since $(x_T,y_T)$ is the saddle point of $\sum_{t=1}^{T-1} \EL_t (x,y)$.
By the induction hypothesis, see Equation \eqref{ind_hypo}, and the definition of $(x_T,y_T)$ we have
\begin{align*}
& \sum_{t=1}^{T-1} \EL_t(x_T,y_T) + \EL_{T}(x_{T+1},y_T)\\
 \geq & \sum_{t=1}^{T-1} \EL_t(x_{t+1},y_{t+1}) - G \sum_{t=1}^{T-1} \Vert y_t-y_{t+1}\Vert + \EL_{T}(x_{T+1},y_T)\\
= & \sum_{t=1}^{T} \EL_t(x_{t+1},y_{t+1})- G \sum_{t=1}^{T-1} \Vert y_t-y_{t+1}\Vert+ \EL_{T} (x_{T+1},y_T) - \EL_{T}(x_{T+1},y_{T+1})\\
\geq & \sum_{t=1}^{T} \EL_t(x_{t+1},y_{t+1})- G \sum_{t=1}^{T-1} \Vert y_t-y_{t+1}\Vert - G \Vert y_T - y_{T+1}\Vert \qquad \qquad \text{since $\EL_T$ is $G$-Lipschitz}\\
= & \sum_{t=1}^{T} \EL_t(x_{t+1},y_{t+1})- G \sum_{t=1}^{T} \Vert y_t-y_{t+1}\Vert.
\end{align*}
This proves the second inequality in the lemma.

Using a similar argument, we now show by induction the first inequality in the statement of the lemma, namely that 
\begin{align*}
\MM \sum_{t=1}^T \EL_t(x,y) \leq \sum_{t=1}^T \EL_t(x_{t+1},y_{t+1}) + G \sum_{t=1}^T \Vert x_t-x_{t+1}\Vert.   
\end{align*}
Indeed, $t=1$ follows from the definition of $(x_2,y_2)$:
\begin{align*}
\MM \EL_1(x,y) - \EL_1(x_2,y_2) = \EL_1(x_2,y_2) - \EL_1(x_2,y_2 = 0 \leq G \Vert x_1,x_2 \Vert.
\end{align*}
We now assume the following  claim holds for $T-1$,

\begin{equation}\label{ind_hypo_2}
\MM \sum_{t=1}^{T-1} \EL_t(x,y) \leq \sum_{t=1}^{T-1} \EL_t(x_{t+1},y_{t+1}) + G \sum_{t=1}^{T-1} \Vert x_t-x_{t+1}\Vert
\end{equation}
and prove it for $T$.
By definition of $(x_{T+1},y_{T+1})$, we have
\begin{align*}
 \MM \sum_{t=1}^T \EL_t(x,y) &=  \sum_{t=1}^T \EL_t(x_{T+1},y_{T+1})\\ 
 &\leq \sum_{t=1}^{T-1} \EL_t(x_{T},y_{T+1}) + \EL_T(x_T, y_{T+1})\\
 &\leq \sum_{t=1}^{T-1} \EL_t(x_{T},y_{T}) + \EL_T(x_T, y_{T+1}).
 \end{align*}
 
 The first inequality holds because $(x_{T+1},y_{T+1})$ is the saddle point of $\sum_{t=1}^T \EL_t(x,y)$, so $x_{T+1}$ is a minimizer of $\sum_{t=1}^T \EL_t(x,y_{T+1})$, see Equation \eqref{def_sp}. Similarly, the second inequality follows since $(x_T,y_T)$ is the saddle point of $\sum_{t=1}^{T-1}\EL_t(x,y)$ so $y_T$ is the maximizer of $\sum_{t=1}^{T-1}\EL_t(x_T,y)$. 
 
 By the induction hypothesis (see Equation \eqref{ind_hypo_2}) and the definition of $(x_T,y_T)$, we have
 \begin{align*}
 & \sum_{t=1}^{T-1} \EL_t(x_{T},y_{T}) + \EL_T(x_T, y_{T+1})\\
 &\leq \sum_{t=1}^{T-1} \EL_t(x_{t+1},y_{t+1}) + G \sum_{t=1}^{T-1}\Vert
 x_{t}-x_{t+1}\Vert + \EL_T(x_T, y_{T+1})\\
 & =  \sum_{t=1}^{T} \EL_t(x_{t+1},y_{t+1}) + G \sum_{t=1}^{T-1}\Vert x_{t}-x_{t+1}\Vert + \EL_T(x_T, y_{T+1}) - \EL_T(x_{T+1},y_{T+1})\\
 &\leq  \sum_{t=1}^{T} \EL_t(x_{t+1},y_{t+1}) + G \sum_{t=1}^{T}\Vert x_{t}-x_{t+1}\Vert, \qquad \qquad \text{since $\EL_T$ is $G$-Lipschitz}. 
\end{align*}
This concludes the proof. 
\endproof

{
\begin{lemma}\label{FTL_close_iterates} Let $\{\EL_t\}_{t=1}^T$ be an arbitrary sequence of $H$-strongly convex-concave functions (with respect to norm $\Vert \cdot \Vert$) which is also $G$-Lipschitz with respect to the same norm. Let $\{(x_t,y_t)\}_{t=1}^T$ be the iterates of $\mathsf{SP}$-$\mathsf{FTL}$ run on the sequence $\{\EL_t\}_{t=1}^T$. It holds that
\begin{align*}
\Vert x_t - x_{t+1}\Vert + \Vert y_t - y_{t+1}\Vert \leq \frac{4G}{H t}.
\end{align*}
\end{lemma}

\proof{Proof.}
Consider a fixed period $t$. Define 
\begin{align*}
J(x,y) \triangleq \sum_{\tau=1}^{t-1} \EL_\tau (x,y) + \EL_t(x,y)
\end{align*}
so that $(x_{t+1},y_{t+1})$ is a saddle point of $J$. Since $J$ is $Ht$-strongly convex it holds that for any $x\in X$ and any $y\in Y$
\begin{align*}
J(x,y) \geq J(x_{t+1},y) + \nabla_x J(x_{t+1},y)^{\top}(x - x_{t+1})  + \frac{Ht}{2}\Vert x-x_{t+1}\Vert^2.
\end{align*}
Plugging in $y=y_{t+1}$ and recalling the KKT condition $\nabla_x J(x_{t+1},y_{t+1})^{\top}(x - x_{t+1}) \geq 0$ (see Chapter 2 in \cite{hazan2016introduction}), we have that for any $x\in X$
\begin{equation}\label{J_KKT_convex}
\frac{2}{Ht}\big[ J(x,y_{t+1}) - J(x_{t+1},y_{t+1})\big] \geq \Vert x - x_{t+1}\Vert^2.
\end{equation} 
Similarly, since $J$ is $Ht$-strongly concave, for any $y\in Y$ we have
\begin{align*}
J(x_{t+1},y) \leq J(x_{t+1},y_{t+1}) + \nabla_y J(x_{t+1},y_{t+1})^{\top}(y-y_{t+1}) - \frac{Ht}{2} \Vert y-y_{t+1}\Vert^2.
\end{align*}
Together with the KKT condition $\nabla_y J(x_{t+1},y_{t+1})^{\top}(y-y_{t+1})\leq 0$, we get that for any $y\in Y$
\begin{equation} \label{J_KKT_concave}
\frac{2}{Ht}\big[ J(x_{t+1},y_{t+1}) - J(x_{t+1},y)\big] \geq \Vert y-y_{t+1}\Vert^2.
\end{equation}
Adding up Equations \eqref{J_KKT_convex} and \eqref{J_KKT_concave}, plugging $x=x_t$ and $y=y_t$, we get
\begin{align*}
& \frac{2}{Ht} \big[ J(x_t,y_{t+1})-J(x_{t+1},y_t) \big] \geq \Vert x_t - x_{t+1}\Vert^2 + \Vert y_t-y_{t+1}\Vert^2.
\end{align*}
Plugging in the definition of function $J(\cdot)$, we get
\begin{align*}
 \frac{2}{Ht} \big[ \sum_{\tau=1}^{t-1}\EL_\tau (x_t,y_{t+1}) + \EL_t (x_t,y_{t+1}) - [ \sum_{\tau=1}^{t-1} \EL_{\tau}(x_{t+1},y_t) + \EL_t (x_{t+1},y_t)]\big]  \geq \Vert x_t - x_{t+1}\Vert^2 + \Vert y_t-y_{t+1}\Vert^2.
\end{align*}
Since $(x_t,y_t)$ is the saddle point of $\sum_{\tau=1}^{t-1} \EL_\tau(x,y)$, it holds that $\sum_{\tau=1}^{t-1}\EL_\tau (x_t,y_{t+1}) \leq \sum_{\tau=1}^{t-1}\EL_\tau (x_t,y_{t})$. Therefore, we have
\begin{align*}
 &\frac{2}{Ht} \big[ \sum_{\tau=1}^{t-1}\EL_\tau (x_t,y_{t}) + \EL_t (x_t,y_{t+1}) - [ \sum_{\tau=1}^{t-1} \EL_{\tau}(x_{t+1},y_t) + \EL_t (x_{t+1},y_t)]\big]  \geq \Vert x_t - x_{t+1}\Vert^2 + \Vert y_t-y_{t+1}\Vert^2.
\end{align*} 
Additionally, since $(x_t,y_t)$ is the saddle point of $\sum_{\tau=1}^{t-1} \EL_\tau(x,y)$, it holds that %$\sum_{\tau=1}^{t-1}\EL_\tau (x_t,y_{t}) \leq \sum_{\tau=1}^{t-1}\EL_\tau (x_{t+1},y_{t})$, or which is that same: 
$- \sum_{\tau=1}^{t-1}\EL_\tau (x_t,y_{t}) \geq -\sum_{\tau=1}^{t-1}\EL_\tau (x_{t+1},y_{t})$. This implies
\begin{align*}
&\frac{2}{Ht} \big[ \sum_{\tau=1}^{t-1}\EL_\tau (x_t,y_{t}) + \EL_t (x_t,y_{t+1}) -  \sum_{\tau=1}^{t-1} \EL_{\tau}(x_{t},y_t) - \EL_t (x_{t+1},y_t) \big] \geq \Vert x_t - x_{t+1}\Vert^2 + \Vert y_t-y_{t+1}\Vert^2.
\end{align*}
Notice the two summations cancel, thus
\begin{align*}
\frac{2}{Ht} \big[  \EL_t (x_t,y_{t+1}) - \EL_t (x_{t+1},y_t) \big]  \geq \Vert x_t - x_{t+1}\Vert^2 + \Vert y_t-y_{t+1}\Vert^2.
\end{align*}

Since $\EL_t$ is $G$-Lipschitz with respect to norm $\Vert \cdot \Vert$, it holds that
\begin{align*}
& \frac{2}{Ht} G \Vert[x_t;y_{t+1}] - [x_{t+1};y_t]\Vert   \geq \Vert x_t - x_{t+1}\Vert^2 + \Vert y_t-y_{t+1}\Vert^2,
\end{align*}
which then implies that
\begin{align*}
\frac{2}{Ht} G\big[ \Vert x_t - x_{t+1}\Vert + \Vert y_t - y_{t+1}\Vert\big]   \geq \Vert x_t - x_{t+1}\Vert^2 + \Vert y_t-y_{t+1}\Vert^2.
\end{align*}
Rearranging the terms of the inequality above, we get
\begin{align*}
\frac{2G}{Ht} \geq \frac{\Vert x_t - x_{t+1}\Vert^2 + \Vert y_t-y_{t+1}\Vert^2}{\Vert x_t - x_{t+1}\Vert + \Vert y_t - y_{t+1}\Vert}.
\end{align*}
%Since $x^2$ is a convex function, by Jensen's inequality it holds that for any $a,b \in \mathbb{R}$, $\frac{a^2}{2}  + \frac{b^2}{2}  \geq \big( \frac{a+b}{2}\big)^2$. 
For any $a,b \in \mathbb{R}$, we have $a^2 + b^2 \geq \frac{(a+b)^2}{2}$, which implies
\begin{align*}
\Vert x_t - x_{t+1}\Vert^2 + \Vert y_t-y_{t+1}\Vert^2 \geq \frac{\left(\Vert x_t - x_{t+1}\Vert + \Vert y_t-y_{t+1}\Vert \right)^2}{2}.
\end{align*}
Therefore, we have
\begin{align*}
 \frac{2G}{Ht} \geq \frac{ \left(\Vert x_t - x_{t+1}\Vert + \Vert y_t-y_{t+1}\Vert \right)^2}{2 \left(\Vert x_t - x_{t+1}\Vert + \Vert y_t - y_{t+1}\Vert \right)}.
\end{align*}
Rearranging the terms, we get
\begin{align*} 
\frac{4G}{Ht}\geq \Vert x_t - x_{t+1}\Vert + \Vert y_t - y_{t+1}\Vert. 
\end{align*}
This concludes the proof. 

\endproof
}

Now we are ready to prove Theorem  \ref{thm:sp_regret_str}.

\proof{Proof of Theorem \ref{thm:sp_regret_str}.}
We first prove one side of the inequality,
\begin{align*}
 \sum_{t=1}^T \EL_t(x_t,y_t) -\MM \sum_{t=1}^T \EL_t(x,y) \leq \frac{8G^2}{H} (1+ \ln T).
\end{align*}
We have
\begin{align*}
&\quad \sum_{t=1}^T \EL_t(x_t,y_t) -\MM \sum_{t=1}^T \EL_t(x,y) \\
&\leq \sum_{t=1}^T \EL_t(x_t,y_t) - \sum_{t=1}^T \EL_t(x_{t+1},y_{t+1}) + G \sum_{t=1}^T \Vert y_t-y_{t+1}\Vert & \text{by Lemma \ref{loss_BTL}}\\
& \leq \sum_{t=1}^T G \Vert [x_t;y_t] - [x_{t+1};y_{t+1}]\Vert + G \sum_{t=1}^T \Vert y_t-y_{t+1}\Vert & \text{since $\EL_t$ is $G$-Lipschitz}\\
&\leq G\sum_{t=1}^T  \Vert x_t - x_{t+1}\Vert + \Vert y_t- y_{t+1}\Vert + G \sum_{t=1}^T \Vert y_t-y_{t+1}\Vert \\
&\leq G\sum_{t=1}^T \frac{4G}{Ht} + G \sum_{t=1}^T \frac{4G}{Ht} & \text{by Lemma \ref{FTL_close_iterates}}\\
&\leq \frac{8G^2}{H} (1+ \int_{1}^T \frac{1}{t}dt)\\
&=  \frac{8G^2}{H} (1+ \ln T).
 \end{align*}
 
 {
We now prove the other side of the inequality,
\begin{align*}
\MM \sum_{t=1}^T \EL_t(x,y) - \sum_{t=1}^T \EL_t(x_t,y_t) \leq \frac{8G^2}{H} (1+ \ln T),
\end{align*}
using a similar argument.  In particular, we have
\begin{align*}
& \quad \MM \sum_{t=1}^T \EL_t(x,y) - \sum_{t=1}^T \EL_t(x_t,y_t) \\
& \leq \sum_{t=1}^T \EL_t(x_{t+1},y_{t+1}) - \sum_{t=1}^T \EL_t(x_t,y_t) + G \sum_{t=1}^T \Vert x_t - x_{t+1} \Vert & \text{by Lemma \ref{loss_BTL}}\\
& \leq \sum_{t=1}^T G \Vert [x_t;y_t] - [x_{t+1};y_{t+1}]\Vert + G \sum_{t=1}^T \Vert x_t-x_{t+1}\Vert & \text{since $\EL_t$ is $G$-Lipschitz}\\
&\leq G\sum_{t=1}^T  \Vert x_t - x_{t+1}\Vert + \Vert y_t- y_{t+1}\Vert + G \sum_{t=1}^T \Vert x_t-x_{t+1}\Vert \\
&\leq G\sum_{t=1}^T \frac{4G}{Ht} + G \sum_{t=1}^T \frac{4G}{Ht} & \text{by Lemma \ref{FTL_close_iterates}}\\
&\leq \frac{8G^2}{H} (1+ \int_{1}^T \frac{1}{t}dt)\\
&=  \frac{8G^2}{H} (1+ \ln T).
\end{align*}
This concludes the proof. 
}
\endproof

We note that the rate in Theorem~\ref{thm:sp_regret_str} is optimal with respect to $T$,
since when $Y$ is a singleton, the problem reduces to the OCO problem with strongly convex loss functions.
In that case, it is well known that no algorithm can achieve regret smaller than $\Omega(\frac{G^2}{H} \log (T))$ \cite{hazan2014beyond}. 

\subsection{The General Convex-Concave Case}

In this section we propose an algorithm to solve the online Saddle Point Problem in the full information setting when the payoff functions are arbitrary convex-concave Lipschitz functions, and the action sets of Player 1 and Player 2 ($X\subset \mathbb{R}^{d_1}$ and $Y\subset \mathbb{R}^{d_2}$ respectively) to be arbitrary convex compact sets.

Let the sequence of convex-concave functions be $\{\bar{\EL}_t(x,y)\}_{t=1}^T$, which are $G_{\bar{\EL}}$-Lipschitz with respect to some norm $\Vert \cdot \Vert$. 
We propose an algorithm called Saddle Point Regularized Follow the Leader (\textsf{SP-RFTL}), shown in Algorithm~\ref{alg:SPRFTL}.

\begin{algorithm}[tbh]
\caption{Saddle-Point Regularized-Follow-the-Leader (\textsf{SP-RFTL})}
\label{alg:SPRFTL}
\begin{algorithmic}
  \STATE {\bfseries input:} $x_1 \in X$, $y_1 \in Y$, parameters: $\eta>0$, strongly convex functions $R_X$, $R_Y$
  \FOR{$t=1,...T$}
  \STATE \quad Play $(x_t,y_t)$ 
  \STATE \quad Observe $\bar{\EL}_t$
  \STATE \quad $\EL_t(x,y) \gets \bar{\EL}_t + \frac{1}{\eta}R_X(x)-\frac{1}{\eta}R_Y(y)$
  \STATE \quad $x_{t+1}\leftarrow \arg \min_{x\in X} \max_{y \in Y} \sum_{\tau=1}^t \EL_t(x,y)$
  \STATE \quad $y_{t+1}\leftarrow \arg \max_{y \in Y} \min_{x\in X}  \sum_{\tau=1}^t \EL_t(x,y) $
  \ENDFOR
\end{algorithmic}
\end{algorithm}

The regularizers $R_X, R_Y$ are used as input for the algorithm. We will choose regularizers that are strongly convex with respect to norm $ \Vert \cdot \Vert $, and $G_{R_1}$ and $G_{R_2}$ Lipschitz with respect to norm $ \Vert \cdot \Vert $, which means that $\Vert \nabla R_X(x)\Vert_* \leq G_{R_1}$ for all $x\in X$, and $\Vert \nabla R_Y(y)\Vert_* \leq G_{R_2}$ for, all $y\in Y$. Finally, we assume $R_X(x)\geq 0 $ for all $x\in X$ and $R_Y(y)\geq 0 $ for all $y\in Y$. 

We have the following guarantee for \textsf{SP-RFTL}.

{
\begin{theorem}\label{theorem:sp_regret_convex_concave}
Let $X\subset \mathbb{R}^{d_1}$ and $Y\subset \mathbb{R}^{d_2}$ be convex and compact sets. Let $\{\bar{\EL}_{t}(x,y)\}_{t=1}^T$ be any sequence of convex-concave functions. For $t=1,...,T$, let $\bar{\EL}_{t}$ be $G_{\bar{\EL}}$-Lipschitz with respect to norm $\Vert \cdot \Vert$. Let $R_X$, $R_Y$ be two strongly convex regularization functions with respect to the same norm, and let $G_{R_X},G_{R_Y}$ be the Lipschitz constants of $R_X$, $R_Y$. Let $\{(x_t,y_t)\}_{t=1}^T$ be the iterates generated by \textsf{SP-RFTL} when run on the sequence $\{\bar{\EL}_{t}(x,y)\}_{t=1}^T$. It holds that 
\begin{align*}
& \left| \sum_{t=1}^T \bar{\EL}_t(x_t,y_t) - \MM \sum_{t=1}^T \bar{\EL}_t(x,y) \right|  \\
\leq & 8 \eta \left[G_{\bar{\EL}}+ \frac{1}{\eta}\max(G_{R_X}, G_{R_Y})\right]^2 ( 1 + \ln(T) )+ \frac{T}{\eta} \max_{y\in Y} R_Y(y) +  \frac{T}{\eta} \max_{x\in X} R_X(x),
\end{align*}
where $\eta>0$ is the parameter chosen in Algorithm~\ref{alg:SPRFTL}. 
\end{theorem}

As a corollary, we have the following result that shows our algorithm guarantees a sublinear SP-Regret.

\begin{corollary}\label{first_corollary}
Let $X\subset \mathbb{R}^{d_1}$ and $Y\subset \mathbb{R}^{d_2}$ be convex and compact sets containing the origin such that $\max_{x\in X}\Vert x \Vert_2, \max_{y\in Y}\Vert y \Vert_2 \leq D$ for some $0 < D \in \mathbb{R}$. Let $R_X(x) = \Vert x \Vert_2^2$ and $R_Y(y) = \Vert y \Vert_2^2$. For $t=1,...,T$, let $\bar{\EL}_{t}$ be $G_{\bar{\EL}}$-Lipschitz with respect to norm $\Vert \cdot \Vert_2$. Setting $\eta = \frac{D \sqrt{T}}{G_{\bar{\EL}}\sqrt{\ln(T)}}$ in \textsf{SP-RFTL} guarantees
\begin{align*}
\left| \sum_{t=1}^T \bar{\EL}_t(x_t,y_t) - \MM \sum_{t=1}^T \bar{\EL}_t(x,y) \right| \leq O\left(G_{\bar{\EL}} D \sqrt{\ln(T) T}\right),
\end{align*}
where the $O(\cdot)$ notation hides an absolute constant.

\proof{Proof of Corollary \ref{first_corollary}.}
We will instantiate the result from Theorem \ref{theorem:sp_regret_convex_concave}. By our choice of regularizers, we have $\max_{x\in X} R_X(x), \max_{y\in Y} R_Y(y) \leq D^2$, as well as $\Vert \nabla R_X (x) \Vert_2 \leq 2 D$. Thus, we have 
\begin{align*}
 &\left| \sum_{t=1}^T \bar{\EL}_t(x_t,y_t) - \MM \sum_{t=1}^T \bar{\EL}_t(x,y) \right| \leq 8 \eta \left[G_{\bar{\EL}}+ \frac{2D}{\eta}\right]^2 ( 1 + \ln(T) )+ \frac{D^2 T}{\eta}  +  \frac{D^2 T}{\eta}.
\end{align*}
Since for any $a,b \in \mathbb{R}$ it holds that $(a+b)^2\leq 2a^2 + 2b^2$, we have
\begin{align*}
8 \eta \left[G_{\bar{\EL}}+ \frac{2D}{\eta}\right]^2 ( 1 + \ln(T) )+ \frac{D^2 T}{\eta}  +  \frac{D^2 T}{\eta} & \leq 8\eta \left( 2 G_{\bar{\EL}}^2 + \frac{8D^2}{\eta^2}\right) ( 1 + \ln(T) )+ \frac{2 D^2 T}{\eta}\\
& \leq O\left(G_{\bar{\EL}} D \sqrt{\ln(T) T}\right),
\end{align*}
where the last inequality follows by the choice of $\eta$ and the $O(\cdot)$ hides an absolute constant. 

\endproof

\end{corollary}

We note that the bound in Corollary \ref{first_corollary} is optimal up to the $\sqrt{\ln(T)}$ factor. This is because our setup is a special case of Online Convex Optimization and there is a known lower bound $\Omega(d \sqrt{T})$, see Chapter 3 in \cite{hazan2016introduction}. In our setup if $\bar{\EL}_t$ is bilinear, that is $\bar{\EL}_t = x^{\top} A_t y$ for some $d_1 \times d_2$ matrix $A_t$ with bounded entries, and $X,Y$ are unit boxes of dimensions $d_1$ and $d_2$ respectively, we have $G_{\bar{\EL}}, D = O(\max\{d_1,d_2\})$ and so our SP-Regret bound becomes $O(\max\{d_1,d_2\} \sqrt{\ln(T)\sqrt{T}})$.

\begin{remark}[Computational Complexity]
Although the focus of our work is mainly concerned with showing sublinear rate of SP-Regret, it is worth discussing the computation complexity for each iteration of our algorithms. Notice that in each iteration we must solve a strongly convex strongly concave constrained saddle point problem. It is well known that by simultaneously playing two no Individual Regret algorithms for strongly convex functions (such as those in \cite{hazan2007logarithmic} which achieve Individual Regret $O(\log(K)))$,  one can generate  after $K$ rounds a solution to the problem that is $O(\log(K)/K)$ close to the Nash equilibrium (in terms of the value of the game) (See Theorem 9 in \cite{abernethy2018faster}). Recently \cite{abernethy2018faster} showed that with additional smoothness assumptions it is possible to obtain linear convergence rates for some static saddle point problems. It is also possible to solve the subproblem for each iteration using the (Stochastic Approximation) Mirror Descent algorithm from \cite{nemirovski2009robust}. All the previously discussed algorithms are variants of the seminal work of \cite{arrow1958studies}.
\end{remark}
}    

In the rest of this subsection we will prove Theorem \ref{theorem:sp_regret_convex_concave}.
Define $\EL_t(x,y) \triangleq \bar{\EL}_t(x,y) + \frac{1}{\eta}R_X(x) - \frac{1}{\eta}R_Y(y)$. Notice that it is $\frac{1}{\eta}$-strongly convex in $x$ with respect to norm $\Vert \cdot \Vert$ for all $y\in Y$ and $\frac{1}{\eta}$-strongly concave with respect to norm $\Vert \cdot \Vert$ for all $x \in X$. Additionally, notice that $\EL_t$ is $G_\EL\triangleq G_{\bar{\EL}} + \frac{1}{\eta} (G_{R_X}+G_{R_Y})$-Lipschitz with respect to norm $ \Vert \cdot \Vert $. Finally, notice that by nonnegativity of $R_X$ and $R_Y$ for $t=1,...,T$, all $x \in X$ and all $y\in Y$ it holds that
\begin{equation}\label{eq:diff_bar_not_bar}
-\frac{1}{\eta}R_Y(y) \leq \EL_t(x,y) - \bar{\EL}_t(x,y) \leq \frac{1}{\eta} R_X(x).
\end{equation}

The following lemma shows that the value of the convex-concave games defined by $\sum_{t=1}^T \EL_t$ and $\sum_{t=1}^T \bar{\EL}_t$ are not too far from each other.
{
\begin{lemma}\label{lemma:mm_bar_not_bar}
Let $X \subseteq \mathbb{R}^{d_1}$, $Y \subseteq \mathbb{R}^{d_2}$ be convex and compact sets. Let $\{\EL_t\}_{t=1}^T$ be any sequence of convex-concave functions where $\EL_t: X \times Y \rightarrow \mathbb{R}$ for all $t=1,...,T$. Let
\begin{align*}
    \textstyle \bar{x}_{T+1} \in \arg \min_{x\in X} \max_{y\in Y} \sum_{t=1}^T\bar{\EL}_t(x,y), \\
    \textstyle \bar{y}_{T+1} \in \arg \max_{y\in Y} \min_{x\in X}  \sum_{t=1}^T\bar{\EL}_t(x,y).
\end{align*} It holds that 
\begin{align*}
 -\frac{T}{\eta} R_Y(\bar{y}_{T+1}) \leq \MM \sum_{t=1}^T \EL_t (x,y) - \MM \sum_{t=1}^T \bar{\EL}_t(x,y) \leq \frac{T}{\eta} R_X(\bar{x}_{T+1}).
\end{align*}
\end{lemma}
}

%\begin{proof}[Proof of Lemma \ref{lemma:mm_bar_not_bar}]
\proof{Proof of Lemma \ref{lemma:mm_bar_not_bar}.}

We will first show that 
\begin{align*}
\MM \sum_{t=1}^T \EL_t(x,y) - \MM \sum_{t=1}^T \bar{\EL}_t(x,y) \leq \frac{T}{\eta}R_X(\bar{x}_{T+1}).
\end{align*}

Plugging in the definition of $\EL_t$, we have
\begin{align*}
\MM \sum_{t=1}^T \EL_t(x,y) &= \sum_{t=1}^T[\bar{\EL}_t(x_{T+1},y_{T+1}) + \frac{1}{\eta}R_X(x_{T+1}) - \frac{1}{\eta} R_Y(y_{T+1}) ]\\
&\leq \sum_{t=1}^T[\bar{\EL}_t(\bar{x}_{T+1},y_{T+1}) + \frac{1}{\eta}R_X(\bar{x}_{T+1}) - \frac{1}{\eta} R_Y(y_{T+1}) ]\\
&\leq \sum_{t=1}^T[\bar{\EL}_t(\bar{x}_{T+1},\bar{y}_{T+1}) + \frac{1}{\eta}R_X(\bar{x}_{T+1}) - \frac{1}{\eta} R_Y(y_{T+1}) ],
\end{align*}
where the first inequality holds since $(x_{T+1},y_{T+1})$ is the saddle point of $\sum_{t=1}^T \EL_t(x,y)$ and thus $x_{T+1}$ is the minimizer of $\sum_{t=1}^T \EL_t(x,y_{T+1})$ see Equation \eqref{def_sp}. The second inequality holds since $(\bar{x}_{T+1},\bar{y}_{T+1})$ is a saddle point of $\sum_{t=1}^T \bar{\EL}_{t}(x,y)$, and thus $\bar{y}_{T+1}$ is the maximizer of $\sum_{t=1}^T \EL_t(\bar{x}_{T+1},y)$, see Equation \eqref{def_sp}. By definition of $(\bar{x}_{T+1},\bar{y}_{T+1})$, we have  
\begin{align*}
&\sum_{t=1}^T[\bar{\EL}_t(\bar{x}_{T+1},\bar{y}_{T+1}) + \frac{1}{\eta}R_X(\bar{x}_{T+1}) - \frac{1}{\eta} R_Y(y_{T+1}) ]\\
&= \MM  \sum_{t=1}^T[\bar{\EL}_t(x,y) + \frac{T}{\eta}R_X(\bar{x}_{T+1}) - \frac{T}{\eta} R_Y(y_{T+1}) ]\\
& \leq \MM  \sum_{t=1}^T \bar{\EL}_t(x,y) + \frac{T}{\eta}R_X(\bar{x}_{T+1}),
\end{align*}
where the inequality holds by nonnegativity of $R_Y$.

Using a similar argument we now show that 
\begin{align*}
-\frac{T}{\eta}R_Y(\bar{y}_{T+1}) \leq \MM \sum_{t=1}^T \EL_t(x,y) - \MM \sum_{t=1}^T \bar{\EL}_t(x,y).
\end{align*}

Plugging in the definition of $\EL_t$, we have 
\begin{align*}
\MM \sum_{t=1}^T \EL_t(x,y) &= \sum_{t=1}^T[\bar{\EL}_t(x_{T+1},y_{T+1}) + \frac{1}{\eta}R_X(x_{T+1}) - \frac{1}{\eta} R_Y(y_{T+1}) ]\\
& \geq \sum_{t=1}^T[\bar{\EL}_t(x_{T+1},\bar{y}_{T+1}) + \frac{1}{\eta}R_X(x_{T+1}) - \frac{1}{\eta} R_Y(\bar{y}_{T+1}) ] \\
&\geq \sum_{t=1}^T[\bar{\EL}_t(\bar{x}_{T+1},\bar{y}_{T+1}) + \frac{1}{\eta}R_X(x_{T+1}) - \frac{1}{\eta} R_Y(\bar{y}_{T+1}) ],
\end{align*}
where the first inequality holds since $(x_{T+1},y_{T+1})$ is a saddle point of $\sum_{t=1}^T \EL_t(x,y)$ and thus $y_{T+1}$ is a maximizer of $\sum_{t=1}^T \EL_t(x_{T+1},y)$, see Equation \eqref{def_sp}. The second inequality holds since $(\bar{x}_{T+1},\bar{y_{T+1}})$ is a saddle point of $\sum_{t=1}^T \bar{\EL}_t(x,y)$ thus $\bar{x}_{T+1}$ is a minimizer of $\sum_{t=1}^T \bar{\EL}_t(x,\bar{y}_{T+1})$, see Equation \eqref{def_sp}.

By definition of $(\bar{x}_{T+1},\bar{y}_{T+1})$, we have 
\begin{align*}
\sum_{t=1}^T[\bar{\EL}_t(\bar{x}_{T+1},\bar{y}_{T+1}) + \frac{1}{\eta}R_X(x_{T+1}) - \frac{1}{\eta} R_Y(\bar{y}_{T+1}) ] &= \MM  \sum_{t=1}^T[\bar{\EL}_t(x,y) + \frac{T}{\eta}R_X(x_{T+1}) - \frac{T}{\eta} R_Y(\bar{y}_{T+1}) ]\\
& \geq \MM  \sum_{t=1}^T \bar{\EL}_t(x,y) - \frac{T}{\eta}R_Y(\bar{y}_{T+1}),
\end{align*}
where the inequality holds by nonnegativity of $R_Y$. This concludes the proof. 
%\end{proof}
\endproof

To prove the SP-Regret bound, we note that \textsf{SP-RFTL} is running \textsf{SP-FTL} on functions $\{\EL_{t=1}^T\}$ so the proof will be similar to that of Theorem \ref{thm:sp_regret_str}.

\proof{Proof of Theorem \ref{theorem:sp_regret_convex_concave}.}
We first prove one side of the inequality.
\begin{align*}
& \quad \sum_{t=1}^T \bar{\EL}_t(x_t,y_t) - \MM \sum_{t=1}^T \bar{\EL}_t(x,y)\\
& \leq \sum_{t=1}^T \EL_t(x_t,y_t) - \MM \sum_{t=1}^T \bar{\EL}_t(x,y) + \sum_{t=1}^T \frac{1}{\eta} R_Y(y_t) \quad \text{by Equation \eqref{eq:diff_bar_not_bar}}\\
& \leq \sum_{t=1}^T \EL_t(x_t,y_t) - \MM \sum_{t=1}^T \EL_t(x,y) + \sum_{t=1}^T \frac{1}{\eta} R_Y(y_t) + \frac{T}{\eta} R_X(x_{T+1})  \quad \text{by Lemma \ref{lemma:mm_bar_not_bar}}\\
& \leq \sum_{t=1}^T \EL_t(x_t,y_t) - \sum_{t=1}^T \EL_t(x_{t+1},y_{t+1}) + \sum_{t=1}^T \frac{1}{\eta} R_Y(y_t) + \frac{T}{\eta} R_X(x_{T+1}) + G_{\EL} \sum_{t=1}^T \Vert y_t- y_{t+1}\Vert  \quad \text{by Lemma \ref{loss_BTL}}\\
& \leq \sum_{t=1}^T G_{\EL} (\Vert x_t - x_{t+1}\Vert + \Vert y_{t}-y_{t+1}\Vert) + \sum_{t=1}^T \frac{1}{\eta} R_Y(y_t) + \frac{T}{\eta} R_X(x_{T+1}) + G_{\EL} \sum_{t=1}^T \Vert y_t- y_{t+1}\Vert \\
& \quad \text{since $\EL_t$ is $G_\EL$-Lipschitz} \\
%& \leq \sum_{t=1}^T G_{\EL} (\Vert x_t - x_{t+1}\Vert + \Vert y_{t}-y_{t+1}\Vert) + \sum_{t=1}^T \frac{1}{\eta} R_Y(y_t) + \frac{T}{\eta} R_X(x_{T+1}) + G_{\EL} \sum_{t=1}^T \Vert y_t- y_{t+1}\Vert \\
& \leq 2\sum_{t=1}^T G_{\EL} (\Vert x_t - x_{t+1}\Vert + \Vert y_{t}-y_{t+1}\Vert) + \sum_{t=1}^T \frac{1}{\eta} R_Y(y_t) + \frac{T}{\eta} R_X(x_{T+1}).
\end{align*}

Applying Lemma \ref{FTL_close_iterates} using
$H=\frac{1}{\eta}$ and $G_\EL = G_{\bar{\EL}}+ \frac{1}{\eta}\max(G_{R_X}, G_{R_Y})$, we have
\begin{align*}
&\quad 2\sum_{t=1}^T G_{\EL} (\Vert x_t - x_{t+1}\Vert + \Vert y_{t}-y_{t+1}\Vert) + \sum_{t=1}^T \frac{1}{\eta} R_Y(y_t) + \frac{T}{\eta} R_X(x_{T+1})\\
& \leq 8 G_{\EL} \eta [G_{\bar{\EL}}+ \frac{1}{\eta}\max(G_{R_X}, G_{R_Y})] ( 1 + \int_{1}^T \frac{1}{t} dt )+ \sum_{t=1}^T \frac{1}{\eta} R_Y(y_t) + \frac{T}{\eta} R_X(x_{T+1})\\
& \leq 8 G_{\EL}  \eta [G_{\bar{\EL}}+ \frac{1}{\eta}\max(G_{R_X}, G_{R_Y})] ( 1 + \ln(T) )+ \frac{T}{\eta} \max_{y\in Y} R_Y(y) +  \frac{T}{\eta} \max_{x\in X} R_X(x)\\
& \leq 8 \eta [G_{\bar{\EL}}+ \frac{1}{\eta}\max(G_{R_X}, G_{R_Y})]^2 ( 1 + \ln(T) )+ \frac{T}{\eta} \max_{y\in Y} R_Y(y) +  \frac{T}{\eta} \max_{x\in X} R_X(x).
\end{align*}
%Notice that  $\MM \sum_{t=1}^T \bar{\EL}_t(x,y) - \sum_{t=1}^T \bar{\EL}_t(x_t,y_t)$ can be upper bounded by the same quantity using the same argument. 
This completes the proof for one side of the inequality.
We now prove the other side of the inequality.
\begin{align*}
& \quad \MM \sum_{t=1}^T \bar{\EL}_t(x,y) - \sum_{t=1}^T \bar{\EL}_t(x_t,y_t)\\
& \leq \MM \sum_{t=1}^T \bar{\EL}_t(x,y) -  \sum_{t=1}^T \EL_t(x_t,y_t)  + \sum_{t=1}^T \frac{1}{\eta} R_X(x_t) \quad \text{by Equation \eqref{eq:diff_bar_not_bar}}\\
& \leq \MM \sum_{t=1}^T \EL_t(x,y) - \sum_{t=1}^T \EL_t(x_t,y_t)  + \sum_{t=1}^T \frac{1}{\eta} R_X(x_t) + \frac{T}{\eta} R_Y(y_{T+1})  \quad \text{by Lemma \ref{lemma:mm_bar_not_bar}}\\
& \leq \sum_{t=1}^T \EL_t(x_{t+1},y_{t+1}) - \sum_{t=1}^T \EL_t(x_t,y_t)  + \sum_{t=1}^T \frac{1}{\eta} R_X(x_t) + \frac{T}{\eta} R_Y(y_{T+1}) + G_{\EL} \sum_{t=1}^T \Vert x_t- x_{t+1}\Vert  \quad \text{by Lemma \ref{loss_BTL}}\\
& \leq \sum_{t=1}^T G_{\EL} (\Vert x_t - x_{t+1}\Vert + \Vert y_{t}-y_{t+1}\Vert) + \sum_{t=1}^T \frac{1}{\eta} R_X(x_t) + \frac{T}{\eta} R_Y(y_{T+1}) + G_{\EL} \sum_{t=1}^T \Vert x_t- x_{t+1}\Vert \\
& \quad \text{since $\EL_t$ is $G_\EL$-Lipschitz} \\
%& \leq \sum_{t=1}^T G_{\EL} (\Vert x_t - x_{t+1}\Vert + \Vert y_{t}-y_{t+1}\Vert) + \sum_{t=1}^T \frac{1}{\eta} R_Y(y_t) + \frac{T}{\eta} R_X(x_{T+1}) + G_{\EL} \sum_{t=1}^T \Vert y_t- y_{t+1}\Vert \\
& \leq 2\sum_{t=1}^T G_{\EL} (\Vert x_t - x_{t+1}\Vert + \Vert y_{t}-y_{t+1}\Vert) + \sum_{t=1}^T \frac{1}{\eta} R_X(x_t) + \frac{T}{\eta} R_Y(y_{T+1})  \\
& \leq 2\sum_{t=1}^T G_{\EL} ( \frac{4 \eta }{t} [G_{\bar{\EL}}+ \frac{1}{\eta}\max(G_{R_X}, G_{R_Y})]) + \sum_{t=1}^T \frac{1}{\eta} R_X(x_t) + \frac{T}{\eta} R_Y(y_{T+1}).
\end{align*}

Applying Lemma \ref{FTL_close_iterates} using $H=\frac{1}{\eta}$ and $G_\EL = G_{\bar{\EL}}+ \frac{1}{\eta}\max(G_{R_X}, G_{R_Y})$, we have 
\begin{align*}
& \quad 2\sum_{t=1}^T G_{\EL} ( \frac{4 \eta }{t} [G_{\bar{\EL}}+ \frac{1}{\eta}\max(G_{R_X}, G_{R_Y})]) + \sum_{t=1}^T \frac{1}{\eta} R_X(x_t) + \frac{T}{\eta} R_Y(y_{T+1})\\
& \leq 8 G_{\EL} \eta [G_{\bar{\EL}}+ \frac{1}{\eta}\max(G_{R_X}, G_{R_Y})] ( 1 + \int_{1}^T \frac{1}{t} dt )+ \sum_{t=1}^T \frac{1}{\eta} R_X(x_t) + \frac{T}{\eta} R_Y(y_{T+1})\\
& \leq 8 G_{\EL}  \eta [G_{\bar{\EL}}+ \frac{1}{\eta}\max(G_{R_X}, G_{R_Y})] ( 1 + \ln(T) )+ \frac{T}{\eta} \max_{y\in Y} R_Y(y) +  \frac{T}{\eta} \max_{x\in X} R_X(x)\\
& \leq 8 \eta [G_{\bar{\EL}}+ \frac{1}{\eta}\max(G_{R_X}, G_{R_Y})]^2 ( 1 + \ln(T) )+ \frac{T}{\eta} \max_{y\in Y} R_Y(y) +  \frac{T}{\eta} \max_{x\in X} R_X(x).
\end{align*}

This concludes the proof. 

\endproof

\section{Online Matrix Games}
{
In Section~\ref{setup_problem}, we analyzed the OSP problem by treating the payoff functions as general convex-concave functions and the action spaces as general convex compact sets. We explained that, in general, one should expect to achieve SP-Regret which depends \emph{linearly} in the dimension of the problem (see discussion after Corollary~\ref{first_corollary}).

In this section, we consider a special case of the OSP problem with bilinear payoff functions, which we call Online Matrix Games (OMG). In this setting, player 1 has $d_1$ available actions and player 2 has $d_2$ available actions. At each time step $t=1,...,T$, the payoff of the players will be given by a payoff matrix $A_t \in [-1,1]^{d_1\times d_2}$, where the  $(i,j)$-th entry specifies the loss of player 1 and the reward of player 2 when they choose actions $i$ and $j$ respectively. We allow the players to choose probability distributions over their available actions. That is, the decision sets of player 1 and player 2 are the probability simplexes, $\Delta_X\subset \mathbb{R}^{d_1}$ and $\Delta_Y\subset \mathbb{R}^{d_2}$, respectively. Here $\Delta_X$ denotes the probability simplex over $d_1$ actions, that is $\Delta_X \triangleq \{x\in \mathbb{R}^{d_1}: x\geq 0, \Vert x \Vert_1\ = 1\}$, $\Delta_Y$ is defined similarly. Notice that this is a special case of the OSP problem studied in Section~\ref{setup_problem}, where the convex-concave function is defined as $\EL_t(x,y) = x^\top A_t y$, which specifies the expected payoff for the players when they choose distributions $x\in \Delta_X,y\in \Delta_Y$. 
%For simplicity of exposition, in the remainder of the section, we allow the players to `play' probability distributions over actions instead of a single action. If we wanted to enforce choosing only one action, the players could sample an action from their probability distributions.

Our goal in this section is to obtain sharper SP-Regret bounds that scale \emph{logarithmically} in the dimensions $d_1$ and $d_2$. This will allow us to solve games that may have exponentially many actions, which often arise in combinatorial optimization settings. 

To achieve this goal, we exploit the geometry of the probability simplexes $\Delta_X, \Delta_Y$ and the bilinear structure of the payoff functions. 
%From now on we omit the subscripts $X,Y$ in the simplexes.
%The plan to obtain the desired SP-Regret bounds in this more restrictive setting is to 
We use the negative entropy as a regularization function (which is strongly convex with respect to $\Vert \cdot \Vert_1$), that is $R_X(x) = \sum_{i=1}^{d_1} x_i\ln(x_i) + \ln(d_1)$ and $R_Y(y)= \sum_{i=1}^{d_2} y_i \ln(y_i)+\ln(d_2)$ where the extra logarithmic terms ensure $R_X, R_Y$ are nonnegative everywhere in their respective simplexes. Unfortunately, the negative entropy is not Lipschitz over the simplex, so we can not leverage our result from Theorem~\ref{theorem:sp_regret_convex_concave}. To deal with this challenge, we will restrict the new algorithm to play over a restricted simplex:\footnote{We will also use the notation $\Delta_{X,\theta}$ and $\Delta_{Y,\theta}$ to mean the restricted simplex of Player 1 and 2, respectively}
\begin{equation}
\Delta_\theta = \{z \in \mathbb{R}^d: \Vert z\Vert_1=1, z_i\geq \theta, i=1,...,d \}.
\end{equation}
The tuning parameter $\theta \in [0,1/d]$ used for the algorithm will be defined later in the analysis. (Notice that when $\theta > {1}/{d}$, the set is empty). We have the following result.

\begin{lemma}\label{lemma:negative_entropy_lipschitz}
The function $R(x)\triangleq \sum_{i=1}^d x_i \ln(x_i)$ is $G_R$-Lipschitz continuous with respect to $\Vert \cdot \Vert_1$ over $\Delta_\theta$ with $G_R = \max\{|\ln(\theta)|,1\}$.
\end{lemma}

%\begin{proof}[Proof of Lemma \ref{lemma:negative_entropy_lipschitz}]
\proof{Proof of Lemma \ref{lemma:negative_entropy_lipschitz}.}
We need to find $G_R>0$ such that $\Vert \nabla R(x)\Vert_\infty \leq G_R$ for all $x\in \Delta_\theta$. Notice that $[\nabla R(x)]_i = 1 + \ln(x_i)$ for $i=1,...d$. Moreover, since for every $i=1,...,d$ we have $\theta \leq x_i\leq 1$ the following sequence of inequalities hold: $\ln(\theta)\leq 1 +\ln(\theta) \leq 1+ \ln(x_i) \leq 1$. It follows that $G_R = \max\{|\ln(\theta)|,1\}$.
%\end{proof}
\endproof

The  algorithm \textsf{Online-Matrix-Games Regularized-Follow-the-Leader (OMG-RFTL)} is an instantiation of \textsf{SP-RFTL} with a particular choice of regularization functions, which are nonegative and Lipschitz with respect to the $\Vert \cdot \Vert_1$ norm over the sets $\Delta_{X,\theta}$, $\Delta_{Y,\theta}$. With this, we can %use Theorem \ref{theorem:sp_regret_convex_concave} to 
prove a SP-Regret bound for the OMG problem. For the remainder of the section, the regularization functions will be set as follows:
\[
R_X(x) \triangleq  \textstyle \sum_{i=1}^{d_1} x_i \ln(x_i)+\ln(d_1), \quad
R_Y(y)  \triangleq  \textstyle \sum_{i=1}^{d_2} y_i\ln(y_i)+\ln(d_2).
\]
\begin{algorithm}[!tbh]
\caption{Online-Matrix-Games Regularized-Follow-the-Regularized-Leader (\textsf{OMG-RFTL})}
\label{alg:OMG-RFTL}
\begin{algorithmic}
  \STATE {\bfseries input:} $x_1 \in \Delta_{X,\theta}\subset \mathbb{R}^{d_1}$, $y_1 \in \Delta_{Y,\theta}\subset \mathbb{R}^{d_2}$, parameters: $\eta>0$, $\theta<\min \{\frac{1}{d_1}, \frac{1}{d_2}\}$.
  \FOR{$t=1,...T$}
  \STATE \quad Play $(x_t,y_t)$, observe matrix $A_t$
  \STATE \quad $\bar{\EL}_t \gets x^\top A_t y$
  \STATE \quad $\EL_t(x,y) \gets \bar{\EL}_t + \frac{1}{\eta}R_X(x)-\frac{1}{\eta}R_Y(y)$
  \STATE \quad $x_{t+1}\leftarrow \arg \min_{x\in \Delta_{X,\theta}} \max_{y \in \Delta_{Y, \theta}} \sum_{\tau=1}^t \EL_t(x,y)$
  \STATE \quad $y_{t+1}\leftarrow \arg \max_{y \in \Delta_{Y, \theta}} \min_{x\in \Delta_{X,\theta}}  \sum_{\tau=1}^t \EL_t(x,y) $
  \ENDFOR
\end{algorithmic}
\end{algorithm}

We have the following guarantee for \textsf{OMG-RFTL}. 
\begin{theorem}\label{thm:omg_rftl_regret}
Let $\{A_t\}_{t=1}^T$ be an arbitrary sequence of matrices in $[-1, 1]^{d_1 \times d_2}$. Let $G_{\bar{\EL}}$ be the Lipschitz constant (with respect to $\Vert \cdot \Vert_1$) of $\bar{\EL}_t \triangleq x^\top A_t y$ for $t=1,...,T$. Let $\{(x_t,y_t)\}_{t=1}^T$ be the iterates of \textsf{OMG-RFTL} and choose $\theta = e^{-\eta G_{\bar{\EL}}}\leq \min\{\frac{1}{d_1}, \frac{1}{d_2}\}$ such that $\frac{|\ln(\theta)|}{\eta} = G_{\bar{\EL}}$. By setting $\eta = \frac{\sqrt{T}}{G_{\bar{\EL}}}$ in Algorithm~\ref{alg:OMG-RFTL},  it holds that
\begin{align*}
& \quad \left|\sum_{t=1}^T x_t^\top A_t y_t - \MMD \sum_{t=1}^T x^\top A_t y \right|\\
& \leq 32 G_{\bar{\EL}} \sqrt{T} (1 + \ln(T)) + 2 \sqrt{T} \max\{\ln d_1 ,\ln d_2\} + 2 \max\{d_1,d_2\} G_{\bar{\EL}} T e^{-\sqrt{T}}\\
& = O\left(\ln(T)\sqrt{T} +  \sqrt{T} \max\{\ln d_1 ,\ln d_2\}\right) + o(1)\max\{d_1,d_2\}.
\end{align*}
\end{theorem}

To prove the theorem, we require a few intermediate results.
Since Algorithm~\ref{alg:OMG-RFTL} selects actions over a restricted simplex, we must quantify the loss in the SP-Regret bound imposed by this restriction. The next two lemmas make this precise.

\begin{lemma}\label{lemma:dist_proj_sp}
Let $z^* \in \Delta \subset \mathbb{R}^d$ define $z^*_p \triangleq \arg \min_{z\in \Delta_\theta} \Vert z - z^*\Vert_1$, with $\theta \leq \frac{1}{d}$. Notice $z^*_p$ is unique since it is a projection. It holds that $\Vert z^*_p - z^*\Vert_1 \leq 2\theta(d-1)$.
\end{lemma}

%\begin{proof}[Proof of Lemma \ref{lemma:dist_proj_sp}]
\proof{Proof of Lemma \ref{lemma:dist_proj_sp}.}
Choose $z^*=[1;0;0;...;0;0]$, it is easy to see that $z^*_p = [1-\theta (d-1); \theta; \theta; ...; \theta; \theta]$ and $\Vert z^* - z^*_p \Vert_1 = 2\theta(d-1).$ 
 %{\color{red} Formalize this argument?}
%\end{proof}
\endproof

\begin{lemma}\label{lemma:sp_val_error_theta}

Let $\{\bar{\EL}_t(x,y)\}_{t=1}^T$ be an arbitrary sequence of convex-concave functions, $\bar{\EL}_t:\Delta_X \times \Delta_Y \rightarrow \mathbb{R}$, that are $G_{\bar{\EL}}$-Lipschitz with respect to $\Vert \cdot \Vert_1$. 
%With $\Delta_X \subseteq \mathbb{R}^{d_1}$, and $\Delta_Y \subseteq \mathbb{R}^{d_2}$, 
It holds that 
\begin{align*}
- G_{\bar{\EL}} T \Vert x^*_p - x^*\Vert_1 
\leq  \MMD \sum_{t=1}^T \bar{\EL}_t(x,y) - \MMDt \sum_{t=1}^T \bar{\EL}_t(x,y) 
\leq G_{\bar{\EL}}T \Vert y^*_p - y^*\Vert_1.
\end{align*}
\end{lemma}

%\begin{proof}[Proof of Lemma \ref{lemma:sp_val_error_theta}]
\proof{Proof of Lemma \ref{lemma:sp_val_error_theta}.}
Let $(x^*,y^*)$ be any saddle point pair for $\sum_{t=1}^T \bar{\EL}_t(x,y)$ with $x^*\in \Delta_X, y^*\in \Delta_Y$.  Let $(x^*_\theta,y^*_\theta)$ be any saddle point pair for $\sum_{t=1}^T \bar{\EL}_t(x,y)$ with $x^*_\theta \in \Delta_{X,\theta}, y^*_\theta \in \Delta_{Y,\theta}$. Let $x^*_p, y^*_p$ be the projection of $x^*, y^*$ onto $\Delta_{X,\theta}, \Delta_{Y,\theta}$ respectively, using the $\Vert \cdot \Vert_\infty$ norm. We first show the second inequality. 

Since $(x^*,y^*)$ is a saddle point for $\sum_{t=1}^T \bar{\EL}_t (x,y)$ over $\Delta_X$ and $\Delta_Y$, and Player 1 deviated to $x^*_\theta$, we have
\begin{align*}
\sum_{t=1}^T \bar{\EL}_t (x^*, y^*) & \leq \sum_{t=1}^T \bar{\EL}_t (x^*_\theta, y^*)\\
& \leq \sum_{t=1}^T \bar{\EL}_t (x^*_\theta, y^*_p) + G_{\bar{\EL}}T \Vert y^*_p - y^*\Vert_1 \quad \text{since $\bar{\EL}_t$ is $G_{\bar{\EL}}$-Lipschitz}\\
& \leq \sum_{t=1}^T \bar{\EL}_t (x^*_\theta, y^*_\theta) + G_{\bar{\EL}}T \Vert y^*_p - y^*\Vert_1,
\end{align*}
where the last inequality holds since $(x^*_\theta, y^*_\theta)$ is a saddle point for $\sum_{t=1}^T \bar{\EL}_t (x,y)$ over $\Delta_{X,\theta}$ and $\Delta_{Y,\theta}$.

To show the first inequality in the statement of the lemma, by using similar argument, we have
\begin{align*}
\sum_{t=1}^T \bar{\EL}_t (x^*, y^*) & \geq \sum_{t=1}^T \bar{\EL}_t (x^*, y^*_\theta) \\
& \geq \sum_{t=1}^T \bar{\EL}_t (x^*_p, y^*_\theta) - G_{\bar{\EL}} T \Vert x^*_p - x^*\Vert_1\\
& \geq \sum_{t=1}^T \bar{\EL}_t(x^*_\theta, y^*_\theta) - G_{\bar{\EL}} T \Vert x^*_p - x^*\Vert_1.
\end{align*}
This concludes the proof. 
%\end{proof}
\endproof

Combining the previous two lemmas and Theorem \ref{theorem:sp_regret_convex_concave}, we can show the SP-Regret bound for \textsf{OMG-RFTL} holds. We are ready to prove Theorem \ref{thm:omg_rftl_regret}.

%\begin{proof}[Proof of Theorem \ref{thm:omg_rftl_regret}]
\proof{Proof of Theorem \ref{thm:omg_rftl_regret}.}
 For convenience, we define $\bar{\EL}_t(x,y) = x^\top A_t y$. Let $(x^*, y^*)$ be any saddle point of $\MMD \sum_{t=1}^T x^\top A_t y$, and let $(x^*_p, y^*_p)$ be the respective projections onto $\Delta_{X,\theta}, \Delta_{Y,\theta}$ using $\Vert \cdot \Vert_\infty$ norm. 
 %By the choice of $\theta$, we have that $|\ln(\theta)|>1$. 
 %Additionally, notice that $\max_{z\in \Delta_\theta} \sum_{i=1}^d z_i \ln(z_i) + \ln(d) \leq 0 +\ln(d)$ by Jensen's inequality, so we have
Using Lemma~\ref{lemma:dist_proj_sp}, Lemma~\ref{lemma:sp_val_error_theta} and Theorem~\ref{theorem:sp_regret_convex_concave}, we have
\begin{align*}
& \quad \sum_{t=1}^T x_t^\top A_t y_t - \MMD \sum_{t=1}^T x^\top A_t y \\
& \leq \sum_{t=1}^T x_t^\top A_t y_t - \MMDt \sum_{t=1}^T x^\top A_t y + G_{\bar{\EL}} T \Vert x^* - x^*_p\Vert_1  \quad \text{by Lemma \ref{lemma:sp_val_error_theta} }\\
& \leq \sum_{t=1}^T x_t^\top A_t y_t - \MMDt \sum_{t=1}^T x^\top A_t y + 2G_{\bar{\EL}} T \theta (d_1-1)  \quad \text{by Lemma \ref{lemma:dist_proj_sp}}\\
&\leq  8 \eta [G_{\bar{\EL}}+ \frac{1}{\eta}\max(G_{R_X}, G_{R_Y})]^2 ( 1 + \ln(T) )+ \frac{T}{\eta} \max_{y\in \Delta_{Y,\theta}} R_Y(y) +  \frac{T}{\eta} \max_{x\in \Delta_{X\theta}} R_X(x)\\
&\qquad + 2G_{\bar{\EL}} T \theta (d_1-1)  \quad \text{by Theorem \ref{theorem:sp_regret_convex_concave}}.
\end{align*}

By Lemma \ref{lemma:negative_entropy_lipschitz}, we know that $G_{R_X},G_{R_Y} \leq \max\{\vert \ln(\theta),1\vert\}$. Our choice of $\theta$ will ensure that $1\leq \vert \ln(\theta)\vert$, so $\max(G_{R_X},G_{R_Y})\leq \vert \ln(\theta)\vert$. Therefore, we have
\begin{align*}
& 8 \eta [G_{\bar{\EL}}+ \frac{1}{\eta}\max(G_{R_X}, G_{R_Y})]^2 ( 1 + \ln(T) )+ \frac{T}{\eta} \max_{y\in \Delta_{Y,\theta}} R_Y(y) +  \frac{T}{\eta} \max_{x\in \Delta_{X\theta}} R_X(x) + 2G_{\bar{\EL}} T \theta (d_1-1)\\
 &\leq  8 \eta [G_{\bar{\EL}}+ \frac{|\ln(\theta)|}{\eta}]^2 ( 1 + \ln(T) )+ \frac{T}{\eta} \max_{y\in \Delta_{Y,\theta}} R_Y(y) +  \frac{T}{\eta} \max_{x\in \Delta_{X,\theta}} R_X(x) + 2G_{\bar{\EL}} T \theta (d_1-1)\\
 &\leq 32 \eta G_{\bar{\EL}}^2 (1+\ln(T)) + \frac{T}{\eta} \max_{y\in \Delta_{Y,\theta}} R_Y(y) +  \frac{T}{\eta} \max_{x\in \Delta_{X,\theta}} R_X(x) + 2G_{\bar{\EL}} T e^{-\eta G_{\bar{\EL}}} (d_1-1),
 \end{align*}
 where the last inequality holds by the choice of $\theta$.
 
Notice that $\max_{z\in \Delta_\theta} R(z) \triangleq \max_{z\in \Delta_\theta} \sum_{i=1}^d z_i \ln(z_i) + \ln(d) \leq 0 +\ln(d)$. Therefore, we have
 \begin{align*}
  &\quad 32 \eta G_{\bar{\EL}}^2 (1+\ln(T)) + \frac{T}{\eta} \max_{y\in \Delta_{Y,\theta}} R_Y(y) +  \frac{T}{\eta} \max_{x\in \Delta_{X,\theta}} R_X(x) + 2G_{\bar{\EL}} T e^{-\eta G_{\bar{\EL}}} (d_1-1)\\
 & \leq 32 \eta G_{\bar{\EL}}^2 (1+\ln(T)) + \frac{T}{\eta}\ln(d_2) +  \frac{T}{\eta} \ln(d_1)+ 2G_{\bar{\EL}} T e^{-\eta G_{\bar{\EL}}} (d_1-1)\\
 & \leq 32 G_{\bar{\EL}} \sqrt{T} (1 + \ln(T)) + \sqrt{T} (\ln d_1+\ln d_2) + 2 d_1 G_{\bar{\EL}} T e^{-\sqrt{T}} \\
& = O\left(\ln(T)\sqrt{T} +  \sqrt{T} \max\{\ln d_1 ,\ln d_2\}\right) + o(1)\max\{d_1,d_2\}.
\end{align*}
The last line follows because $G_{\bar{\EL}}\leq 1$, since each entry of $A_t$ is bounded between $[-1, 1]$. 

We now prove the other side of the inequality:
\begin{align*}
& \quad \MMD \sum_{t=1}^T x^\top A_t y - \sum_{t=1}^T x_t^\top A_t y_t  \\
& \leq \MMDt \sum_{t=1}^T x^\top A_t y - \sum_{t=1}^T x_t^\top A_t y_t + G_{\bar{\EL}} T \Vert y^* - y^*_p\Vert_1  \quad \text{by Lemma \ref{lemma:sp_val_error_theta} }\\
& \leq \MMDt \sum_{t=1}^T x^\top A_t y -  \sum_{t=1}^T x_t^\top A_t y_t + 2G_{\bar{\EL}} T \theta (d_2-1)  \quad \text{by Lemma \ref{lemma:dist_proj_sp}}\\
&\leq  8 \eta [G_{\bar{\EL}}+ \frac{1}{\eta}\max(G_{R_X}, G_{R_Y})]^2 ( 1 + \ln(T) )+ \frac{T}{\eta} \max_{y\in \Delta_{Y,\theta}} R_Y(y) +  \frac{T}{\eta} \max_{x\in \Delta_{X,\theta}} R_X(x)\\
&\qquad + 2G_{\bar{\EL}} T \theta (d_2-1)  \quad \text{by Theorem \ref{theorem:sp_regret_convex_concave}}\\
 &\leq  8 \eta [G_{\bar{\EL}}+ \frac{|\ln(\theta)|}{\eta}]^2 ( 1 + \ln(T) )+ \frac{T}{\eta} \max_{y\in \Delta_{Y,\theta}} R_Y(y) +  \frac{T}{\eta} \max_{x\in \Delta_{X,\theta}} R_X(x) + 2G_{\bar{\EL}} T \theta (d_2-1)\\
 &\leq 32 \eta G_{\bar{\EL}}^2 (1+\ln(T)) + \frac{T}{\eta} \max_{y\in \Delta_{Y,\theta}} R_Y(y) +  \frac{T}{\eta} \max_{x\in \Delta_{X,\theta}} R_X(x) + 2G_{\bar{\EL}} T e^{-\eta G_{\bar{\EL}}} (d_2-1),
 \end{align*}
 \\
 where the last inequality holds by the choice of $\theta$.
 Again, notice that $\max_{z\in \Delta_\theta} R(z) \triangleq \max_{z\in \Delta_\theta} \sum_{i=1}^d z_i \ln(z_i) + \ln(d) \leq 0 +\ln(d)$. We have
 \begin{align*}
  & 32 \eta G_{\bar{\EL}}^2 (1+\ln(T)) + \frac{T}{\eta} \max_{y\in \Delta_{Y,\theta}} R_Y(y) +  \frac{T}{\eta} \max_{x\in \Delta_{X,\theta}} R_X(x) + 2G_{\bar{\EL}} T e^{-\eta G_{\bar{\EL}}} (d_2-1)\\
 & \leq 32 \eta G_{\bar{\EL}}^2 (1+\ln(T)) + \frac{T}{\eta}\ln(d_2) +  \frac{T}{\eta} \ln(d_1)+ 2G_{\bar{\EL}} T e^{-\eta G_{\bar{\EL}}} (d_2-1)\\
 & \leq 32 G_{\bar{\EL}} \sqrt{T} (1 + \ln(T)) + \sqrt{T} (\ln d_1+\ln d_2) + 2 d_2 G_{\bar{\EL}} T e^{-\sqrt{T}} \\
& = O\left(\ln(T)\sqrt{T} +  \sqrt{T} \max\{\ln d_1 ,\ln d_2\}\right) + o(1)\max\{d_1,d_2\}.
\end{align*}
The last line follows because $G_{\bar{\EL}}\leq 1$, since each entry of $A_t$ is bounded between $[-1, 1]$.
This concludes the proof. 
%\end{proof}
\endproof
}

\subsection{Online Matrix Games with Bandit Feedback}\label{section:omg_bandit}
{ 
The results we proved for the OMG problem can be extended to a setting with bandit feedback. In the bandit setting, the players observe in every round only the payoff corresponding to the chosen actions. In other words, if Player 1 chooses action $i$, Player 2 chooses action $j$, and the payoff matrix at that time step is $A_t$, then the players observe only $(A_t)_{ij}$ instead of the full matrix $A_t$. The limited feedback makes the problem significantly more challenging than the full information one, as the players must balance the exploration-exploitation tradeoff.
%find a way to \textit{exploit} (i.e., try to achieve Nash equilibrium) and \textit{explore} (i.e., try to estimate unobserved entries of $A_t$). 
This problem resembles that of Online Bandit Optimization \cite{flaxman2005online,auer1995gambling,bubeck2016kernel,hazan2016optimal}, albeit with two players.
%while the challenge in our setting is that we must estimate the whole matrix $A_t$ with \emph{one} function evaluation.
%instead of the gradients $\nabla_x \EL_t(x,y)$ and $\nabla_y \EL_t(x,y)$ where $\EL_t = x^\top A_t y$.

For convenience, we define some useful notation. 
For $i=1,...,d$, let $e_i \in \mathbb{R}^d$  be the collection of standard unit vectors i.e. $e_i$ is the vector that has a $1$ in the $i$-th entry and $0$ in the rest. Let $e_{x,t}$ be the standard unit vector corresponding to the decision made by Player 1 for round $t$, define $e_{y,t}$ similarly. Notice that under bandit feedback, in round $t$ both players only observe the quantity $e_{x,t}^{\top} A_t e_{y,t}$.

\subsubsection{One-Point Estimate for Payoff Function}
As explained previously, in each round $t$ the players must estimate $A_t$ by observing only one of its entries. To this end, we allow the players to share with each other their decisions and to randomize \textit{jointly} (a similar assumption is used to define correlated equilibria in zero-sum games, see \cite{aumann1987correlated}). The following result shows how to build a random estimate of $A$ by observing only one of its entries.
 
\begin{theorem}\label{thm:hess_estimate}
Let $x\in \Delta_{X,\delta}, y\in \Delta_{Y,\delta}$ with $d_1,d_2\geq2$ and $\delta >0$. Sample $i' \sim x, j'\sim y$. Let $\hat{A}$ be the $d_1\times d_2$ matrix with $\hat{A}_{i,j}=0$ for all $i,j$ such that $i\neq i'$ and $j\neq j'$ and $\hat{A}_{i',j'} = \frac{A_{i',j'}}{x_{i'}y_{j'}}$. It holds that 
\begin{equation*} 
\mathbb{E}_{i' \sim x, j' \sim y} [ \hat{A}] = A.
\end{equation*}
\end{theorem}

\proof{Proof of Theorem \ref{thm:hess_estimate}.}
Let $B_{i,j}$ be the matrix of zeros everywhere except in the $i,j$ entry where it is equal to $ \frac{A_{i,j}}{x(i)y(j)}$. We have
\begin{align*}
\mathbb{E}_{i' \sim x, j' \sim y} [ \hat{A}] = \sum_{i=1}^{d_1} \sum_{j=1}^{d_2} x_i y_j B_{i,j} = A. 
\end{align*}

\endproof

\subsubsection{Algorithm under Bandit Feedback}
We now present an algorithm that ensures sublinear (i.e. $o(T)$) SP-Regret under bandit feedback for the OMG problem that holds against an adaptive adversary. By adaptive adversary, we mean that the payoff matrices $A_t$ can depend on the players' actions up to time $t-1$; in particular, we assume the adversary does not observe the actions chosen by the players for time period $t$ when choosing $A_t$. We consider an algorithm that runs \textsf{OMG-RFTL} on a sequence of functions $\hat{\EL}_t \triangleq x^\top \hat{A}_t y$, where $\hat{A}_t$ is the unbiased one-point estimate of $A_t$ derived in Theorem~\ref{thm:hess_estimate}. Recall that the iterates of \textsf{OMG-RFTL} algorithm are distributions over the possible actions of both players. In order to generate the estimate $\hat{A}_t$, both players will sample an action from their distributions and weigh their observation with the inverse probability of obtaining that observation. 

\begin{algorithm}[tbh]
\caption{Bandit Online-Matrix-Games Regularized-Follow-the-Leader (\textsf{Bandit-OMG-RFTL})}
\label{alg: BOMGFTRL}
\begin{algorithmic}
  \STATE  {\bfseries input:} $x_1 \in \Delta_{X,\delta}\subset \mathbb{R}^{d_1}$, $y_1 \in \Delta_{Y,\delta}\subset \mathbb{R}^{d_2}$, parameters: $\eta>0$, $0<\delta<\min \{\frac{1}{d_1}, \frac{1}{d_2}\}$.
  \FOR{$t=1,...T$}
  \STATE \quad Sample independently $e_{x,t} \sim x_t$ and $e_{y,t} \sim y_t$
  \STATE \quad Observe $e_{x,t}^{\top} A_t e_{y,t}$
  \STATE \quad  Build $\hat{A_t}$ as in Theorem \ref{thm:hess_estimate} using  $e_{x,t}^{\top} A_t e_{y,t}, x_t, y_t$
    \STATE \quad $\hat{\EL}_t \gets x^\top \hat{A}_t y$
    \STATE \quad $\EL_t(x,y) \gets \hat{\EL}_t + \frac{1}{\eta}R_X(x)-\frac{1}{\eta}R_Y(y)$
  \STATE \quad $x_{t+1}\leftarrow \arg \min_{x\in \Delta_{X,\theta}} \max_{y \in \Delta_{Y, \theta}} \sum_{\tau=1}^t \EL_t(x,y)$
  \STATE \quad $y_{t+1}\leftarrow \arg \max_{y \in \Delta_{Y, \theta}} \min_{x\in \Delta_{X,\theta}}  \sum_{\tau=1}^t \EL_t(x,y) $
  \ENDFOR
\end{algorithmic}
\end{algorithm}

We have the following guarantee for \textsf{Bandit-OMG-RFTL}. 
\begin{theorem}\label{no_bandit_regret}
Let $\{A_t\}_{t=1}^T$ be any sequence of payoff matrices chosen by an adaptive adversary, where $A_t\in [-1,1]^{d_1 \times d_2}$ for all $t=1,...,T$. Let $\{e_{x,t},e_{y,t}\}_{t=1}^T$ be the iterates generated by \textsf{Bandit-OMG-FTRL}. Setting $\delta = \frac{1}{T^{1/6}}$, $\eta = T^{1/6}$ ensures
\begin{align*} 
\left| \mathbb{E}  \left[\sum_{t=1}^T e_{x,t}^{\top}A_t e_{y,t} - \MM  \sum_{t=1}^T x^{\top}A_t y \right] \right|  \leq O((d_1 + d_2) \ln(T) T^{5/6})
 \end{align*}
where the expectation is taken with respect to all the randomization used in the algorithm.
\end{theorem}

%A complete proof of the theorem can be found in the Appendix. 
The full proof of this Theorem will be given shortly. We now present a few lemmas. 
The total payoff given to each of the players is given by $\sum_{t=1}^T e_{x,t}^\top A_t e_{y,t}$ so we must relate this quantity to the iterates $\{x_t,y_t\}_{t=1}^T$ of \textsf{OMG-RFTL} when run on sequence of matrices $\{\hat{A}_t \}_{t=1}^T$. The following two lemmas will allow us to do so. 

\begin{lemma}\label{e_to_x}
Let $\{e_{x,t}, e_{y,t}\}_{t=1}^T$ be the sequence of iterates generated by \textsf{Bandit-OMG-RFTL}. It holds that
\begin{align*} \textstyle
 \mathbb{E}\left[  \sum_{t=1}^T e_{x,t}^{\top}A_t e_{y,t}\right] =  \mathbb{E}\left[ \sum_{t=1}^T x^{\top}_t A_t y_t\right],
 \end{align*}
 where the expectation is taken with respect to the internal randomness of the algorithm.
 \end{lemma}
 
%\begin{proof}[Proof of Lemma \ref{e_to_x}]
\proof{Proof of Lemma \ref{e_to_x}.} Let $\mathbb{E}[X | \tau=1,...,T-1]$ be the expectation of random variable $X$ conditioned on all the randomness from time steps $\tau=1,...,T-1$.
\begin{align*}
&\quad \mathbb{E}[\sum_{t=1}^T e_{x,t}^{\top} A_t e_{y,t}]\\
& = \mathbb{E}[\sum_{t=1}^{T-1} e_{x,t}^{\top} A_t e_{y,t}] + \mathbb{E}[e_{x,T}^{\top} A_T e_{y,T}] \\
& = \mathbb{E}[\sum_{t=1}^{T-1} e_{x,t}^{\top} A_t e_{y,t}] + \mathbb{E}[ \mathbb{E}_{e_{x,T} \sim x_t, e_{y,T} \sim y_t}[e_{x,T}^{\top} A_T e_{y,T}|\tau=1,...,T-1]].
\end{align*}

Since the adversary can can not observe $e_{x,T}, e_{y,T}$ when selecting $A_T$, $e_{x,T}, e_{y,T}$ and $A_T$ are all independent from each other, thus it holds that $\mathbb{E}[ \mathbb{E}_{e_{x,T} \sim x_t, e_{y,T} \sim y_t}[e_{x,T}^{\top} A_T e_{y,T}|\tau=1,...,T-1]] = \mathbb{E}[ x_T^{\top} \mathbb{E}_{e_{x,T} \sim x_t, e_{y,T} \sim y_t}[ A_T |\tau=1,...,T-1]y_T ]$. Therefore

\begin{align*}
&\quad \mathbb{E}[\sum_{t=1}^{T-1} e_{x,t}^{\top} A_t e_{y,t}] + \mathbb{E}[ \mathbb{E}_{e_{x,T} \sim x_t, e_{y,T} \sim y_t}[e_{x,T}^{\top} A_T e_{y,T}|\tau=1,...,T-1]] \\
 & = \mathbb{E}[\sum_{t=1}^{T-1} e_{x,t}^{\top} A_t e_{y,t}] + \mathbb{E}[ x_T^{\top} \mathbb{E}_{e_{x,T} \sim x_t, e_{y,T} \sim y_t}[ A_T |\tau=1,...,T-1]y_T ]\\
 & = \mathbb{E}[\sum_{t=1}^{T-1} x_t^{\top} A_t y_t] + \mathbb{E}[ x_T^{\top} A_T y_T ] .
\end{align*}
Repeating the argument $T-1$ more times yields the result. 
%\end{proof}
\endproof

\begin{lemma}\label{A_hat_no_hat}
Let $\{A_t\}_{t=1}^T$ be any sequence of payoff matrices chosen by an adaptive adversary, where $A_t\in \mathbb{R}^{d_1 \times d_2}$ for all $t=1,...,T$. Let $\{x_t,y_t,\hat{A}_t\}_{t=1}^T$ be generated by \textsf{Bandit-OMG-FTRL}. It holds that
\begin{align*} \textstyle
\mathbb{E}\left[\sum_{t=1}^T x_t^{\top} \hat{A}_t y_t\right] = \mathbb{E}\left[\sum_{t=1}^T x_t^{\top} A_t y_t\right],
\end{align*}
where the expectation is with respect to all the internal randomness of the algorithm.
\end{lemma}

%\begin{proof}[Proof of Lemma \ref{A_hat_no_hat}]
\proof{Proof of Lemma \ref{A_hat_no_hat}.}
We have
\begin{align*}
& \quad \mathbb{E}[\sum_{t=1}^T x_t^{\top} \hat{A}_t y_t]\\
& = \mathbb{E}[\sum_{t=1}^{T-1} x_t^{\top} \hat{A}_t y_t] + \mathbb{E}[x_T^{\top} \hat{A}_T y_T]\\
& = \mathbb{E}[\sum_{t=1}^{T-1} x_t^{\top} \hat{A}_t y_t] + \mathbb{E}[ \mathbb{E}[x_T^{\top} \hat{A}_T y_T|\tau=1,...,T-1]].
\end{align*}
Since $(x_T,y_T)$ is deterministic conditioned on everything that has happened up to time $T-1$ it holds that $\mathbb{E}[ \mathbb{E}[x_T^{\top} \hat{A}_T y_T|\tau=1,...,T-1]] = \mathbb{E}[ x_T^{\top} \mathbb{E}[ \hat{A}_T |\tau=1,...,T-1]y_T ]$. It then follows that 
\begin{align*}
& \quad \mathbb{E}[\sum_{t=1}^{T-1} x_t^{\top} \hat{A}_t y_t] + \mathbb{E}[ \mathbb{E}[x_T^{\top} \hat{A}_T y_T|\tau=1,...,T-1]]\\
& = \mathbb{E}[\sum_{t=1}^{T-1} x_t^{\top} \hat{A}_t y_t] + \mathbb{E}[ x_T^{\top} \mathbb{E}[ \hat{A}_T |\tau=1,...,T-1]y_T ]\\
& = \mathbb{E}[\sum_{t=1}^{T-1} x_t^{\top} \hat{A}_t y_t] + \mathbb{E}[ x_T^{\top} A_T y_T ],
\end{align*}
where the last equality holds by Theorem \ref{thm:hess_estimate}. Repeating the argument $T-1$ more times yields the result. 
%\end{proof}
\endproof

We will then bound the difference between the comparator term 
 $\MMD \sum_{t=1}^T x^{\top} A_t y$ and the comparator term Theorem \ref{thm:omg_rftl_regret} gives us by running \textsf{OMG-RFTL} on functions $\{\hat{\EL}\}_{t=1}^T$ and sets $\Delta_{X,\delta}, \Delta_{Y,\delta}$, $\MMDd \sum_{t=1}^T x^{\top} \hat{A}_t y$. Special care must be taken to ensure this difference holds even against an adaptive adversary. To this end, we use the next two lemmas; as we will see, the proof of Lemma \ref{lemma_with_alphas} relies heavily on Theorem \ref{thm:hess_estimate}.

\begin{lemma}\label{close_sp_vals}
With probability 1, for any $y \in \Delta_{Y,\delta}$, it holds that
\begin{align*}
\left| \min_{x\in \Delta_{X,\delta}}\max_{y\in \Delta_{Y,\delta}} \sum_{t=1}^T x^{\top}A_t y - \min_{x\in \Delta_{X,\delta}}\max_{y\in \Delta_{Y,\delta}} \sum_{t=1}^T x^{\top}\hat{A}_t y \right|
\leq  \max_{y\in \Delta_{Y,\delta}} \left \Vert \sum_{t=1}^T A_t y - \hat{A}_t y \right \Vert_2.
\end{align*}
\end{lemma}

%\begin{proof}[Proof of Lemma \ref{close_sp_vals}]
\proof{Proof of Lemma \ref{close_sp_vals}.}
Let us fist bound $|\sum_{t=1}^T x^\top A_t y - \sum_{t=1}^T x^\top \hat{A}_t y|$  for any $x \in \Delta_X$ and $y \in \Delta_Y$ with probability 1. For any $x \in \Delta_X$ and $y \in \Delta_Y$ we have
\begin{align*}
|\sum_{t=1}^T x^\top A_t y - \sum_{t=1}^T x^\top \hat{A}_t y| 
& = | x^{\top} ( \sum_{t=1}^T A_t y - \sum_{t=1}^T \hat{A}_t y ) |\\
& \leq \Vert x\Vert_2 \Vert \sum_{t=1}^T A_t y - \hat{A}_t y \Vert_2 \quad \text{by Cauchy-Schwarz}\\
& \leq \Vert \sum_{t=1}^T A_t y - \hat{A}_t y \Vert_2 \quad \text{since $x\in \Delta_X$}.
\end{align*}

This implies that
\begin{align*}
& \sum_{t=1}^T x^{\top} \hat{A}_t y \leq \sum_{t=1}^T x^{\top}A_t y + \Vert \sum_{t=1}^T A_t y - \hat{A}_t y \Vert_2 \quad \text{$ \forall x \in \Delta_X$, $ \forall y \in \Delta_Y$},
\end{align*}
which implies that
\begin{align*}
& \min_{x\in \Delta_{X,\delta}} \sum_{t=1}^T x^{\top} \hat{A}_t y \leq \sum_{t=1}^T x^{\top}A_t y + \Vert \sum_{t=1}^T A_t y - \hat{A}_t y \Vert_2 \quad \forall x \in \Delta_{X,\delta} ,y \in \Delta_{Y,\delta}.
\end{align*}
Therefore, it holds that
\begin{align*}
&\min_{x\in \Delta_{X,\delta}} \sum_{t=1}^T x^{\top} \hat{A}_t y \leq \max_{y\in \Delta_{Y,\delta}} \sum_{t=1}^T x^{\top}A_t y + \Vert \sum_{t=1}^T A_t y - \hat{A}_t y \Vert_2 \quad \forall x \in \Delta_{X,\delta},y \in \Delta_{Y,\delta},
\end{align*}
Thus
\begin{align*}
& \max_{y\in \Delta_{Y,\delta}} \min_{x\in \Delta_{X,\delta}} \sum_{t=1}^T x^{\top} \hat{A}_t y \leq \min_{x\in \Delta_{X,\delta}} \max_{y\in \Delta_{Y,\delta}} \sum_{t=1}^T x^{\top}A_t y + \max_{y\in \Delta_{Y,\delta}}\Vert \sum_{t=1}^T A_t y - \hat{A}_t y \Vert_2  . 
\end{align*}

Since $\max_{y\in \Delta_{Y,\delta}} \min_{x\in \Delta_{X,\delta}} \sum_{t=1}^T x^{\top} \hat{A}_t y = \min_{x\in \Delta_{X,\delta}} \max_{y\in \Delta_{Y,\delta}} \sum_{t=1}^T x^{\top} \hat{A}_t y$ (the function is convex-concave and the sets $ \Delta_Y^\delta$ and $\Delta_X^\delta$ are convex and compact), we have shown that
\begin{align*}
\min_{x\in \Delta_{X,\delta}} \max_{y\in \Delta_{Y,\delta}}  \sum_{t=1}^T x^{\top} \hat{A}_t y \leq \min_{x\in \Delta_{X,\delta}} \max_{y\in \Delta_{Y,\delta}} \sum_{t=1}^T x^{\top}A_t y + \max_{y\in \Delta_{Y,\delta}} \Vert \sum_{t=1}^T A_t y - \hat{A}_t y \Vert_2  \quad \forall y \in \Delta_{Y,\delta}.
\end{align*}
The other side of the inequality follows from a similar argument. Indeed we know that
\begin{align*}
& \sum_{t=1}^T x^{\top} A_t y \leq \sum_{t=1}^T x^{\top} \hat{A}_t y + \Vert \sum_{t=1}^T A_t y - \hat{A}_t y \Vert_2 \quad \text{$ \forall x \in \Delta_X$, $ \forall y \in \Delta_Y$}.
\end{align*}
The previous inequality implies that
\begin{align*}
 & \min_{x\in \Delta_{X,\delta}} \sum_{t=1}^T x^{\top} A_t y \leq \sum_{t=1}^T x^{\top}\hat{A}_t y + \Vert \sum_{t=1}^T A_t y - \hat{A}_t y \Vert_2 \quad \forall x \in \Delta_{X,\delta} ,y \in \Delta_{Y,\delta},
 \end{align*}
or
\begin{align*}
& \min_{x\in \Delta_{X,\delta}} \sum_{t=1}^T x^{\top} A_t y \leq \max_{y\in \Delta_{Y,\delta}} \sum_{t=1}^T x^{\top} \hat{A}_t y + \Vert \sum_{t=1}^T A_t y - \hat{A}_t y \Vert_2 \quad \forall x \in \Delta_{X,\delta},y \in \Delta_{Y,\delta}.
\end{align*}
Therefore, we have
\begin{align*}
& \max_{y\in \Delta_{Y,\delta}} \min_{x\in \Delta_{X,\delta}} \sum_{t=1}^T x^{\top} A_t y \leq \min_{x\in \Delta_{X,\delta}} \max_{y\in \Delta_{Y,\delta}} \sum_{t=1}^T x^{\top} \hat{A}_t y + \max_{y\in \Delta_{Y,\delta}} \Vert \sum_{t=1}^T A_t y - \hat{A}_t y \Vert_2 . 
\end{align*}

Since $\max_{y\in \Delta_{Y,\delta}} \min_{x\in \Delta_{X,\delta}} \sum_{t=1}^T x^{\top} A_t y = \min_{x\in \Delta_{X,\delta}} \max_{y\in \Delta_{Y,\delta}} \sum_{t=1}^T x^{\top} A_t y$ we get the result. 
%\end{proof}
\endproof

\begin{lemma}\label{lemma_with_alphas}
Let $\{A_t\}$ be any sequence of payoff matrices chosen by an adaptive adversary, where with $A_t\in \mathbb{R}^{d_1 \times d_2}$ for all $t=1,...,T$. Let $\{\hat{A}_t\}$ be the sequence of matrices generated by \textsc{Bandit-OMG-RFTL}. For any $y\in \Delta_{Y}$, it holds that
\begin{align*}
\mathbb{E}\left[ \left \Vert \sum_{t=1}^T A_t y - \hat{A}_t y \right \Vert_2 \right] \leq \frac{2 \sqrt{T} \min (d_1,d_2)}{\delta^2},
\end{align*}
where the expectation is taken with respect to the internal randomness of the algorithm.
\end{lemma}

%\begin{proof}[Proof of Lemma \ref{lemma_with_alphas}]
\proof{Proof of Lemma \ref{lemma_with_alphas}.}
For any $y$ define $\alpha_t \triangleq A_t y - \hat{A}_t y$.
We first show that for all $t, t'$ such that $t<t'$ it holds that $\mathbb{E}[\alpha_t^{\top}\alpha_{t'}]=0$. Indeed
\begin{align*}
\mathbb{E}[\alpha_t^{\top} \alpha_{t'}] &= \mathbb{E}[(A_t y -\hat{A}_t y )^{\top}(A_{t'} y -\hat{A}_{t'} y )]\\
&= \mathbb{E}[(A_t y)^{\top}A_{t'} y - (A_t y)^{\top}\hat{A}_{t'} y - (\hat{A}_t y )^{\top}A_{t'} y + (\hat{A}_t y )^{\top} \hat{A}_{t'} y ]\\
& = (A_t y)^{\top}A_{t'} y  - (A_t y)^{\top}A_{t'} y  - (A_t y )^{\top}A_{t'} y + \mathbb{E}[(\hat{A}_t y )^{\top} \hat{A}_{t'} y ]\\
& = (A_t y)^{\top}A_{t'} y  - (A_t y)^{\top}A_{t'} y  - (A_t y )^{\top}A_{t'} y + (A_t y )^{\top} A_{t'} y\\
&= 0,
\end{align*}
where the second to last line follows since 
\begin{align*}
\mathbb{E}[(\hat{A}_t y )^{\top} \hat{A}_{t'} y ] &= \mathbb{E}_{1,...,t'-1}[ \mathbb{E}[(\hat{A}_t y )^{\top} \hat{A}_{t'} y |\tau = 1,..., t'-1]]\\
&= \mathbb{E}_{1,...,t'-1}[ (\hat{A}_t y )^{\top} \mathbb{E}[ \hat{A}_{t'} y |\tau = 1,..., t'-1]]\\
&= \mathbb{E}_{1,...,t'-1}[ (\hat{A}_t y )^{\top} A_{t'}y ]\\
&= (A_t y )^{\top} A_{t'}y.
\end{align*}

Now, we have
\begin{align*}
\mathbb{E}[ \Vert \sum_{t=1}^T A_t y - \hat{A}_t y \Vert_2 ] & = \sqrt{\mathbb{E}[\Vert \sum_{t=1}^T \alpha_t \Vert_2 ]^2}\\
& \leq \sqrt{ \mathbb{E}[\Vert \sum_{t=1}^T \alpha_t \Vert_2^2 ]} \qquad \text{by Jensen's Inequality}\\
& = \sqrt{ \sum_{t=1}^T \mathbb{E}[\Vert \alpha_t\Vert_2^2] + 2 \sum_{t < t'} \mathbb{E}[\alpha_t^{\top}\alpha_{t'}]}\\
& = \sqrt{\sum_{t=1}^T \mathbb{E}[\Vert A_t y - \hat{A}_t y\Vert_2^2]}\\
& \leq \sqrt{ \sum_{t=1}^T \mathbb{E}[2 \Vert A_t y\Vert ^2 + 2\Vert \hat{A}_t y\Vert_2^2 ] }.
\end{align*}
We proceed to bound $\Vert \hat{A}_t y\Vert_2$, the upper bound we obtain will also bound $\Vert A_t y\Vert $ because of the following fact. If the random vector $\tilde{a}$ satisfies $\Vert \tilde{a}\Vert \leq c$ for some constant c with probability 1 then $\Vert \mathbb{E}\tilde{a}\Vert \leq c$. Indeed by Jensen's inequality, we have  $\Vert \mathbb{E}\tilde{a}\Vert \leq \mathbb{E} \Vert \tilde{a}\Vert \leq c$. Let us omit the subscript $t$ for the rest of the proof. Let $\hat{A}_{[i,:]}$ be the $i$-th row of matrix $\hat{A}$.

\begin{align*}
\Vert \hat{A}y\Vert_2 &= \sqrt{\sum_{i=1}^{d_1} \big[\sum_{j=1}^{d_2} \hat{a}_{i,j} y_j\big]^2}\\
&\leq \sum_{i=1}^{d_1} \sqrt{\big[\sum_{j=1}^{d_2} \hat{a}_{i,j} y_j\big]^2}\\
&= \sum_{i=1}^{d_1} \big| \sum_{j=1}^{d_2} \hat{a}_{i,j} y_j \big| \\
& \leq \sum_{i=1}^{d_1} \Vert \hat{A}_{[i,:]} \Vert_{\infty} \Vert y\Vert_1 \quad \text{by generalized Cauchy Schwarz}\\
& \leq d_1 \max_{i,j} |\frac{A_{i,j}}{\delta^2}| \quad \text{by definition of $\hat{A}$ and using the fact that $ x_t \in \Delta_{X,\delta}$ and $y_t \in \Delta_{Y,\delta}$}\\
& \leq  \frac{d_1}{\delta^2}.
\end{align*}
Notice the upper bound $\frac{d_2}{\delta^2}$ can also be obtained by interchanging the summations and repeating the argument. This yields the desired result.
%\end{proof}
\endproof

Before we prove Theorem~\ref{no_bandit_regret}, we need the following lemma.

\begin{lemma}\label{bilinear_lipschitz}
Consider a matrix $A\in \mathbb{R}^{d_1\times d_2}$. If the absolute value of each entry of $A$ is bounded by $c>0$, then the function $\EL(x,y) = x^{\top}A y$ is $ G_{\EL}^{\Vert \cdot \Vert_2}$-Lipschitz continuous with respect to $\Vert \cdot \Vert_2$ over the sets $\Delta_X\subset \mathbb{R}^{d_1}$ and $\Delta_Y\subset \mathbb{R}^{d_2}$ , where $ G_{\EL}^{\Vert \cdot \Vert_2}  = \sqrt{c}\left(\sqrt{d_1}+\sqrt{d_2}\right)$. The function $\EL$ is also $G_{\EL}^{\Vert \cdot \Vert_1}$-Lipschitz continuous (over the same sets) with respect to norm $\Vert \cdot \Vert_1$, where $G_{\EL}^{\Vert \cdot \Vert_1} = c$.
\end{lemma}

%\begin{proof}[Proof of Lemma \ref{bilinear_lipschitz}]
\proof{Proof of Lemma \ref{bilinear_lipschitz}.}

%We omit the subscript $t$.
\begin{align*}
\Vert \nabla x^\top Ay \Vert_2 &= 
\left\Vert
\begin{bmatrix}
\nabla_x x^\top Ay;
\nabla_y x^\top Ay
\end{bmatrix}^\top \right\Vert_2 \\
&=
\left\Vert \begin{bmatrix}
A_{[1,:]}^\top y;
 ... ;
 A_{[d_1,:]}^\top y;
 A_{[:,1]}^\top x  ;
 ... ;
 A_{[:,d_2]}^\top x
\end{bmatrix}^\top \right\Vert_2 \\
& \leq 
\left\Vert \begin{bmatrix}
A_{[1,:]}^\top y  ;
 ... ;
 A_{[d_1,:]}^\top y\\
\end{bmatrix}^\top \right\Vert_2 
+
\left\Vert \begin{bmatrix}
 A_{[:,1]}^\top x  ;
 ... ;
 A_{[:,d_2]}^\top x
\end{bmatrix}^\top \right\Vert_2 \\
&\leq
\sqrt{\sum_{i=1}^{d_1}(A_{[i,:]}^\top y)^2} 
+
 \left\Vert \begin{bmatrix}
 A_{[:,1]}^\top x  ;
 ... ;
 A_{[:,d_2]}^\top x
\end{bmatrix}^\top \right\Vert_2 \\
&\leq
\sqrt{d_1(\Vert A_{[i,:]}\Vert_{\infty} \Vert y\Vert_1)^2} 
+
\left\Vert \begin{bmatrix}
 A_{[:,1]}^\top x  ;
 ... ;
 A_{[:,d_2]}^\top x
\end{bmatrix}^\top \right\Vert_2 \quad \text{by Generalized Cauchy Schwarz}\\
&\leq
\sqrt{c d_1} 
+
 \left\Vert \begin{bmatrix}
 A_{[:,1]}^\top x  ;
 ... ;
 A_{[:,d_2]}^\top x
\end{bmatrix}^\top \right\Vert_2 \\
&\leq
\sqrt{c d_1} 
+ 
\sqrt{c d_2} \quad \text{by using the same reasoning}.
\end{align*}

 We now prove the second part of the claim by bounding $\Vert \nabla x^\top Ay \Vert_\infty$.
 \begin{align*}
\Vert \nabla x^\top Ay \Vert_\infty &= 
\left\Vert
\begin{bmatrix}
\nabla_x x^\top Ay  ;
\nabla_y x^\top Ay
\end{bmatrix}^\top \right\Vert_\infty \\
&=
\left\Vert \begin{bmatrix}
A_{[1,:]}^\top y  ;
 ... ;
 A_{[d_1,:]}^\top y;
 A_{[:,1]}^\top x  ;
 ... ;
 A_{[:,d_2]}^\top x
\end{bmatrix}^\top \right\Vert_\infty
\end{align*}

By Cauchy-Schwarz inequality, for any  $i=1,...,d_1$, we have $A_{[i,:]}^\top y \leq \Vert A_{[i,:]} \Vert_\infty \Vert y\Vert_1 \leq c \Vert y\Vert_1 \leq c$, since $y \in \Delta_Y$. Similarly, for any $j=1,...,d_2$ $A_{[:,j]}^\top x \leq c$. This shows that  $\Vert \nabla x^\top Ay \Vert_\infty\leq c$.  
%\end{proof}
\endproof

 The proof of Theorem~\ref{no_bandit_regret} follows by combining Lemmas \ref{e_to_x} through \ref{lemma_with_alphas}, with careful choice of tuning parameters.
 
% \begin{proof}[Proof of Theorem \ref{no_bandit_regret}]
 \proof{Proof of Theorem \ref{no_bandit_regret}.}
We first focus on one side of the inequality, 
\begin{align*}
& \quad \mathbb{E}[  \sum_{t=1}^T e_{x,t}^{\top}A_t e_{y,t} - \MMD  \sum_{t=1}^T x^{\top}A_t y ] \\
& =  \mathbb{E}[  \sum_{t=1}^T e_{x,t}^{\top}A_t e_{y,t}] - \mathbb{E}[\MMD  \sum_{t=1}^T x^{\top}A_t y ]\\
& = \mathbb{E}[  \sum_{t=1}^T x_t^{\top}A_t y_t] - \mathbb{E}[\MMD  \sum_{t=1}^T x^{\top}A_t y ] \quad \text{by Lemma \ref{e_to_x} }\\
& = \mathbb{E}[  \sum_{t=1}^T x_t^{\top}A_t y_t] - \mathbb{E}[ \min_{x\in \Delta_{X,\delta}}\max_{y\in \Delta_{Y,\delta}} \sum_{t=1}^T x^{\top}A_t y ] + 2\delta G_{\bar{\EL}}^{\Vert \cdot \Vert_1}(d_1-1) T  \quad {\text{by Lemmas \ref{lemma:dist_proj_sp} and \ref{lemma:sp_val_error_theta} }}\\
& \leq \mathbb{E}[  \sum_{t=1}^T x_t^{\top}A_t y_t] - \mathbb{E}[ \min_{x\in \Delta_{X,\delta}}\max_{y\in \Delta_{Y,\delta}} \sum_{t=1}^T x^{\top}\hat{A}_t y ] + \frac{2 \sqrt{T}\min(d_1,d_2)}{\delta^2}  + 2\delta G_{\bar{\EL}}^{\Vert \cdot \Vert_1}(d_1-1) T\\
& \qquad \text{by Lemmas \ref{close_sp_vals} and \ref{lemma_with_alphas}}\\
& \leq \mathbb{E}[  \sum_{t=1}^T x_t^{\top}\hat{A}_t y_t] - \mathbb{E}[ \min_{x\in \Delta_{X,\delta}}\max_{y\in \Delta_{Y,\delta}} \sum_{t=1}^T x^{\top}\hat{A}_t y ] + \frac{2 \sqrt{T}\min(d_1,d_2)}{\delta^2}   + 2\delta G_{\bar{\EL}}^{\Vert \cdot \Vert_1}(d_1-1) T\\
& \qquad \text{by Lemma \ref{A_hat_no_hat}}\\
& \leq  8 \eta [G_{\hat{\EL}}^{\Vert \cdot \Vert_1} + \frac{\vert \ln(\delta)\vert}{\eta}]^2 (1+\ln(T)) +\frac{T}{\eta} (\ln(d_1)+ \ln(d_2)) + \frac{2 \sqrt{T}\min(d_1,d_2) }{\delta^2} + 2\delta G_{\bar{\EL}}^{\Vert \cdot \Vert_1}(d_1-1) T.
\end{align*}

The last inequality follows by the same reasoning we used in the proof of Theorem \ref{thm:omg_rftl_regret}. Indeed $\{(x_t,y_t)\}_{t=1}^T$ are the iterates of \textsf{OMG-RFTL} run on the sequence of payoff functions $\{x^\top A_t y\}_{t=1}^T$ so the same proof holds. By Lemma \ref{bilinear_lipschitz}, since the absolute value of all the entries in $A_t$ is bounded above by 1, it holds that
\begin{align*}
& 8 \eta [G_{\hat{\EL}}^{\Vert \cdot \Vert_1} + \frac{\vert \ln(\delta)\vert}{\eta}]^2 (1+\ln(T)) +\frac{T}{\eta} (\ln(d_1)+ \ln(d_2)) + \frac{2 \sqrt{T}\min(d_1,d_2) }{\delta^2} + 2\delta G_{\bar{\EL}}^{\Vert \cdot \Vert_1}(d_1-1) T\\
& =  8 \eta [\frac{1}{\delta^2}+ \frac{\vert \ln(\delta)\vert}{\eta}]^2 (1+\ln(T)) +\frac{T}{\eta} (\ln(d_1)+ \ln(d_2)) + \frac{2 \sqrt{T} \min(d_1,d_2) }{\delta^2} + 2\delta (d_1-1) T\\
& = O((d_1 + d_2) \ln(T) T^{5/6}),
\end{align*}
where the last equality holds since we use $\delta = \frac{1}{T^{1/6}}$, $\eta = T^{1/6}$.

We now show the other side of the inequality.
\begin{align*}
& \quad \mathbb{E}[ \MMD  \sum_{t=1}^T x^{\top}A_t y - \sum_{t=1}^T e_{x,t}^{\top}A_t e_{y,t} ] \\
& =  \mathbb{E}[\MMD  \sum_{t=1}^T x^{\top}A_t y ]  - \mathbb{E}[  \sum_{t=1}^T e_{x,t}^{\top}A_t e_{y,t}] \\
& = \mathbb{E}[\MMD  \sum_{t=1}^T x^{\top}A_t y ]  -  \mathbb{E}[  \sum_{t=1}^T x_t^{\top}A_t y_t]\quad \text{by Lemma \ref{e_to_x} }\\
& = \mathbb{E}[ \min_{x\in \Delta_{X,\delta}}\max_{y\in \Delta_{Y,\delta}} \sum_{t=1}^T x^{\top}A_t y ]  - \mathbb{E}[  \sum_{t=1}^T x_t^{\top}A_t y_t] + 2\delta G_{\bar{\EL}}^{\Vert \cdot \Vert_1}(d_2-1) T  \quad {\text{by Lemmas \ref{lemma:dist_proj_sp} and \ref{lemma:sp_val_error_theta} }}\\
& \leq \mathbb{E}[ \min_{x\in \Delta_{X,\delta}}\max_{y\in \Delta_{Y,\delta}} \sum_{t=1}^T x^{\top}\hat{A}_t y ]  - \mathbb{E}[  \sum_{t=1}^T x_t^{\top}A_t y_t] + \frac{2 \sqrt{T}\min(d_1,d_2)}{\delta^2}  + 2\delta G_{\bar{\EL}}^{\Vert \cdot \Vert_1}(d_2-1) T\\
& \qquad \text{by Lemmas \ref{close_sp_vals} and \ref{lemma_with_alphas}}\\
& \leq \mathbb{E}[ \min_{x\in \Delta_{X,\delta}}\max_{y\in \Delta_{Y,\delta}} \sum_{t=1}^T x^{\top}\hat{A}_t y ]  - \mathbb{E}[  \sum_{t=1}^T x_t^{\top}\hat{A}_t y_t] + \frac{2 \sqrt{T}\min(d_1,d_2)}{\delta^2}   + 2\delta G_{\bar{\EL}}^{\Vert \cdot \Vert_1}(d_2-1) T\\
& \qquad \text{by Lemma \ref{A_hat_no_hat}}\\
& \leq  8 \eta [G_{\hat{\EL}}^{\Vert \cdot \Vert_1} + \frac{\vert \ln(\delta)\vert}{\eta}]^2 (1+\ln(T)) +\frac{T}{\eta} (\ln(d_1)+ \ln(d_2)) + \frac{2 \sqrt{T}\min(d_1,d_2) }{\delta^2} + 2\delta G_{\bar{\EL}}^{\Vert \cdot \Vert_1}(d_2-1) T.
\end{align*}

The last inequality follows by the same reasoning we used in the proof of Theorem \ref{thm:omg_rftl_regret}. Indeed $\{(x_t,y_t)\}_{t=1}^T$ are the iterates of \textsf{OMG-RFTL} run on the sequence of payoff functions $\{x^\top A_t y\}_{t=1}^T$ so the same proof holds. By Lemma \ref{bilinear_lipschitz}, since the absolute value of all the entries in $A_t$ is bounded above by 1, it holds that
\begin{align*}
& 8 \eta [G_{\hat{\EL}}^{\Vert \cdot \Vert_1} + \frac{\vert \ln(\delta)\vert}{\eta}]^2 (1+\ln(T)) +\frac{T}{\eta} (\ln(d_1)+ \ln(d_2)) + \frac{2 \sqrt{T}\min(d_1,d_2) }{\delta^2} + 2\delta G_{\bar{\EL}}^{\Vert \cdot \Vert_1}(d_2-1) T\\
& =  8 \eta [\frac{1}{\delta^2}+ \frac{\vert \ln(\delta)\vert}{\eta}]^2 (1+\ln(T)) +\frac{T}{\eta} (\ln(d_1)+ \ln(d_2)) + \frac{2 \sqrt{T} \min(d_1,d_2) }{\delta^2} + 2\delta (d_2-1) T \\
& = O((d_1 + d_2) \ln(T) T^{5/6}).
\end{align*}
The last equality holds since we use $\delta = \frac{1}{T^{1/6}}$, $\eta = T^{1/6}$.
This completes the proof. 
\endproof

}

\section{Relationship between SP-regret and Individual-regret}
\label{choose_one}

We have defined two regret metrics for the OSP problem, namely the \emph{SP-regret} \eqref{eq:sp-reg} and the \emph{individual-regret} \eqref{eq:indiv-reg}. In the previous subsection, we proposed an algorithm (\textsf{SP-FTL}) with sublinear SP-regret.
We have also mentioned that any OCO algorithm (e.g., online gradient descent, online mirror descent, Follow-the-Leader) can achieve sublinear individual-regret.
A natural question is whether there exists a \emph{single} algorithm that has both sublinear SP-regret and individual-regret. Surprisingly, the answer is negative.

{

%\begin{theorem}\label{thm:choose_one_omg}
\begin{theorem}\label{thm:impossible}
Consider any algorithm that selects a sequence of $x_t, y_t$ pairs given the past payoff matrices $A_1, \ldots, A_{t-1}$. Consider the following three objectives:
\begin{eqnarray}
 \label{eq:xyregret} \left\vert \sum_{t=1}^Tx_t^\top A_t y_t - \MMD  \sum_{t=1}^Tx^\top A_t y \right \vert & = & o(T), \\
  \label{eq:xregret} \sum_{t=1}^Tx_t^\top A_t y_t - \min_{x\in \Delta_X} \sum_{t=1}^T x^\top A_t y_t  & = & o(T), \\
  \label{eq:yregret} \max_{y\in \Delta_Y} \sum_{t=1}^T x_t^\top A_t y - \sum_{t=1}^Tx_t^\top A_t y_t  & = & o(T).
\end{eqnarray}
Then there exists an (adversarially-chosen) sequence $A_1, A_2, \ldots$ such that not all of \eqref{eq:xyregret}, \eqref{eq:xregret}, and \eqref{eq:yregret}, are true.
\end{theorem} 
}

A formal proof of the result is shown shortly, but here we give a sketch. The main idea is to construct two parallel scenarios, each with their own sequences of payoff matrices. The two scenarios will be identical for the first $T/2$ periods but are different for the rest of the horizon. In our particular construction, in both scenarios the players play the well known ``matching-pennies'' game for the first $T/2$ periods, then in first scenario they play a game with equal payoffs for all of their actions and in the second scenario they play a game where Player 1 is indifferent between its actions. One can show that if all three quantities in the statement of the theorem are $o(T)$ in the first scenario, then we prove that at least one of them is $\Omega(T)$ in the second one which yields the result. This suggests that the machinery for OCO, which minimizes individual regret, cannot be directly applied to the OMG problem.

 We note that despite the negative result in Theorem~\ref{thm:impossible}, it is possible to achieve both sublinear SP-Regret and individual-regret with further assumptions on the payoff functions $\{\EL_t\}_{t=1}^T$.  

One such example is where $\EL_t(x,y)$ is sampled i.i.d.\ ; this case is discussed in \S\ref{OCOwK_setup}. However, in light of Theorem~\ref{thm:impossible}, in the general case where $\{\EL_t\}_{t=1}^T$ is an arbitrary sequence, the best one can hope for is achieve either SP-Regret or individual-regret, but not both. In \S\ref{subsec:regret-simulation}, we include a numerical example to further illustrate the relationship between SP-Regret and individual-regret.

\subsection{Proof of the Impossibility Result}
We now present a formal proof of the impossibility result.

{

\proof{Proof of Theorem~\ref{thm:impossible}.}
We assume there exists an algorithm such that 
\begin{align*}
\vert \sum_{t=1}^Tx_t^\top A_t y_t - \min_{x\in \Delta_X}\max_{y \in \Delta_Y}  \sum_{t=1}^Tx^\top A_t y\vert &\leq o(T),\\
\sum_{t=1}^Tx_t^\top A_t y_t - \min_{x\in \Delta_X} \sum_{t=1}^T x^\top A_t y_t &\leq o(T), \\
\max_{y\in \Delta_Y} \sum_{t=1}^T x_t^\top A_T y - \sum_{t=1}^Tx_t^\top A_t y_t &\leq o(T),
\end{align*}
for all possible sequences of matrices $\{A_t\}_{t=1}^T$ with bounded entries between $[-1,1]$. We now construct two sequences of functions for which all the three guarantees hold and lead that to a contradiction. Let $T$ be divisible by $2$. 

In scenario 1: $A_t = 
\begin{bmatrix}
1 &-1 \\
-1 &1\\
\end{bmatrix}$
for $1\leq t \leq \frac{T}{2}$ and $A_t=
\begin{bmatrix}
0 &0 \\
0 &0\\
\end{bmatrix}$
for $\frac{T}{2}<t\leq T$.

In scenario 2: $A_t = 
\begin{bmatrix}
1 &-1 \\
-1 &1\\
\end{bmatrix}$
for $1\leq t \leq \frac{T}{2}$ and $A_t=
\begin{bmatrix}
1 &-1 \\
1 &-1\\
\end{bmatrix}$
for $\frac{T}{2}<t\leq T$.

 It is easy to see that for both scenarios it holds that $\MM \sum_{t=1}^T x^\top A_t y = 0$.
Since $d_1=d_2=2$ and we can parametrize any $x\in \Delta_X$ as $x = [\alpha; 1-\alpha]$ and any $y\in \Delta_Y$ as $y=[\beta;1-\beta]$ for some $0\leq \alpha, \beta \leq 1$. By assumption, we have 
\[
\max_{y\in \Delta_Y} \sum_{t=1}^T x_t^\top A_t y - \MM \sum_{t=1}^T x^\top A_t y \leq o(T)\]
for all sequences of matrices $\{A_t\}_{t=1}^T$. This implies for scenario 1 that 
\[
\max_{0\leq \beta \leq 1} \sum_{t=1}^{\frac{T}{2}}4\alpha_t\beta - 2\beta + 1 -2\alpha_t \leq o(T),\]
which also implies that $\sum_{t=1}^{\frac{T}{2}}2\alpha_t -1 \leq o(T)$ and $\sum_{t=1}^{\frac{T}{2}} 1-2\alpha_t \leq o(T)$ since $\sum_{t=1}^{\frac{T}{2}}4\alpha_t\beta - 2\beta + 1 -2\alpha_t$ is a linear function of $\beta$ and thus its maximum occurs at $\beta=0$ or $\beta=1$.

For scenario 2 $\max_{y\in \Delta_Y} \sum_{t=1}^T x_t^\top A_t y - \MM \sum_{t=1}^T x^\top A_t y \leq o(T)$ reduces to
\[
\max_{0\leq \beta \leq 1} \sum_{t=1}^{\frac{T}{2}} 4\alpha_t \beta - 2\beta + 1 - 2\alpha_t + \frac{T}{2} (2\beta-1)\leq o(T),
\]
which implies $\sum_{t=1}^{\frac{T}{2}} 2\alpha_t -1 + \frac{T}{2} \leq o(T)$ and $\sum_{t=1}^{\frac{T}{2}} 1 -2\alpha_t  + \frac{T}{2} \leq o(T)$. Finally, notice that $\sum_{t=1}^{\frac{T}{2}} 2\alpha_t -1 + \frac{T}{2} \leq o(T)$ implies $\frac{T}{2} \leq o(T) +\sum_{t=1}^{\frac{T}{2}} 1 - 2\alpha_t$. But from scenario 1, we have  $\sum_{t=1}^{\frac{T}{2}} 1 - 2\alpha_t \leq o(T)$ since $\frac{T}{2}\leq o(T)$ is a contradiction we get the result.
 
\endproof

}

\section{Online Convex Optimization with Knapsacks}\label{OCOwK_setup}
 
In this section, we consider the online convex optimization with knapsacks (OCOwK) problem. This problem is motivated by various applications in dynamic pricing, online ad auctions, and crowdsourcing (see \cite{badanidiyuru2018bandits} and discussion in \S\ref{Intro}). The OCOwK model generalizes the standard OCO framework by having an additional set of resource constraints.
We will show that OCOwK is closely related to the OSP problem studied in \S\ref{setup_problem}.

In the OCOwk problem, the decision maker has a set of resources $i=1,\ldots,m$ with given budgets $b=[b_1;b_2;...;b_m]$.
There are $T$ time periods. At each time period, the decision maker chooses $x_{t}\in X\subset\mathbb{R}^{n}$, where $X$ is a convex compact set. 
{After the decision is chosen, Nature reveals two functions: a concave reward function
$r_{t}: X\to\mathbb{R}_+$, which is assumed to be $G$-Lipschitz (with respect to $\Vert \cdot \Vert_2$),
and a vector-valued resource consumption function $c_{t}: X\to\mathbb{R}^{m}_+$, where each entry of $c_t$ is a convex, $G$-Lipschitz function (with respect to $\Vert \cdot \Vert_2$). }

The objective is to maximize cumulative reward while satisfying the budget constraints. In particular, we assume that if a decision $x_t$ violates any of the budget constraints, no reward is collected at period $t$. Therefore, the decision maker's cumulative reward is given by
\begin{equation}\label{eq:ocowk-reward}
R(x_{1},x_{2},\cdots,x_{T})=\sum_{t=1}^{T} \left( r_{t}(x_{t})\mathbb{I}\bigl[\sum_{\tau =1}^{t}c_{\tau}(x_{\tau})\leq b \bigr] \right),
\end{equation}
where $\mathbb{I}[\cdot]$ denotes the indicator function. In \eqref{eq:ocowk-reward}, if $b = +\infty$, the problem reduces to the standard OCO setting. In the special case where $r_t$ and $c_t$ are linear functions, our problem is related to the Bandits with Knapsack (BwK) model studied in \cite{badanidiyuru2018bandits}. A similar problem with general concave reward and convex constraints is studied by \cite{agrawal2014bandits}. However, unlike our model, both of these papers assume bandit feedback.

In order to guarantee that the budget constraint can always be satisfied, we assume there exists a ``null action'' that doesn't consume any resource or generate any reward. 
\begin{assumption}\label{assumption:null}
There exists an action $x_0 \in X$ such that $r_t(x_0)\equiv  0$ and $c_t(x_0) \equiv 0$ for all $t=1,\ldots,T$. 
\end{assumption}
The ``null action'' assumption is often satisfied in real-world applications of OCOwK.
For example, in dynamic pricing, the ``null action'' is equivalent to charging an extremely high price so there is no customer demand; in online auctions (see \cite{balseiro2017learning}), the ``null action'' corresponds to bidding at \$0.

If the reward and consumption functions are chosen arbitrarily, it can be shown that no algorithm can achieve sublinear regret for OCOwK. Intuitively, if the reward and consumption functions shift at $\lfloor T/2 \rfloor$, no algorithm can recover the mistake made before  $T/2$ in the remaining periods (which is similar to the case in \S\ref{choose_one}). Therefore, we consider the setting where the reward and consumption functions are stochastic.

\begin{assumption}\label{assumption:iid}
For $t=1,\ldots,T$, the reward function $r_t$ and consumption function $c_t$ are sampled i.i.d.\ from a fixed and unknown joint distribution.
\end{assumption}

Notice that even when the reward and consumption distribution is known, the optimal policy for the OCOwK problem is not a static decision, as the optimal decision depends on the remaining time and remaining budget. Therefore, defining the offline benchmark for OCOwK is not as straightforward as in the stochastic OCO setting.
However, it has been shown in the literature that the following offline convex problem provides an \emph{upper bound} of the expected reward of the optimal offline policy under Assumption~\ref{assumption:iid} (see e.g.~\cite{badanidiyuru2018bandits,besbes2012blind}):
\begin{align}
r^{*}\triangleq \max_{x\in X} &\left\{ \sum_{t=1}^{T}\Ex[r_t(x)],\; \text{subject to}  \sum_{t=1}^{T}\Ex[c_{t}(x)]\leq b \right\}.
\label{eq:hindsight}
\end{align}
Therefore, we define the expected regret for the OCOwK problem as 
\begin{align*}
\COCOreg (T) \triangleq r^* - \Ex [ R(x_{1},x_{2},\cdots,x_{T})],
\end{align*}
where the expectation is taken with respect to the random realizations of functions $r_t$ and $c_t$.

\subsection{Reduction to a Saddle Point Problem}

We relate the OCOwK problem to the OSP problem studied in \S\ref{setup_problem} by defining the function
\begin{equation}\label{eq:def_L_t}
L_{t}(x,y)\triangleq -r_{t}(x)-y^{\top}(b/T-c_{t}(x)),
\end{equation}
with $y \in \mathbb{R}^m_+$.
Note that $L_t(x,y)$ is convex in $x$ and concave in $y$, so we can treat $L_{t}(x,y)$ as a payoff function in the OSP problem. 
Here, $y$ can be viewed as the dual prices associated with the budget constraints in \eqref{eq:hindsight}, and the function $L_t(x,y)$ penalizes the payoff if consumption at iteration $t$ exceeds the average budget per period.

We let constant $y_{max,i}$ be the maximum reward that can be gained by adding one unit of resource $i$ ($\forall i \in [m]$), and define set $Y = \prod_{i=1}^m [0,y_{max,i}]$. Namely, $y_{max,i}$ is an upper bound on the dual variables for problem \eqref{eq:hindsight}. {We also define vector $y_{max} = [y_{max,1}; \cdots; y_{max,m} ]$.}
For any sequence of decisions $x_1,\cdots, x_T$, 
we claim that the decision maker's total reward is bounded by
\begin{equation}\label{eq:reward}
R(x_{1},x_{2},\cdots,x_{T})\geq \sum_{t=1}^{T}r_{t}(x_{t}) + \min_{y \in Y}\left\{ y^{\top}\sum_{t=1}^{T}(b/T-c_{t}(x_{t}))\right\} =- \max_{y \in Y}\sum_{t=1}^{T}L_{t}(x_{t},y).
\end{equation}
To see this, consider a modified OCOwK problem where resource consumption is allowed to go over the budget, but the decision maker must pay $y_{max,i}$ for each additional unit of resource $i$ used over $b_i$. By the definition of $y_{max,i}$, the decision maker's profit under the modified problem is given by the right-hand side of \eqref{eq:reward}, which can be no more than the reward in the original problem.

We now consider the benchmark \eqref{eq:hindsight}. By Assumption~\ref{assumption:null}, the Slater condition holds for the convex optimization problem \eqref{eq:hindsight}; so by using strong
duality, we have 
\[
r^{*}= - \min_{x\in X}\max_{y \in Y}\left\{ -\sum_{t=1}^{T}\Ex[r_{t}(x)]-y^{\top}\sum_{t=1}^{T}(b/T-\Ex[c_{t}(x)])\right\} = - \min_{x\in X}\max_{y \in Y} \Ex[\sum_{t=1}^{T}L_{t}(x,y)].
\]
Therefore, the expected regret for OCOwK is bounded by
\begin{align}
& \COCOreg (T) = r^* - \Ex [ R(x_{1},x_{2},\cdots,x_{T})] \nonumber \\
 \leq  & \Ex\Bigl[\max_{y \in Y}\sum_{t=1}^{T}L_{t}(x_{t},y)\Bigr] -  \min_{x\in X}\max_{y \in Y} \Ex\Bigl[\sum_{t=1}^{T}L_{t}(x,y)\Bigr] \nonumber\\
= & \underbrace{\Ex\Bigl[\max_{y\in Y}\sum_{t=1}^{T}L_{t}(x_{t},y)-\sum_{t=1}^{T}L_{t}(x_{t},y_{t})\Bigr]}_{(\dagger)}+\underbrace{\Ex\Bigl[\sum_{t=1}^{T}L_{t}(x_{t},y_{t})\Bigr]-\min_{x\in X}\max_{y\in Y}\Ex\Bigl[\sum_{t=1}^{T}L_{t}(x,y)\Bigr]}_{(\ddagger)}. \label{eq:regret-bound}
\end{align}
We have bounded the regret of the OCOwK problem by two quantities in a related OSP problem. In particular, the term $(\dagger)$ is equal to the expectation of player 2's individual-regret (see Eq~\eqref{eq:indiv-reg-y}), and the second term is related to the SP-Regret.

\subsection{Algorithms for OCOwK}
\label{subsec:algorithms-for-ocowk}

Motivated by its connection to the OSP problem, we propose two algorithms for the OCOwK problem. For clarity we defer all proofs to the next two subsections.

First, we consider \textsf{SP-FTL} defined in Algorithm~\ref{alg: SP-FTL}.
In view of Eq~\eqref{eq:regret-bound}, we can bound the regret for OCOwK by the sum of an individual-regret and the SP-Regret.
Theorem~\ref{theorem:sp_regret_convex_concave} has already provided a SP-Regret bound for \textsf{SP-FTL}, so we just need to prove a sublinear individual-regret bound. In general, this is impossible due to the negative result in Theorem~\ref{thm:impossible}. However, since we made the additional assumption that $r_t$ and $c_t$ are sampled i.i.d., we are able to get a sublinear individual-regret for \textsf{SP-FTL} in the OCOwK problem.

We start by establishing a high probability bound on the individual-regret for the general OSP problem when the payoff function $\EL_t(x,y)$ is strongly convex-concave.

\begin{lemma}\label{no_sp_indiv_reg_str}
Let $\{\EL_t(x,y)\}_{t=1}^T$ be an i.i.d. sequence of functions that is $H$-strongly convex-concave with respect norm $\Vert \cdot \Vert_2$, and $G$-Lipschitz continuous with respect norm $\Vert \cdot \Vert_2$. Here $\EL_t:X\times Y \rightarrow \mathbb{R}$. Let $d$ be the dimension of $X \times Y$, and $D_{XY}>0$ be some constant such that $\max_{z_1,z_2\in X\times Y}\Vert z_1 - z_2\Vert_2\leq D_{XY}$. \textsf{SP-FTL} run on the sequence of functions $\{\EL_t\}_{t=1}^T$ guarantees that with probability at least $1-1/T$
\begin{align*}
& \max_{y\in Y}\sum_{t=1}^T \EL_t(x_t,y) -\sum_{t=1}^T\EL_t(x_t,y_t) \leq \frac{8G^2}{H}(1+\ln(T)) + O\big( \frac{G^{3/2} D_{XY}^{1/2} (d \ln(T) \ln(dT) )^{1/4} T^{3/4}\ln^{1/4}(T)}{H^{1/2}}\big).
\end{align*}
\end{lemma}

The proof of Lemma~\ref{no_sp_indiv_reg_str} uses a concentration inequality for Lipschitz functions by Shalev-Shwartz et al.\ \cite{shalevstochastic}. The key step in the proof is to show that the solution of the sample average approximation at step $t$ i.e. $x_t$ is close to $x^*$, the saddle point of the expected game.

However, we cannot directly use Lemma~\ref{no_sp_indiv_reg_str} to bound the individual-regret term $(\dagger)$ in \eqref{eq:regret-bound} because the function $L_t(x,y)$ is linear in $y$ and thus not strongly convex-concave. We add a regularization term to $L_t(x,y)$ to make it $H$-strongly convex-concave. Notice our choice of the regularization term here is not the same as in Theorem~\ref{theorem:sp_regret_convex_concave}, which leads to a $\tilde{O}(T^{5/6})$ bound in the following theorem.

{
\begin{theorem}\label{no_sp_indiv_reg}
Let $X\subset \mathbb{R}^n$ be a convex compact set. Let $\{r_t(x)\}_{t=1}^T$ be an i.i.d. sequence of concave reward functions with $r_t:X\rightarrow \mathbb{R}_+$ which is $G$-Lipschitz with respect to norm $\Vert \cdot \Vert_2$. Let $\{c_t(x)\}_{t=1}^T$ be an i.i.d. sequence of vector-valued functions with $c_t:X \rightarrow \mathbb{R}^m$ where each entry of $c_t$ is convex in $x$ and $G$-Lipschitz with respect to norm $\Vert \cdot \Vert_2$. Let $L_t$ be defined as in equation (\ref{eq:def_L_t}). Define $\bar{\EL}_t(x,y) \triangleq L_t(x,y) + H\Vert x \Vert^2- H\Vert y \Vert^2$, where $H \triangleq T^{-1/6}$. Applying the \textsf{SP-FTL} algorithm on functions $\{\bar{\EL}_t\}_{t=1}^T$
guarantees that with probability at least  $1-1/T$ it holds that
\begin{align*}\label{eq:high-probability-bound}
% & \quad\max_{y\in Y} \sum_{t=1}^T L_t(x_t,y) - \sum_{t=1}^T L_t(x_t,y_t)\\
% &\leq \frac{8(G+2H(D_X+D_Y))^2}{H}(1+\ln(T)) +\\
% &\quad O\big( \frac{\beta(G,H,D_X,D_Y) D_{XY}^{1/2} (d \ln(T) \ln(dT) )^{1/4} T^{3/4}\ln^{1/4}(T)}{H^{1/2}}\big) + TH(D_X^2 + D_Y^2).
& \quad\max_{y\in Y} \sum_{t=1}^T L_t(x_t,y) - \sum_{t=1}^T L_t(x_t,y_t)\\
& \leq \frac{8(G+2H(D_X+D_Y))^2}{H}(1+\ln(T)) \\
&\quad + O\big( \frac{\beta(G,H,D_X,D_Y) D_{XY}^{1/2} (d \ln(T) \ln(dT) )^{1/4} T^{3/4}\ln^{1/4}(T)}{H^{1/2}}\big) + TH(D_X^2 + D_Y^2)\\
& = O\left(poly(G,D_X, D_Y) \ln^{1/2}(T) T^{5/6} \right),
\end{align*}
%  Where the function $\beta$ is defined in the proof of the theorem. 
 where $\beta(G,H,D_X,D_Y)$ is defined in the proof of the theorem. Additionally, for the OCOwK problem, with probability at least  $1-1/T$ it holds that

 \begin{align*}
 \COCOreg(T) = O\left(poly(G,D_X, D_Y) \ln^{1/2}(T) T^{5/6} \right).
 \end{align*}

\end{theorem}
}

Next, we present an algorithm for OCOwK that improves the  regret bound in Theorem~\ref{no_sp_indiv_reg}. 
The key idea of this algorithm is to update primal variable $x$ and dual variable $y$ of $L_t(x,y)$ in parallel. Each variable can be updated using any algorithm for Online Convex Optimization such as Online Gradient Descent \cite{zinkevich2003online} or Regularized Follow the Leader \cite{hazan2007logarithmic,hazan2016introduction}.
We call this algorithm Primal-Dual Regularized-Follow-the-Leader (\textsf{PD-RFTL}) (see Algorithm~\ref{alg: PD-RFTL}). 

\begin{algorithm}[tbh]
\caption{Primal-Dual Regularized Follow-the-Leader (\textsf{PD-RFTL})}
\label{alg: PD-RFTL}
\begin{algorithmic}
{
  \STATE {\bfseries input:} Convex compact decision sets $X$,$Y=\Pi_{i=1}^m[0,y_{\max,i}]$. Parameters $\eta_1,\eta_2$.
    \STATE $x_1 \leftarrow \arg \min_{x\in X} \frac{1}{2}\Vert x\Vert_2^2$ 
   \STATE $y_1 \leftarrow \arg \max_{y\in Y} -\frac{1}{2}\Vert y\Vert_2^2 $
  \FOR{$t=1,...,T$}
  \STATE \quad Play $(x_t,y_t)$
  \STATE \quad Observe $L_t$; define $f_{t}(x)\triangleq L_{t}(x,y_{t})$ and $g_{t}(y)\triangleq L_{t}(x_{t},y)$
  \STATE \quad  $x_{t+1} \leftarrow \arg\min_{x\in X}  \left\{\sum_{\tau=1}^{t} \nabla f_\tau (x_\tau)^\top x +\frac{1}{2\eta_1}\Vert x \Vert_2^2\right\}$
  \STATE \quad $y_{t+1} \leftarrow \arg\max_{y \in Y} \left\{\sum_{\tau=1}^{t} \nabla g_\tau (y_\tau)^\top y-\frac{1}{2\eta_2} \Vert y \Vert_2^2\right\}$
  \ENDFOR
  }
\end{algorithmic}
\end{algorithm}

We will bound the regret of \textsf{PD-RFTL} using Eq~\eqref{eq:regret-bound}. Before we proceed we state the well known individual-regret guarantee of \textsf{RFTL}.
\begin{lemma}[Adapted from \cite{hazan2016introduction} Ch. 5]\label{thm:rftl_bound}
Let $\{f_t(x)\}_{t=1}^T$ be any sequence of convex and $G_f$-Lipschitz functions where $f_t:X\rightarrow \mathbb{R}$ and $X\subset\mathbb{R}^d$ is a convex compact set such that $\max_{x_1,x_2}\Vert x_1-x_2\Vert_2\leq D_X$. The \textsf{RFTL} algorithm: $x_1 \leftarrow \arg\min_{x\in X}\frac{1}{2}\Vert x \Vert_2^2$, $x_t\leftarrow \arg \min_{x\in X} \sum_{\tau=1}^t \nabla f_t(x_t)^\top x + \frac{1}{2\eta}\Vert x \Vert_2^2$ guarantees that
\begin{align*}
    \sum_{t=1}^T f_t(x_t) - \min_{x\in X} \sum_{t=1}^T f_t(x)\ \leq 2\eta G_f^2 T + \frac{D_X^2}{\eta}.
\end{align*}
\end{lemma}
Recall that for \textsf{SP-FTL}, it was more challenging to bound the first term $(\dagger)$ and relatively easy to bound the term $(\ddagger)$. For \textsf{PD-RFTL} it is quite the opposite. By defining $g_{t}(y)\triangleq L_{t}(x_{t},y)$, the first term  $(\dagger)$ can be  written as $\Ex[\max_{y\in Y} \sum_{t=1}^T g_t(y) - \sum_{t=1}^T g_t(y_t)]$, so we immediately have
$(\dagger) = O(\sqrt{T})$ using the regret bound for Regularized Follow-the-Leader in the OCO setting. To bound the second term $(\ddagger)$, we have the following result. 

{
\begin{theorem}\label{no_ocowk_regret_pdftl}
Let $X\subset \mathbb{R}^n$ be a convex compact set. Let $\{r_t(x)\}_{t=1}^T$ be an i.i.d. sequence of concave reward functions with $r_t:X\rightarrow \mathbb{R}_+$ which is $G$-Lipschitz with respect to norm $\Vert \cdot \Vert_2$. Let $\{c_t(x)\}_{t=1}^T$ be an i.i.d. sequence of vector-valued functions with $c_t:X \rightarrow \mathbb{R}^m$ where each entry of $c_t$ is convex in $x$ and $G$-Lipschitz with respect to norm $\Vert \cdot \Vert_2$. \textsf{PD-RFTL} run with $\eta_1 = \frac{D_X}{G(1+\Vert y_{max} \Vert_2)\sqrt{T}}, \eta_2 =\frac{\Vert y_{max}\Vert_2}{(\frac{1}{T}\Vert b\Vert_2 + \sqrt{m G D_X})\sqrt{T}} $ guarantees that  
\begin{align*}
\COCOreg(T) \leq 5 G (1+\Vert y_{max}\Vert_1) D_X \sqrt{T} + 5 (\frac{1}{T} \Vert b \Vert_2 + \sqrt{m G D_X}) \Vert y_{max}\Vert_2 \sqrt{T}.
\end{align*}
\end{theorem}
}

Compared to other algorithms for OCOwK, including the UCB-based algorithm in \cite{badanidiyuru2018bandits,agrawal2014bandits} and Thompson sampling-based algorithm in \cite{ferreira2017online}, the proof for Theorem~\ref{no_ocowk_regret_pdftl} is surprisingly simple, as we are able to exploit the connection between OCOwK and the OSP problem.

The $O(\sqrt{T})$ regret bound in Theorem~\ref{no_ocowk_regret_pdftl} also gives the best possible rate in $T$, since OCO is a special case of OCOwK, and it is well-known that any algorithm must have $\Omega(\sqrt{T})$ regret for the general OCO problem. In Section \ref{subsec:numerical-ocowk}, 
we compare the performance of \textsf{SP-FTL} and \textsf{PD-RFTL} in
a numerical experiment.

{
\begin{remark}\label{rmk:bandit-OCOwK}
Our proof for Theorem~\ref{no_ocowk_regret_pdftl} allows the RFTL subroutine in Algorithm~\ref{alg: PD-RFTL} to be replaced with other OCO algorithms with $O(\sqrt{T})$ regret.
In addition, we can extend Algorithm~\ref{alg: PD-RFTL} to the bandit setting of OCOwK, where we only observe the values $r_t(x_t)$ and $c_t(x_t)$ after $x_t$ is chosen. By replacing the RFTL subroutine with any Bandit Convex Optimization (BCO) algorithm \cite{bubeck2017kernel}, we can also establish sublinear regret bounds for OCOwK in the bandit setting.
 \end{remark}
 }

\subsection{Proof of SP-FTL for OCOwK}

In this section we present the analysis of \textsf{SP-FTL}, applied to the OCOwK problem.
The following result from Shalev-Shwartz et al.\ \cite{shalevstochastic} (Theorem 5) will be useful.
\begin{theorem}[\cite{shalevstochastic}]\label{shai_stoch}
Let $f(w,\xi):W \times \Xi \rightarrow \mathbb{R}$ be $G$-Lipschitz in $w$ with respect to norm $\Vert \cdot \Vert_2$, where $W\subset \mathbb{R}^d$ is bounded set such that there exists a constant $D_W>0$ such that $\max_{w_1,w_2 \in W}\Vert w_1 - w_2 \Vert_2 \leq D_W$. Then with probability at least $1-\delta$, for all $w\in W$, it holds that 
\begin{align*}
\Bigl| \sum_{t=1}^T f(w,\xi_t) - T \mathbb{E}_{\xi}[f(x,\xi)]\Bigr| \leq O\big( G D_W \sqrt{d \ln(T) \ln(\frac{d}{\delta}) T}\big).
\end{align*}  
\end{theorem}

First, we prove the following lemma.

\begin{lemma}\label{close_to_true_sp}
Let $\{\EL_t(x,y)\}_{t=1}^T$ be a sequence of i.i.d. functions which are $H$-strongly convex concave with respect $\Vert \cdot \Vert_2$ and $G$-Lipschitz with respect to norm $\Vert \cdot \Vert_2$. Here $\EL_t: X \times Y \rightarrow \mathbb{R}$, where $X\subset \mathbb{R}^{d_1}$, $Y \subset \mathbb{R}^{d_2}$ are convex compact sets. Let $\{(x_t,y_t)\}_{t=1}^T$ be the the iterates of SP-FTL when run on functions $\{\EL_t(x,y)\}_{t=1}^T$. With probability at least $1-\delta$, for any $t=1,...T$ it holds that 
\begin{align}
\Vert x_{t+1}-x^*\Vert_2 \leq O\big( \frac{G^{1/2} D_{XY}^{1/2} (d \ln(t) \ln(\frac{d}{\delta}) )^{1/4}}{H^{1/2} t^{1/4}}\big),
\end{align}
where $D_{XY}>0$ is a constant such that $\max_{z_1,z_2 \in X \times Y}\Vert z_1 - z_2\Vert_2 \leq D_{XY}$ and $d$ is the dimenssion of $X\times Y$.
\end{lemma}

\proof{Proof.}
Define the concentration error at time $t$ as
\begin{align}
CE_t \triangleq O\big( GD_{XY}\sqrt{d \ln(t) \ln(\frac{d}{\delta})t}\big). 
\end{align}
Notice that $\EL_\tau$ satisfies all the assumptions of Theorem \ref{shai_stoch}, so with probability at least $1-\delta$, for all $x\in X, y\in Y$ we have 
\begin{align}\label{eq_el_concentration}
\big| \sum_{\tau=1}^t \EL_\tau (x,y) - t \bar{\EL}(x,y) \big| \leq CE_t.
\end{align}
We now derive some consequences of this fact. Recall that $\bar{\EL}(x,y) \triangleq \mathbb{E}[\EL_1(x,y)]$ and $(x^*,y^*)$ is the saddle point of $\bar{\EL}$. With probability at least $1-\delta$.
\begin{align*}
t \bar{\EL}(x^*,y^*) &\leq t \bar{\EL}(x_{t+1},y^*) & \text{by definition of $x^*$} \\
&\leq \sum_{\tau=1}^t \EL_\tau(x_{t+1},y^*) + CE_t & \text{by Equation \eqref{eq_el_concentration}}\\
&\leq \sum_{\tau=1}^t \EL_\tau(x_{t+1},y_{t+1}) + CE_t. & \text{by definition of $y_{t+1}$}\\
\end{align*}
This implies that
\begin{align}\label{conseq_one}
t \bar{\EL}(x^*,y^*) -  \sum_{\tau=1}^t \EL_\tau(x_{t+1},y_{t+1}) \leq CE_t.
\end{align}
We now show that 
\begin{align}\label{conseq_two}
\sum_{\tau=1}^t \EL_\tau(x^*,y_{t+1}) - t\bar{\EL}(x^*,y^*) \leq CE_t.
\end{align}
Indeed, it holds that
\begin{align*}
\sum_{\tau=1}^t \EL_\tau(x^*,y_{t+1}) &\leq t \bar{\EL}(x^*,y_{t+1}) + CE_t & \text{by Equation \eqref{eq_el_concentration}}\\
&\leq t \bar{\EL}(x^*,y^*) + CE_t & \text{by definition of $y^*$}.
\end{align*}
Now, using the fact that $x_{t+1}$ is the saddle point of $\sum_{\tau=1}^t \EL_\tau(x,y)$, which is $(Ht)$-strongly convex, we have
\begin{align*}
\frac{Ht}{2} \Vert x_{t+1} - x^*\Vert_2^2 &\leq \sum_{\tau=1}^t \EL_\tau(x^*,y_{t+1}) - \sum_{\tau=1}^t \EL_\tau(x_{t+1},y_{t+1}) \\
&= \sum_{\tau=1}^t \EL_\tau(x^*,y_{t+1})-t\bar{\EL}(x^*,y^*) + t\bar{\EL}(x^*,y^*) - \sum_{\tau=1}^t \EL_\tau(x_{t+1},y_{t+1})\\
&\leq 2 CE_t \quad \text{by Equations \eqref{conseq_one} and \eqref{conseq_two}}.
\end{align*}
It follows that 
\begin{align*}
\Vert x_{t+1}-x^*\Vert_2 \leq O\big( \frac{G^{1/2} D_{XY}^{1/2} (d \ln(t) \ln(\frac{d}{\delta}) )^{1/4}}{H^{1/2} t^{1/4}}\big).
\end{align*}
\endproof

We now prove Lemma~\ref{no_sp_indiv_reg_str} in Section~\ref{subsec:algorithms-for-ocowk}.

\proof{Proof of Lemma~\ref{no_sp_indiv_reg_str}.}
For all $y\in Y$, it holds that 
\begin{align*}
\sum_{t=1}^T \EL_t(x_t,y) &\leq \sum_{t=1}^{T-1} \EL_t(x_t,y) + \EL_T(x^*,y) + G \Vert x_T-x^*\Vert_2 \qquad \text{since $\EL_T$ is $G$-Lipschitz}\\
&\leq \sum_{t=1}^T\EL_t(x^*,y) + G\sum_{t=1}^T \Vert x_t-x^*\Vert_2 \quad \qquad \text{since each $\EL_t$ is $G$-Lipschitz}\\
&\leq \sum_{t=1}^T\EL_t(x_{T+1},y)+ GT\Vert x^*-x_{T+1}\Vert_2 + G\sum_{t=1}^T \Vert x_t-x^*\Vert_2.
\end{align*}

It follows that 
\begin{align*}
\max_{y\in Y}\sum_{t=1}^T \EL_t(x_t,y) &\leq \max_{y\in Y} \sum_{t=1}^T\EL_t(x_{T+1},y)+ GT\Vert x^*-x_{T+1}\Vert_2 + G\sum_{t=1}^T \Vert x_t-x^*\Vert_2\\
&=\sum_{t=1}^T\EL_t(x_{T+1},y_{T+1})+ GT\Vert x^*-x_{T+1}\Vert_2 + G\sum_{t=1}^T \Vert x_t-x^*\Vert_2 \\
&  \qquad \text{(by definition of $y_{T+1}$)}\\
& = \MM \sum_{t=1}^T \EL_t(x,y) + GT\Vert x^*-x_{T+1}\Vert_2 + G\sum_{t=1}^T \Vert x_t-x^*\Vert_2.
\end{align*}

Subtracting in both sides $\sum_{t=1}^T\EL_t(x_t,y_t)$, we get
\begin{align*}
&\quad \max_{y\in Y}\sum_{t=1}^T \EL_t(x_t,y) -\sum_{t=1}^T\EL_t(x_t,y_t)\\
&\leq  \MM \sum_{t=1}^T \EL_t(x,y) -\sum_{t=1}^T\EL_t(x_t,y_t) + GT\Vert x^*-x_{T+1}\Vert_2 + G\sum_{t=1}^T \Vert x_t-x^*\Vert_2\\
&\leq \frac{8G^2}{H}(1+\ln(T)) + GT\Vert x^*-x_{T+1}\Vert_2 + G\sum_{t=1}^T \Vert x_t-x^*\Vert_2 \quad \text{By Theorem~\ref{thm:sp_regret_str}}.
\end{align*}

By Lemma \ref{close_to_true_sp}, and a simple union bound, we have that with probability at least $1-\delta T$ 
\begin{align*}
& \quad \frac{8G^2}{H}(1+\ln(T)) + GT\Vert x^*-x_{T+1}\Vert_2 + G\sum_{t=1}^T \Vert x_t-x^*\Vert_2\\
&\leq \frac{8G^2}{H}(1+\ln(T)) + \frac{G^{3/2}D_{XY}^{1/2}(d\ln(T)\ln(\frac{d}{\delta}))^{1/4}T^{3/4}}{H^{1/2}} + G\sum_{t=1}^T \Vert x_t-x^*\Vert_2 \\
&\leq \frac{8G^2}{H}(1+\ln(T)) + \frac{G^{3/2}D_{XY}^{1/2}(d\ln(T)\ln(\frac{d}{\delta}))^{1/4}T^{3/4}}{H^{1/2}} + \frac{G^{3/2} D_{XY}^{1/2}(d\ln(\frac{d}{\delta}))^{1/4}}{H^{1/2}} \sum_{t=1}^T \frac{\ln^{1/4}(t)}{t^{1/4}}.
\end{align*}

Therefore, with probability at least $1-\delta T$ it holds that
\begin{align*}
&\quad \max_{y\in Y}\sum_{t=1}^T \EL_t(x_t,y) -\sum_{t=1}^T\EL_t(x_t,y_t)\\
&\leq  \frac{8G^2}{H}(1+\ln(T)) + \frac{G^{3/2}D_{XY}^{1/2}(d\ln(T)\ln(\frac{d}{\delta}))^{1/4}T^{3/4}}{H^{1/2}} + \frac{G^{3/2} D_{XY}^{1/2}(d\ln(\frac{d}{\delta}))^{1/4}}{H^{1/2}} \int_{1}^T \frac{\ln^{1/4}(t)}{t^{1/4}} dt\\
&= \frac{8G^2}{H}(1+\ln(T)) + O\big( \frac{G^{3/2} D_{XY}^{1/2} (d \ln(T) \ln(\frac{d}{\delta}) )^{1/4} T^{3/4}\ln^{1/4}(T)}{H^{1/2}}\big).
\end{align*}
Setting $\delta=1/T^2$ yields the result. 

\endproof

We are ready to prove  Theorem \ref{no_sp_indiv_reg}.

\proof{Proof of Theorem \ref{no_sp_indiv_reg}.}
Recall that $\bar{\EL}_t(x,y) \triangleq L_t(x,y) + H \Vert x \Vert_2^2 - H \Vert y \Vert_2^2$. If $D_X$, $D_Y$ are constants such that $\max_{x\in X} \Vert x\Vert_2 \leq D_X$ and $\max_{y\in Y} \Vert y\Vert_2 \leq D_Y$, it holds that for all $x\in X$, $y\in Y$
\begin{equation}\label{dif_el_bar_el}
- HD_{Y}^2 \leq \bar{\EL}_t(x,y) - L_t(x,y) \leq HD_X^2, \quad \forall t=1,\ldots,T.
\end{equation}

By adding up equation \eqref{dif_el_bar_el}, we have that for any $y\in Y$
\begin{align*}
\sum_{t=1}^T L_t(x_t,y) \leq \sum_{t=1}^T \bar{\EL}_t(x_t,y) + HD_Y^2 T.
\end{align*}
This implies that
\begin{align*}
\max_{y\in Y}\sum_{t=1}^T L_t(x_t,y) \leq \max_{y \in Y}\sum_{t=1}^T \bar{\EL}_t(x_t,y) + HD_Y^2 T.
\end{align*}

Therefore, we have
\begin{align*}
& \quad\max_{y\in Y} \sum_{t=1}^T L_t(x_t,y) - \sum_{t=1}^T L_t(x_t,y_t)\\
&\leq \max_{y\in Y} \sum_{t=1}^T L_t(x_t,y) - \sum_{t=1}^T \bar{\EL}_t(x_t,y_t) + T H D_X^2 \qquad \text{by Equation \eqref{dif_el_bar_el}}\\
& \leq \max_{y\in Y} \sum_{t=1}^T \bar{\EL}_t(x_t,y) - \sum_{t=1}^T\bar{\EL}_t(x_t,y_t) + T H D_X^2 + THD_Y^2.
\end{align*}

Thus, by Lemma \ref{no_sp_indiv_reg_str} it holds that with probability at least $1-\frac{1}{T}$
{
\begin{align*}
& \quad\max_{y\in Y} \sum_{t=1}^T L_t(x_t,y) - \sum_{t=1}^T L_t(x_t,y_t)\\
&\leq \frac{8(G+2H(D_X+D_Y))^2}{H}(1+\ln(T)) \\
&\quad \quad + O\big( \frac{(G+2H(D_X+D_Y))^{3/2} D_{XY}^{1/2} (d \ln(T) \ln(dT) )^{1/4} T^{3/4}\ln^{1/4}(T)}{H^{1/2}}\big) + TH(D_X^2 + D_Y^2).
\end{align*}

Since 
\begin{align*}
&(G+2H(D_X+D_Y))^{3/2} \\
=\ & (G^3 + 6G^2H(D_X+D_Y) + 12GH^2(D_X+D_Y)^2 + 8H^3(D_X+D_Y)^3 )^{1/2}\\
\leq\ & G^{3/2} + (6G^2H(D_X+D_Y))^{1/2} + (12GH^2(D_X+D_Y)^2)^{1/2} + (8H^3(D_X+D_Y)^3)^{1/2} \\
\triangleq\ & \beta(G,H,D_X,D_Y),
\end{align*}
we have that with probability at least $1-\frac{1}{T}$,
\begin{align*}
& \quad\max_{y\in Y} \sum_{t=1}^T L_t(x_t,y) - \sum_{t=1}^T L_t(x_t,y_t)\\
&\leq \frac{8(G+2H(D_X+D_Y))^2}{H}(1+\ln(T))\\
&\qquad + O\big( \frac{\beta(G,H,D_X,D_Y) D_{XY}^{1/2} (d \ln(T) \ln(dT) )^{1/4} T^{3/4}\ln^{1/4}(T)}{H^{1/2}}\big) + TH(D_X^2 + D_Y^2)\\
& = O\left(poly(G,D_X, D_Y) \ln^{1/2}(T) T^{5/6} \right),
\end{align*}
}
where we plugged in $H=T^{-1/6}$. This concludes the first part of the proof. 

We now prove the second claim. Recall from Eq~\eqref{eq:regret-bound}, that
\begin{align*}
& \quad \COCOreg (T) \\
& \leq \Ex\Bigl[\max_{y\in Y}\sum_{t=1}^{T}L_{t}(x_{t},y)-\sum_{t=1}^{T}L_{t}(x_{t},y_{t})\Bigr] + \Ex\Bigl[\sum_{t=1}^{T}L_{t}(x_{t},y_{t})\Bigr]-\min_{x\in X}\max_{y\in Y}\Ex\Bigl[\sum_{t=1}^{T}L_{t}(x,y)\Bigr] \\
%& \leq O\left(poly(G,D_X, D_Y) \ln^{1/2}(T) T^{5/6} \right) + \Ex\Bigl[\sum_{t=1}^{T}L_{t}(x_{t},y_{t})\Bigr]-\min_{x\in X}\max_{y\in Y}\Ex\Bigl[\sum_{t=1}^{T}L_{t}(x,y)\Bigr] \\
%& \leq \tilde{O}(T^{5/6}) + \tilde{O}(T^{5/6}) = \tilde{O}(T^{5/6}),
& = \Ex\Bigl[\max_{y\in Y}\sum_{t=1}^{T}L_{t}(x_{t},y)-\sum_{t=1}^{T}L_{t}(x_{t},y_{t})\Bigr] + \Ex\Bigl[\sum_{t=1}^{T}L_{t}(x_{t},y_{t}) - \min_{x\in X} \max_{y\in Y} \sum_{t=1}^T L_t(x,y)\Bigr]\\
&\quad + \Ex\Bigl[ \min_{x\in X} \max_{y\in Y} \sum_{t=1}^T L_t(x,y)\Bigr] -\min_{x\in X}\max_{y\in Y}\Ex\Bigl[\sum_{t=1}^{T}L_{t}(x,y)\Bigr].
\end{align*}

Let 
\begin{align*}
A\triangleq & \max_{y\in Y}\sum_{t=1}^{T}L_{t}(x_{t},y)-\sum_{t=1}^{T}L_{t}(x_{t},y_{t}), \\
B \triangleq & \sum_{t=1}^{T}L_{t}(x_{t},y_{t}) - \min_{x\in X} \max_{y\in Y} \sum_{t=1}^T L_t(x,y), \\
C \triangleq & \min_{x\in X} \max_{y\in Y} \sum_{t=1}^T L_t(x,y) -\min_{x\in X}\max_{y\in Y}\Ex\Bigl[\sum_{t=1}^{T}L_{t}(x,y)\Bigr].
\end{align*}

Notice that we already have an upper bound for $A$ that holds with high probability, the term $B$ can be upper bounded using Theorem \ref{theorem:sp_regret_convex_concave}. Let us upper bound $C$ with high probability. As in the proof of Lemma \ref{close_to_true_sp} we know that with probability at least $1-\delta$ it holds that for all $x\in X, y \in Y$ $\left\vert \sum_{t=1}^T L_t(x,y) - \sum_{t=1}^T \mathbb{E}[L(x,y)] \right \vert \leq CE_T $. Therefore, with probability at least $1-\delta$, we have
\begin{align*}
    \sum_{t=1}^T L_t(x,y) \leq \sum_{t=1}^T \mathbb{E}[L(x,y)] + CE_T \quad \forall x\in X, y\in Y.
\end{align*}
This implies that with probability at least $1-\delta$
\begin{align*}
    \min_{x\in X}\sum_{t=1}^T L_t(x,y) \leq \sum_{t=1}^T \mathbb{E}[L(x,y)] + CE_T \quad \forall x\in X, y\in Y,
\end{align*}
which implies that with probability at least $1-\delta$
\begin{align*}
    \max_{y\in Y} \min_{x\in X}\sum_{t=1}^T L_t(x,y) \leq \max_{y\in Y} \sum_{t=1}^T \mathbb{E}[L(x,y)] + CE_T \quad \forall x\in X.
\end{align*}
Therefore it holds that with probability at least $1-\delta$
\begin{align*}
    \min_{x\in X}\max_{y\in Y} \sum_{t=1}^T L_t(x,y) \leq \min_{x\in X} \max_{y\in Y} \sum_{t=1}^T \mathbb{E}[L(x,y)] + CE_T,
\end{align*}
thus we have a high probability bound for $C$.
Our high probability bound for $A$ scales as $\asymp\ln^{1/2}(T)T^{5/6}$, the deterministic bound for $B$ scales as $\asymp\ln^{1/2}(T)T^{5/6}$, and the high probability bound for $C$ scales as
$\asymp\sqrt{\ln(T)T}$. The high probability bounds imply bounds in expectation (please see Lemmas 8 and 9 in \cite{pmlr-v89-cardoso19a} that show how to convert high probability bounds into bounds that hold in expectation). It follows that 
{
\begin{align*}
   \COCOreg (T) &\leq \mathbb{E}[A]+ \mathbb{E}[B] + \mathbb{E}[C]\\
    & = O\left(poly(G,D_X, D_Y) (\ln^{1/2}(T)T^{5/6} + \ln^{1/2}(T)T^{5/6} + \sqrt{\ln(T)T}) \right)\\
    & = O\left(poly(G,D_X, D_Y)\ln^{1/2}(T)T^{5/6}\right).
\end{align*}
}
This concludes the proof. 
\endproof

\subsection{Proof of PD-RFTL for OCOwK}
In this section we present the analysis of \textsf{PD-RFTL}, applied to the OCOwK problem.
%BEGIN Proof for PD FTL

\proof{Proof of Theorem \ref{no_ocowk_regret_pdftl}.}
Notice that \textsf{PD-RFTL} is using two instances of \textsf{RFTL}. One is run on convex functions $\{f_t(x)\}_{t=1}^T$ and another one on concave functions $\{g_t(y)\}_{t=1}^T$ (thus the maximization and the negative sign in the regularizer). Let $R_1 \triangleq 2\eta_1 G_f^2 T + \frac{D_X^2}{\eta_2}$ be the individual-regret guarantee of RFTL on functions $\{f_t(x)\}_{t=1}^T$ from Lemma \ref{thm:rftl_bound}. Let $R_2 \triangleq 2\eta_2 G_g^2 T + \frac{D_Y^2}{\eta_2}$ be the individual-regret guarantee of RFTL on functions $\{g_t(y)\}_{t=1}^T$ from Lemma \ref{thm:rftl_bound}. We have
\begin{equation}\label{eq:oco-ftl}
\sum_{t=1}^{T}L_{t}(x_{t},y_{t})=\sum_{t=1}^{T}f_{t}(x_{t})\leq\min_{x}\sum_{t=1}^{T}f_{t}(x)+R_1=\min_{x}\sum_{t=1}^{T}L_{t}(x,y_{t})+R_1,
\end{equation}
and
\begin{equation}\label{eq:oco-ftl-2}
\sum_{t=1}^{T}L_{t}(x_{t},y_{t})=\sum_{t=1}^{T}g_{t}(y_{t})\geq\max_{y}\sum_{t=1}^{T}g_{t}(y)-R_2=\max_{y}\sum_{t=1}^{T}L_{t}(x_{t},y)-R_2.
\end{equation}

Let $\bar{L}(x,y) = \Ex[L_t(x,y)]$ for any $x\in X,y\in Y$. Let $(x^*, y^*)$ be the saddle point of $\bar{L}$, satisfying
\begin{equation}\label{eq:def-y*}
\bar{L}(x^*, y^*) = \max_{y \in Y} \bar{L}(x^*,y)= \min_{x\in X} \max_{y \in Y} \bar{L}(x,y).
\end{equation}

{
Notice that 
\begin{equation}\label{eq:interchange_expect}
\mathbb{E}\left[\sum_{t=1}^{T}L_{t}(x^*,y_{t})\right] = \mathbb{E}\left[\sum_{t=1}^{T}( \overline{L}(x^*,y_{t}) ) \right],
\end{equation}
where the expectation is taken with respect to the random draw of functions $\{L_t\}_{t=1}^T$. Indeed, we have
\begin{align*}
& \quad \mathbb{E}\left[\sum_{t=1}^{T}L_{t}(x^*,y_{t})\right] \\
&= \mathbb{E}\left[\sum_{t=1}^{T-1}L_{t}(x^*,y_{t})\right] + \mathbb{E}\left[L_{T}(x^*,y_{T})\right]\\
&= \mathbb{E}\left[\sum_{t=1}^{T-1}L_{t}(x^*,y_{t})\right] + \mathbb{E}_{\{L_t\}_{t=1}^{T-1}}\left[ \mathbb{E}_{L_T} \left[L_{T}(x^*,y_{T})\vert\{L_t\}_{t=1}^{T-1}\right] \right]\\
&= \mathbb{E}\left[\sum_{t=1}^{T-1}L_{t}(x^*,y_{t})\right] + \mathbb{E}_{\{L_t\}_{t=1}^{T-1}}\left[ \bar{L}_{T}(x^*,y_{T}) \right],
\end{align*}
where the last equality holds since for any $x\in X, y \in Y$ it holds that $\bar{L}(x,y) = \Ex[L_t(x,y)]$ and $y_T$ is deterministic given $\{L_t\}_{t=1}^{T-1}$. We have $\mathbb{E}\left[\sum_{t=1}^{T}L_{t}(x^*,y_{t})\right] = \mathbb{E}\left[\sum_{t=1}^{T-1}L_{t}(x^*,y_{t})\right] + \mathbb{E}\left[ \bar{L}_{T}(x^*,y_{T}) \right]$, repeating the argument $T-1$ more times shows that $\mathbb{E}\left[\sum_{t=1}^{T}L_{t}(x^*,y_{t})\right] = \mathbb{E}\left[\sum_{t=1}^{T}( \overline{L}(x^*,y_{t}) ) \right]$.
}

We are ready to prove the statement of the theorem.
\begin{align*}
 & \quad \Ex\left[\max_{y}\sum_{t=1}^{T}L_{t}(x_{t},y)\right]-\min_{x\in X}\max_{y\geq0}\Ex\left[\sum_{t=1}^{T}L_{t}(x,y)\right]\\
 & \leq\Ex\left[\min_{x}\sum_{t=1}^{T}L_{t}(x,y_{t})\right]-\min_{x\in X}\max_{y\geq0}\Ex\left[\sum_{t=1}^{T}L_{t}(x,y)\right]+R_1 +R_2 \qquad \text{by Equations \eqref{eq:oco-ftl}, \eqref{eq:oco-ftl-2}}\\
 & =\Ex\left[\min_{x}\sum_{t=1}^{T}L_{t}(x,y_{t})\right] -\sum_{t=1}^T \overline{L}(x^{*},y^{*}) +R_1 +R_2 \qquad \text{by Equation \eqref{eq:def-y*}}\\
 & \leq\Ex\left[\sum_{t=1}^{T}L_{t}(x^*,y_{t})\right] -\sum_{t=1}^T \overline{L}(x^{*},y^{*}) +R_1 +R_2 \qquad \text{because } \min_{x}\sum_{t=1}^{T}L_{t}(x,y_{t}) \leq \sum_{t=1}^{T} L_{t}(x^*,y_{t}) \\ 
& = \Ex\left[\sum_{t=1}^{T}( \overline{L}(x^*,y_{t}) - \overline{L}(x^{*},y^{*}) ) \right] +R_1 +R_2 \qquad \text{by Equation \eqref{eq:interchange_expect}}\\ 
 & \leq0+R_1 +R_2. \qquad \text{by Equation~\eqref{eq:def-y*}}
\end{align*}
 
 {
By Equation \eqref{eq:regret-bound}, the above inequality implies that 
$\COCOreg(T) \leq 2\eta_1 G_f^2 T + \frac{D_X^2}{\eta_1} + 2\eta_2 G_g^2 T + \frac{D_Y^2}{\eta_2}$. Let us now bound $G_f, G_g$ and $D_Y$ from above. We start with $G_f$. To bound $G_f$, it suffices to bound $\Vert \nabla_x f_t(x) \Vert_2$. Let $y_{t,i}$ be the $i$-th entry of vector $y_t$ and $c_t(x)_i$ be the $i$-th entry of vector valued function $c_t$, which by assumption is $G$-Lipschitz continuous. We have
\begin{align*}
\Vert \nabla_x f_t(x) \Vert_2  &= \Vert -\nabla_x r_t(x) + \sum_{i=1}^m y_{t,i}\nabla_x c_t(x)_i\Vert_2\\
& \leq G + \sum_{i=1}^m y_{t,i} \Vert \nabla_x c_t(x)_i \Vert_2\\
& \leq G + G \Vert y_{max} \Vert_1.
\end{align*}
The previous line implies that $G_f\leq G + G \Vert y_{max} \Vert_1$. Let us now upper bound $\Vert \nabla_y g_t(y) \Vert_2$ to obtain a bound on $G_g$. Recall $g_t(y) = -r_t(x) - y^\top(\frac{b}{T}-c_t(x_t))$, and therefore we have
\begin{align*}
\Vert \nabla_y g_t(y) \Vert_2 &= \Vert \frac{b}{T} + c_t(x_t)\Vert_2\\
&\leq \frac{1}{T} \Vert b \Vert_2 + \Vert 0 - c_t(x_t)\Vert_2\\
&\leq \frac{1}{T} \Vert b \Vert_2 + \Vert [G \Vert x_0 - x_t\Vert_2;G \Vert x_0 - x_t\Vert_2;...,G \Vert x_0 - x_t\Vert_2] \Vert_2\\
&\leq \frac{1}{T} \Vert b \Vert_2 + \sqrt{\sum_{i=1}^m G D_X}\\
& = \frac{1}{T} \Vert b \Vert_2 + \sqrt{m G D_X}.
\end{align*}
Plugging in $\eta_1 = \frac{D_X \sqrt{T}}{G_f}, \eta_2 = \frac{D_Y \sqrt{T}}{G_g}$ we get 
\begin{align*}
\COCOreg(T) &\leq 5 G_f D_X \sqrt{T} + 5 G_g D_Y \sqrt{T}\\
&= 5 G (1+\Vert y_{max}\Vert_1) D_X \sqrt{T} + 5 (\frac{1}{T} \Vert b \Vert_2 + \sqrt{m G D_X}) \Vert y_{max}\Vert_2 \sqrt{T}.
\end{align*}
This concludes the proof.
}

\endproof

\section{Numerical Experiments}

\subsection{Individual-Regret and SP-Regret} \label{subsec:regret-simulation}
To further illustrate the relationship between SP-regret and individual-regret and the impossibility result of Theorem~\ref{thm:impossible},  we compare the performance of two online algorithms numerically. The first algorithm is \textsf{SP-FTL} defined in Algorithm~\ref{alg: SP-FTL}. %that achieve sublinear SP-regret by Theorems~\ref{thm:sp_regret_str} and \ref{thm:sp_regret_not_str}. 
In the second algorithm, which we call \textsf{OGDA}, player 1 applies online gradient descent to function $\EL_t(\cdot, y_t)$ and player 2 applies online gradient ascent to function $\EL_t(x_t, \cdot)$.

We generated two different instances. In both instances, we assume $X=Y=[-10,10]$. The payoff functions in both instances are the same for $t=1,...,\lfloor T/3\rfloor$, given by 
$
\EL_t(x,y) = xy + \frac{1}{2}\Vert x-2 \Vert^2 - \frac{1}{2}\Vert x+1 \Vert^2.
$
In Instance 1, for $t=\lfloor T/3\rfloor +1,...T $, we define 
$
\EL_t(x,y) = xy + \frac{1}{2}\Vert x+1 \Vert^2 - \frac{1}{2}\Vert x+2 \Vert^2.
$
In Instance 2 for $t=\lfloor T/3\rfloor +1,...T $, we define 
$
\EL_t(x,y) = xy + \frac{1}{2}\Vert x+1\Vert^2 - \frac{1}{2}\Vert x-3\Vert^2.
$
Since these functions are strongly convex-concave, when players use OGDA with step size $O(\frac{1}{t})$, they are both guaranteed logarithmic individual-regret. 

\begin{figure*}[!htb]
\centering
\includegraphics[width=2.72in, trim={0 0 0 1.3cm}, scale=1.5, clip]{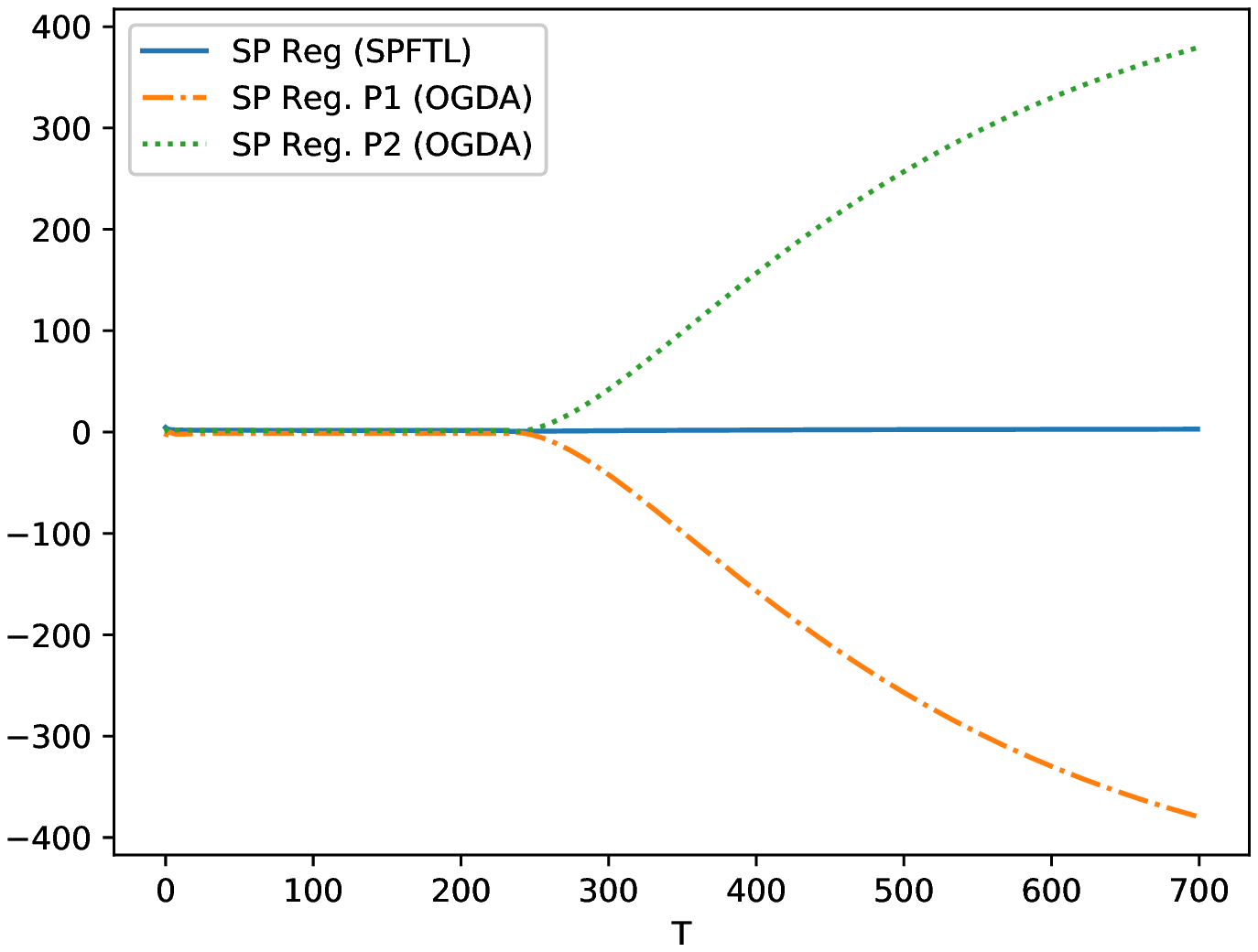}
\includegraphics[width=2.72in, trim={0 0 0 1.3cm}, scale=1.5, clip]{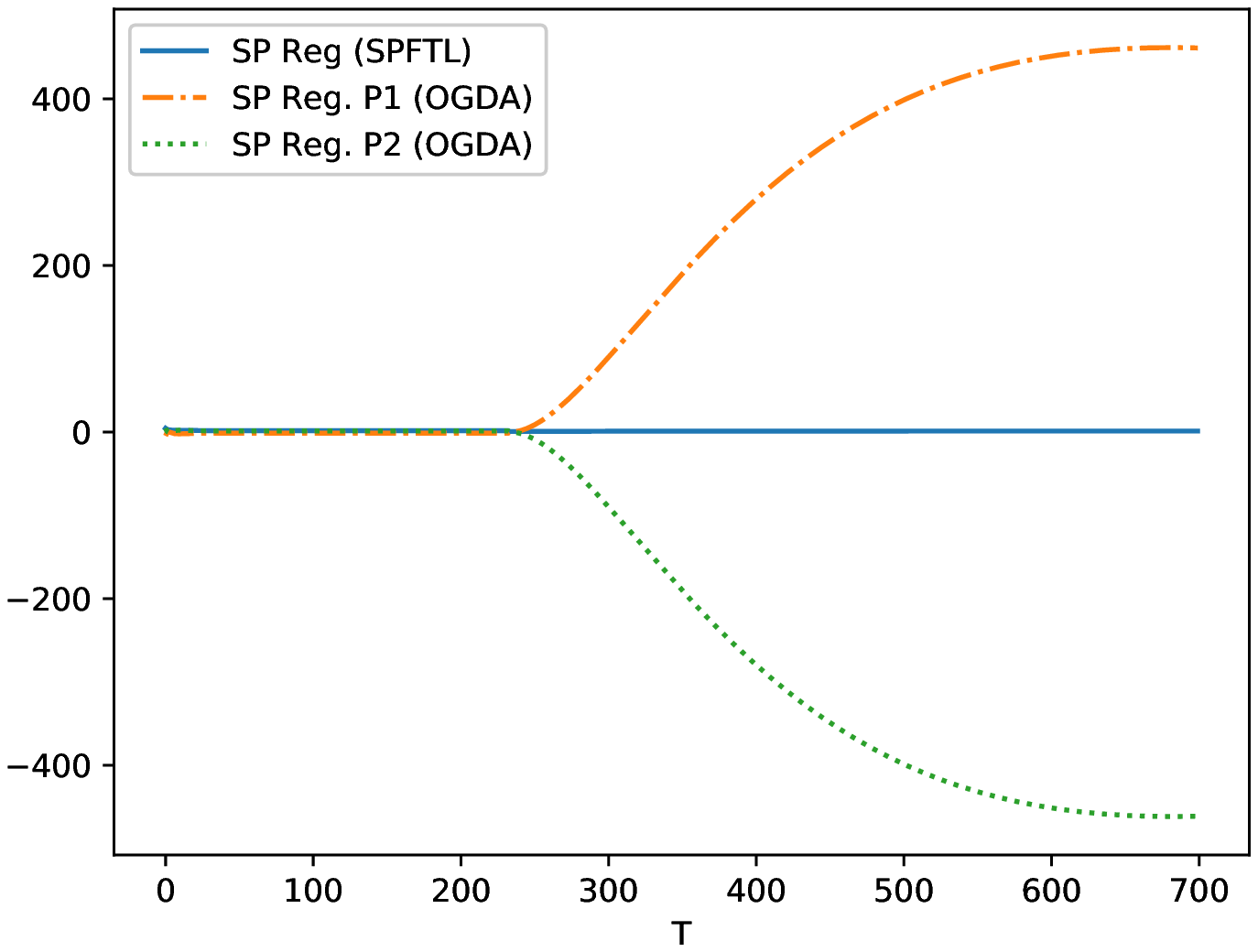}
\caption{SP-regret on instance 1 (left) and instance 2 (right) of SP-FTL and OGDA. 
\newline
\emph{Note.} Here we define the SP Regret of Player 1 (SP Reg.~P1) as $\sum_{t=1}^T \EL_t(x_t,y_t) - \MM \sum_{t=1}^T \EL_t(x,y)$ and the SP Regret of Player 2 (SP Reg.~P2) as $\MM \sum_{t=1}^T \EL_t(x,y)-\sum_{t=1}^T \EL_t(x_t,y_t)$. 
According to this definition, we have \textsc{SP Reg. P1}$=-$\textsc{SP Reg. P2}, 
and SP-regret of OGDA is equal to $|$\textsc{SP Reg. P1}$|=|$\textsc{SP Reg. P2}$|$.
%When using OGDA at one of the players performs considerably better than $\MM \sum_{t=1}^T \EL_t(x,y)$ (negative SP-regret) while the other player is far from  $\MM \sum_{t=1}^T \EL_t(x,y)$.
}
\label{fig:fig_one}
\end{figure*}
In Figure~\ref{fig:fig_one}, we plot the SP-regret of the two instances. On the left, it can be seen that the SP-regret of \textsf{OGDA} increases significantly after the payoff function switches at $\lfloor T/3\rfloor$, while the SP-regret of \textsf{SP-FTL} remains small throughout the entire horizon. 

\begin{figure*}[!htb]
\centering
\includegraphics[width=2.72in, trim={0 0 0 1.3cm}, clip]{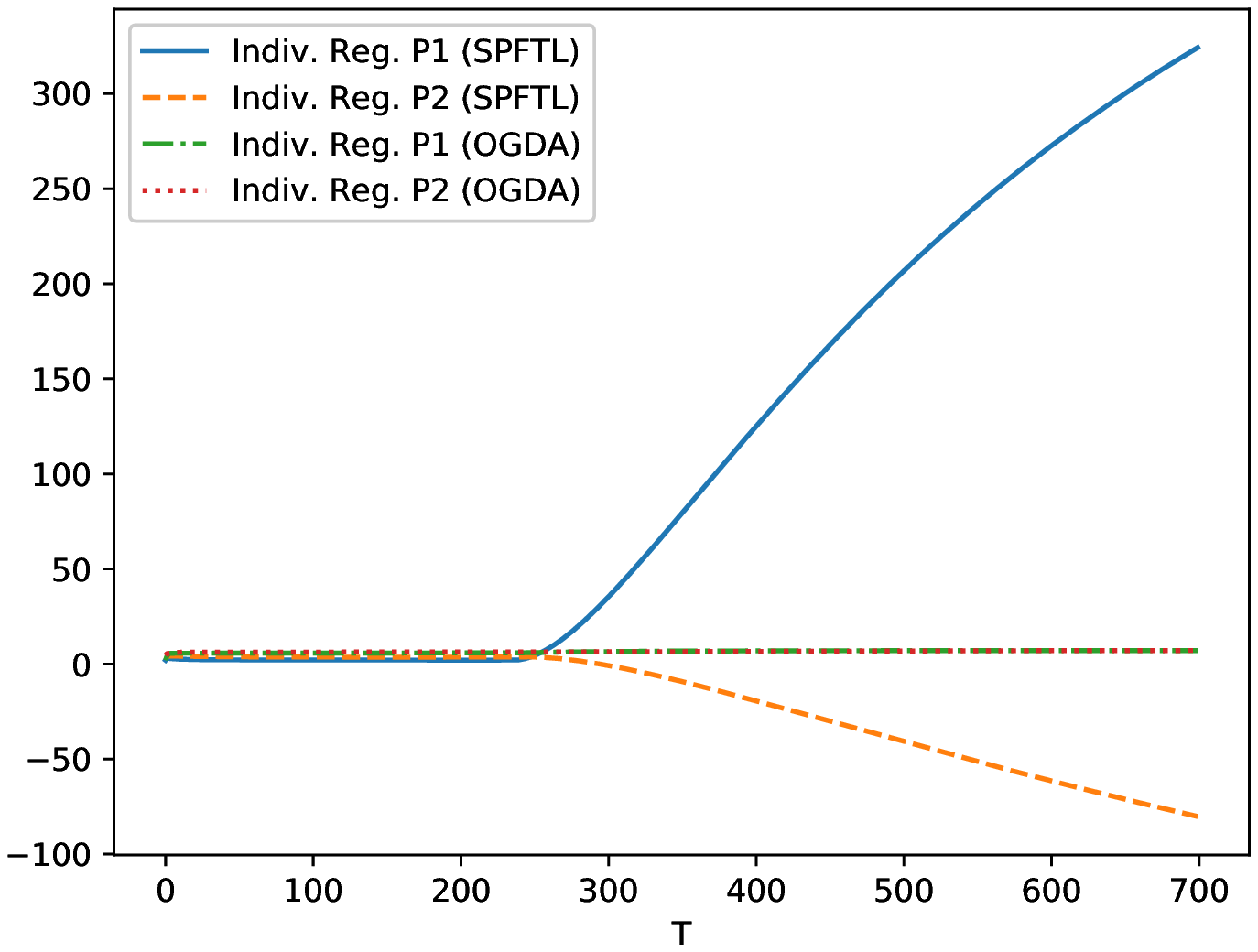}
\includegraphics[width=2.72in, trim={0 0 0 1.3cm}, clip]{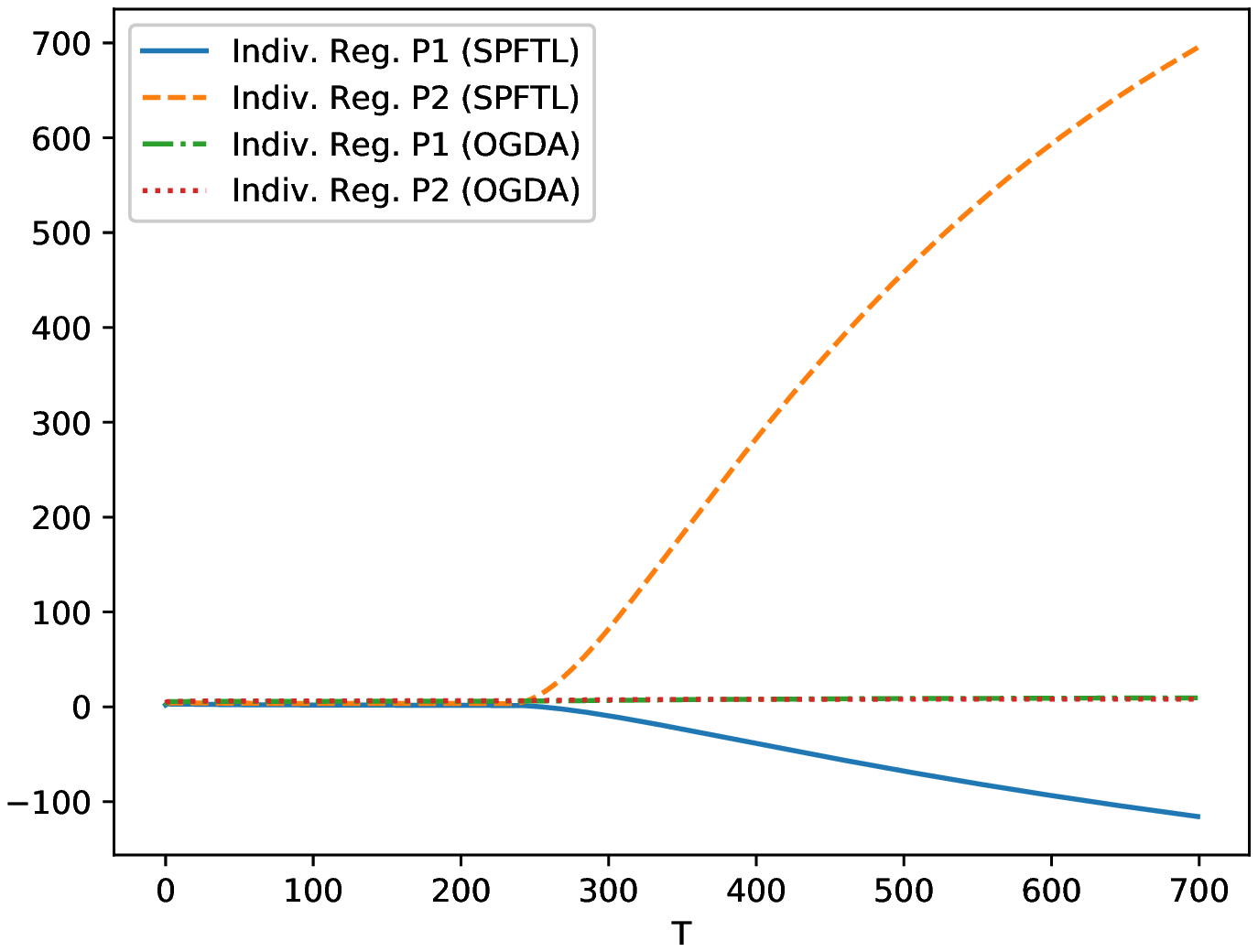}
\caption{Individual Regrets on Instance 1 (left) and Instance 2 (right) of Algorithms: SP-FTL and OGDA. 
\newline
\emph{Note.} The Individual Regret (Indiv.~Reg.) can be negative as we compare a sequence of \textit{dynamic} decisions against the best \textit{fixed} decision in hindsight.}
\label{fig:fig_two}
\end{figure*}

From Figure~\ref{fig:fig_two}, we can observe when both players use the \textsf{OGDA} algorithm, their individual-regrets are small. However, when when they use \textsf{SP-FTL}, at least one player suffers from high individual-regret. Figures~\ref{fig:fig_one} and \ref{fig:fig_two} verify Theorem~\ref{thm:impossible}, which states that no algorithm can achieve both sublinear SP-regret and sublinear individual-regret.

\subsection{SP-FTL and OGDA for the OCOwK problem}
\label{subsec:numerical-ocowk}

In this section, we compare the numerical performance of \textsf{SP-FTL} and \textsf{OGDA} (Online Gradient Descent/Ascent) for solving a OCOwK problem.  
In \textsf{OGDA}, player 1 applies online gradient descent to function $\EL_t(\cdot, y_t)$ and player 2 applies online gradient ascent to function $\EL_t(x_t, \cdot)$.
The proof for Theorem~\ref{no_ocowk_regret_pdftl} can also show that  \textsf{OGDA} has a regret of $O(\sqrt{T})$ (see Remark~\ref{rmk:bandit-OCOwK}).

We construct a numerical example where for each iteration $t=1,...,T$, the decision maker chooses an action $x_t \in X = [0,20]$. The reward function is $r_t = -x^2 + b_t x$ where $b_t \sim U[0,20]$. There are two types of resources with budgets $B_1$ and $B_2$. The consumption function for the first resource is given by $c_{t,1} = (a_t x)^2 + 50x$ where $a_t \sim U[0,3]$, and the consumption function for the  is $c_{t,2}=x$. We assume the budgets are some linear functions of $T$,  $B_1(T)$ and $B_2(T)$ respectively. In our simulations $B_1$ and $B_2$ are chosen so that playing the optimal solution to the problem without budgets is no longer optimal.   

\begin{figure*}[!htbp]
\centering
\includegraphics[width=3.5in, trim={0 0 0 1.3cm}, clip]{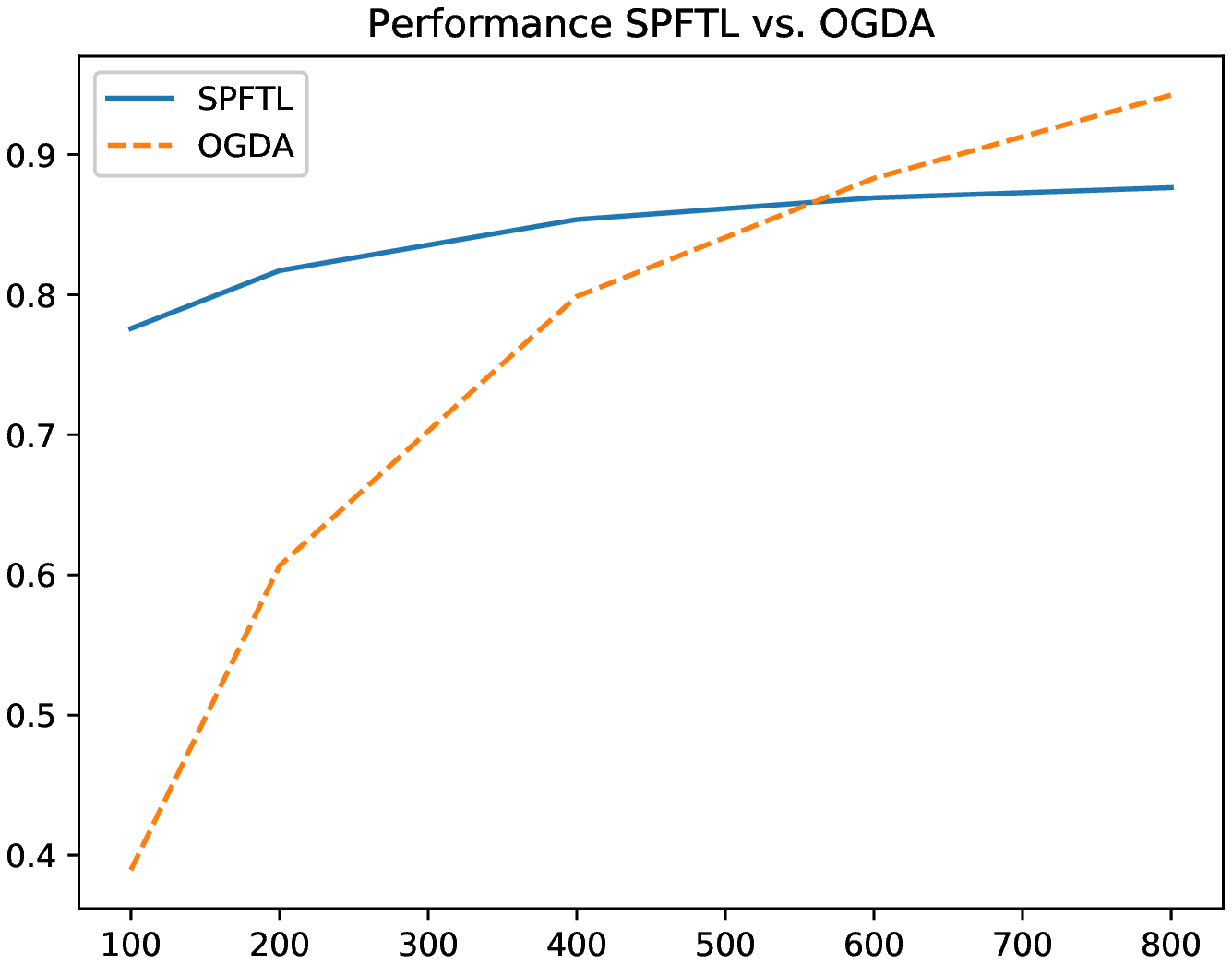}
\centering
\caption{Performance of SP-FTL and OGDA in the OCOwK problem.}
\label{fig:fig_three}
\end{figure*}

Figure \ref{fig:fig_three} compares the performance of \textsf{SP-FTL} vs \textsf{OGDA} on the OCOwK instance defined above. Performance is measured as the ratio of total reward incurred by the algorithm and the solution to Equation \eqref{eq:hindsight} across 25 simulation runs. It can be observed that both algorithms indeed improve their performance as $T$ increases. Moreover, it can be observed that while \textsf{OGDA} has worse performance for small values of $T$, the rate at which performance improves is greater than that for \textsf{SP-FTL}, which is consistent with our theoretical results that \textsf{SP-FTL} has $\tilde{O}(T^{5/6})$ regret and \textsf{OGDA} (or \textsf{PD-FTL}) has $O(\sqrt{T})$ regret.

\section{Conclusion}
In this paper we introduced the Online Saddle Point problem. In this problem, we consider two players that jointly play an arbitrary sequence of convex-concave games against Nature.
This problem is a generalization of the classical Online Convex Optimization problem, which focuses on a single player. The objective is to minimize the saddle-point regret (SP-Regret), defined as the absolute difference between the cumulative payoffs and the saddle point value of the game in hindsight.

We proposed an algorithm \textsf{SP-FTL} for the Online Saddle Point problem and showed that it achieves $\tilde{O}(\sqrt{T})$ SP-Regret for a game with $T$ periods. In the special case where the payoff functions are strongly convex-concave, we showed that the algorithm attains $O(\log T)$ SP-Regret. Furthermore, we proved that if the sequence of payoff functions are chosen arbitrarily, any algorithm with $o(T)$ regret for the Online Convex Optimization problem may incur $\Omega(T)$ SP-Regret in the worst case. 
{
We also consider the special case where the payoff functions are bilinear and the decision sets are the probability simplex. In this setting we are able to design algorithms that reduce the bounds on SP-Regret from a linear dependence in the dimension of the problem to a \textit{logarithmic} one. We also study the problem under bandit feedback and provide an algorithm that achieves sublinear SP-Regret.}
This implies that all existing algorithms for the Online Convex Optimization problem cannot be applied to the Online Saddle Point problem.
Moreover, we showed how our algorithm can be applied to solve the problem of Stochastic Online Convex Optimization with Knapsacks. Finally, we performed some numerical simulations to validate our results.

\bibliography{mybib}

\begin{thebibliography}{10}

\bibitem{abernethy2018faster}
J.~Abernethy, K.~A. Lai, K.~Y. Levy, and J.-K. Wang.
\newblock Faster rates for convex-concave games.
\newblock {\em arXiv preprint arXiv:1805.06792}, 2018.

\bibitem{abernethy2009competing}
J.~D. Abernethy, E.~Hazan, and A.~Rakhlin.
\newblock Competing in the dark: An efficient algorithm for bandit linear
  optimization.
\newblock {\em In Proceedings of the 21st Annual Conference on Learning Theory
  (COLT)}, 2009.

\bibitem{agrawal2014bandits}
S.~Agrawal and N.~R. Devanur.
\newblock Bandits with concave rewards and convex knapsacks.
\newblock In {\em Proceedings of the fifteenth ACM conference on Economics and
  computation}, pages 989--1006, 2014.

\bibitem{agrawal2014fast}
S.~Agrawal and N.~R. Devanur.
\newblock Fast algorithms for online stochastic convex programming.
\newblock In {\em Proceedings of the twenty-sixth annual ACM-SIAM symposium on
  Discrete algorithms}, pages 1405--1424, 2014.

\bibitem{agrawal2014dynamic}
S.~Agrawal, Z.~Wang, and Y.~Ye.
\newblock A dynamic near-optimal algorithm for online linear programming.
\newblock {\em Operations Research}, 62(4):876--890, 2014.

\bibitem{ahmadinejad2016duels}
A.~Ahmadinejad, S.~Dehghani, M.~Hajiaghayi, B.~Lucier, H.~Mahini, and
  S.~Seddighin.
\newblock From duels to battlefields: Computing equilibria of blotto and other
  games.
\newblock {\em Mathematics of Operations Research}, 44(4):1304--1325, 2019.

\bibitem{arrow1958studies}
K.~J. Arrow, L.~Hurwicz, and H.~Uzawa, editors.
\newblock {\em Studies in linear and non-linear programming}. Stanford
  Unversity Press, 1958.

\bibitem{auer1995gambling}
P.~Auer, N.~Cesa-Bianchi, Y.~Freund, and R.~Schapire.
\newblock Gambling in a rigged casino: The adversarial multi-armed bandit
  problem.
\newblock In {\em Proceedings of the 36th Annual Symposium on Foundations of
  Computer Science (FOCS)}, page 322, 1995.

\bibitem{aumann1987correlated}
R.~J. Aumann.
\newblock Correlated equilibrium as an expression of bayesian rationality.
\newblock {\em Econometrica}, pages 1--18, 1987.

\bibitem{badanidiyuru2018bandits}
A.~Badanidiyuru, R.~Kleinberg, and A.~Slivkins.
\newblock Bandits with knapsacks.
\newblock {\em Journal of the ACM (JACM)}, 65(3):13, 2018.

\bibitem{balduzzi2018mechanics}
D.~Balduzzi, S.~Racaniere, J.~Martens, J.~Foerster, K.~Tuyls, and T.~Graepel.
\newblock The mechanics of n-player differentiable games.
\newblock {\em arXiv preprint arXiv:1802.05642}, 2018.

\bibitem{balseiro2017learning}
S.~R. Balseiro and Y.~Gur.
\newblock Learning in repeated auctions with budgets: Regret minimization and
  equilibrium.
\newblock {\em Management Science}, 65(9):3952--3968, 2019.

\bibitem{bernstein2010online}
A.~Bernstein, S.~Mannor, and N.~Shimkin.
\newblock Online classification with specificity constraints.
\newblock In {\em Advances in Neural Information Processing Systems}, pages
  190--198, 2010.

\bibitem{besbes2012blind}
O.~Besbes and A.~Zeevi.
\newblock Blind network revenue management.
\newblock {\em Operations Research}, 60(6):1537--1550, 2012.

\bibitem{bowling2005convergence}
M.~Bowling.
\newblock Convergence and no-regret in multiagent learning.
\newblock In {\em Advances in Neural Information Processing Systems}, pages
  209--216, 2005.

\bibitem{bowling2001convergence}
M.~Bowling and M.~Veloso.
\newblock Convergence of gradient dynamics with a variable learning rate.
\newblock In {\em Proceedings of the Eighteenth International Conference on
  Machine Learning}, pages 27--34, 2001.

\bibitem{boyd2004convex}
S.~Boyd and L.~Vandenberghe.
\newblock {\em Convex optimization}.
\newblock Cambridge University Press, 2004.

\bibitem{bubeck2016kernel}
S.~Bubeck, Y.~T. Lee, and R.~Eldan.
\newblock Kernel-based methods for bandit convex optimization.
\newblock In {\em Proceedings of the 49th ACM Symposium on Theory of Computing
  (STOC)}, pages 72--85, 2017.

\bibitem{bubeck2017kernel}
S.~Bubeck, Y.~T. Lee, and R.~Eldan.
\newblock Kernel-based methods for bandit convex optimization.
\newblock In {\em Proceedings of the 49th Annual ACM SIGACT Symposium on Theory
  of Computing}, pages 72--85, 2017.

\bibitem{buchbinder2009online}
N.~Buchbinder and J.~Naor.
\newblock Online primal-dual algorithms for covering and packing.
\newblock {\em Mathematics of Operations Research}, 34(2):270--286, 2009.

\bibitem{pmlr-v89-cardoso19a}
A.~R. Cardoso and H.~Xu.
\newblock Risk-averse stochastic convex bandit.
\newblock In K.~Chaudhuri and M.~Sugiyama, editors, {\em Proceedings of Machine
  Learning Research}, volume~89 of {\em Proceedings of Machine Learning
  Research}, pages 39--47. PMLR, 16--18 Apr 2019.

\bibitem{cesa2006prediction}
N.~Cesa-Bianchi and G.~Lugosi.
\newblock {\em Prediction, Learning, and Games}.
\newblock Cambridge university press, 2006.

\bibitem{cesa2007improved}
N.~Cesa-Bianchi, Y.~Mansour, and G.~Stoltz.
\newblock Improved second-order bounds for prediction with expert advice.
\newblock {\em Machine Learning}, 66(2-3):321--352, 2007.

\bibitem{chen2018harnessing}
T.~Chen and G.~B. Giannakis.
\newblock Harnessing bandit online learning to low-latency fog computing.
\newblock In {\em 2018 IEEE International Conference on Acoustics, Speech and
  Signal Processing (ICASSP)}, pages 6418--6422. IEEE, 2018.

\bibitem{chowdhury2013experimental}
S.~M. Chowdhury, D.~Kovenock, and R.~M. Sheremeta.
\newblock An experimental investigation of colonel blotto games.
\newblock {\em Economic Theory}, 52(3):833--861, 2013.

\bibitem{conitzer2007awesome}
V.~Conitzer and T.~Sandholm.
\newblock Awesome: A general multiagent learning algorithm that converges in
  self-play and learns a best response against stationary opponents.
\newblock {\em Machine Learning}, 67(1-2):23--43, 2007.

\bibitem{cox2017decomposition}
B.~Cox, A.~Juditsky, and A.~Nemirovski.
\newblock Decomposition techniques for bilinear saddle point problems and
  variational inequalities with affine monotone operators.
\newblock {\em Journal of Optimization Theory and Applications},
  172(2):402--435, 2017.

\bibitem{ferreira2017online}
K.~J. Ferreira, D.~Simchi-Levi, and H.~Wang.
\newblock Online network revenue management using {T}hompson sampling.
\newblock {\em Operations Research}, 66(6):1586--1602, 2018.

\bibitem{flaxman2005online}
A.~D. Flaxman, A.~T. Kalai, and H.~B. McMahan.
\newblock Online convex optimization in the bandit setting: gradient descent
  without a gradient.
\newblock In {\em Proceedings of the sixteenth annual ACM-SIAM symposium on
  Discrete algorithms}, pages 385--394. Society for Industrial and Applied
  Mathematics, 2005.

\bibitem{gupta2016experts}
A.~Gupta and M.~Molinaro.
\newblock How the experts algorithm can help solve {LPs} online.
\newblock {\em Mathematics of Operations Research}, 41(4):1404--1431, 2016.

\bibitem{hazan2016introduction}
E.~Hazan.
\newblock Introduction to online convex optimization.
\newblock {\em Foundations and Trends{\textregistered} in Optimization},
  2(3-4):157--325, 2016.

\bibitem{hazan2007logarithmic}
E.~Hazan, A.~Agarwal, and S.~Kale.
\newblock Logarithmic regret algorithms for online convex optimization.
\newblock {\em Machine Learning}, 69(2-3):169--192, 2007.

\bibitem{hazan2014beyond}
E.~Hazan and S.~Kale.
\newblock Beyond the regret minimization barrier: optimal algorithms for
  stochastic strongly-convex optimization.
\newblock {\em The Journal of Machine Learning Research}, 15(1):2489--2512,
  2014.

\bibitem{hazan2016optimal}
E.~Hazan and Y.~Li.
\newblock An optimal algorithm for bandit convex optimization.
\newblock {\em arXiv preprint arXiv:1603.04350}, 2016.

\bibitem{ho2016role}
N.~Ho-Nguyen and F.~K{\i}l{\i}n{\c{c}}-Karzan.
\newblock The role of flexibility in structure-based acceleration for online
  convex optimization.
\newblock Technical report, Carnegie Mellon University, 2016.
\newblock Technical report.

\bibitem{immorlica2019adversarial}
N.~Immorlica, K.~A. Sankararaman, R.~Schapire, and A.~Slivkins.
\newblock Adversarial bandits with knapsacks.
\newblock In {\em 60th Annual Symposium on Foundations of Computer Science
  (FOCS)}, pages 202--219. IEEE, 2019.

\bibitem{jenatton2015adaptive}
R.~Jenatton, J.~Huang, and C.~Archambeau.
\newblock Adaptive algorithms for online convex optimization with long-term
  constraints.
\newblock {\em arXiv preprint arXiv:1512.07422}, 2015.

\bibitem{kalai2002}
A.~Kalai and S.~Vempala.
\newblock Efficient algorithms for universal portfolios.
\newblock {\em Journal of Machine Learning Research}, 3(Nov):423--440, 2002.

\bibitem{kovenock2012coalitional}
D.~Kovenock and B.~Roberson.
\newblock Coalitional colonel blotto games with application to the economics of
  alliances.
\newblock {\em Journal of Public Economic Theory}, 14(4):653--676, 2012.

\bibitem{lan2016algorithms}
G.~Lan and Z.~Zhou.
\newblock Algorithms for stochastic optimization with expectation constraints.
\newblock {\em arXiv preprint arXiv:1604.03887}, 2016.

\bibitem{laslier2002distributive}
J.-F. Laslier and N.~Picard.
\newblock Distributive politics and electoral competition.
\newblock {\em Journal of Economic Theory}, 103(1):106--130, 2002.

\bibitem{lu2007large}
Z.~Lu, A.~Nemirovski, and R.~D. Monteiro.
\newblock Large-scale semidefinite programming via a saddle point mirror-prox
  algorithm.
\newblock {\em Mathematical Programming}, 109(2-3):211--237, 2007.

\bibitem{mahdavi2012trading}
M.~Mahdavi, R.~Jin, and T.~Yang.
\newblock Trading regret for efficiency: online convex optimization with long
  term constraints.
\newblock {\em Journal of Machine Learning Research}, 13(Sep):2503--2528, 2012.

\bibitem{mahdavi2012online}
M.~Mahdavi, T.~Yang, and R.~Jin.
\newblock Online decision making under stochastic constraints.
\newblock In {\em NIPS workshop on Discrete Optimization in Machine Learning},
  2012.

\bibitem{mannor2009online}
S.~Mannor, J.~N. Tsitsiklis, and J.~Y. Yu.
\newblock Online learning with sample path constraints.
\newblock {\em Journal of Machine Learning Research}, 10(Mar):569--590, 2009.

\bibitem{myerson1993incentives}
R.~B. Myerson.
\newblock Incentives to cultivate favored minorities under alternative
  electoral systems.
\newblock {\em American Political Science Review}, 87(4):856--869, 1993.

\bibitem{neely2017online}
M.~J. Neely and H.~Yu.
\newblock Online convex optimization with time-varying constraints.
\newblock {\em arXiv preprint arXiv:1702.04783}, 2017.

\bibitem{presentation_arkadi}
A.~Nemirovski.
\newblock Deterministic and randomized first order saddle point methods for
  large-scale convex optimization, 2010.
\newblock Technical report.

\bibitem{nemirovski2009robust}
A.~Nemirovski, A.~Juditsky, G.~Lan, and A.~Shapiro.
\newblock Robust stochastic approximation approach to stochastic programming.
\newblock {\em SIAM Journal on Optimization}, 19(4):1574--1609, 2009.

\bibitem{paternain2015online}
S.~Paternain and A.~Ribeiro.
\newblock Online learning of feasible strategies in unknown environments.
\newblock In {\em American Control Conference (ACC), 2015}, pages 4231--4238.
  IEEE, 2015.

\bibitem{shalev2012online}
S.~Shalev-Shwartz et~al.
\newblock Online learning and online convex optimization.
\newblock {\em Foundations and Trends{\textregistered} in Machine Learning},
  4(2):107--194, 2012.

\bibitem{shalevstochastic}
S.~Shalev-Shwartz, O.~Shamir, N.~Srebro, and K.~Sridharan.
\newblock Stochastic convex optimization.
\newblock {\em Proceedings of the 22nd Annual Conference on Learning Theory},
  2009.

\bibitem{singh2000nash}
S.~Singh, M.~Kearns, and Y.~Mansour.
\newblock Nash convergence of gradient dynamics in general-sum games.
\newblock In {\em Proceedings of the Sixteenth conference on Uncertainty in
  artificial intelligence}, pages 541--548. Morgan Kaufmann Publishers Inc.,
  2000.

\bibitem{spall1997one}
J.~C. Spall.
\newblock A one-measurement form of simultaneous perturbation stochastic
  approximation.
\newblock {\em Automatica}, 33(1):109--112, 1997.

\bibitem{wu2015algorithms}
H.~Wu, R.~Srikant, X.~Liu, and C.~Jiang.
\newblock Algorithms with logarithmic or sublinear regret for constrained
  contextual bandits.
\newblock In {\em Advances in Neural Information Processing Systems}, pages
  433--441, 2015.

\bibitem{yu2017online}
H.~Yu, M.~Neely, and X.~Wei.
\newblock Online convex optimization with stochastic constraints.
\newblock In {\em Advances in Neural Information Processing Systems}, pages
  1427--1437, 2017.

\bibitem{yu2016low}
H.~Yu and M.~J. Neely.
\newblock A low complexity algorithm with $ o (\sqrt t) $ regret and finite
  constraint violations for online convex optimization with long term
  constraints.
\newblock {\em arXiv preprint arXiv:1604.02218}, 2016.

\bibitem{yuan2018online}
J.~Yuan and A.~Lamperski.
\newblock Online convex optimization for cumulative constraints.
\newblock {\em arXiv preprint arXiv:1802.06472}, 2018.

\bibitem{zinkevich2003online}
M.~Zinkevich.
\newblock Online convex programming and generalized infinitesimal gradient
  ascent.
\newblock In {\em Proceedings of the 20th International Conference on Machine
  Learning}, pages 928--936, 2003.

\end{thebibliography}
\bibliographystyle{abbrv} 

\end{document}